%% file: 0_main.tex
\documentclass[sigconf]{acmart}
\usepackage{soul}
\usepackage{bm} 
\usepackage{enumitem}
\usepackage{subcaption}
\usepackage{caption}
\usepackage{amsmath}
\usepackage{hyperref}
\usepackage{algorithm}
\usepackage{algorithmic}
\usepackage{stfloats}
\usepackage[table]{xcolor} 
\definecolor{Green}{rgb}{0.0, 0.8, 0.4}   
\definecolor{Blue}{rgb}{0.2, 0.6, 1.0}   
\definecolor{Yellow}{rgb}{0.95, 0.75, 0.1} 
\definecolor{Red}{rgb}{1.0, 0.2, 0.2} 
\definecolor{Orange}{rgb}{1.0, 0.5, 0.0}

\definecolor{lightorange}{RGB}{255,245,230}
\definecolor{lightgreen}{RGB}{230,250,235}
\definecolor{lightpurple}{RGB}{245,240,255}
\definecolor{textorange}{RGB}{255,128,0}
\definecolor{textgreen}{RGB}{0,153,76}
\definecolor{textpurple}{RGB}{102,0,204}
\AtBeginDocument{%
  }

\setcopyright{acmlicensed}
\copyrightyear{2026}
\acmYear{2026}
\acmDOI{XXXXXXX.XXXXXXX}

\acmConference[Conference acronym 'XX]{Make sure to enter the correct
  conference title from your rights confirmation email}{June 03--05,
  2026}{Woodstock, NY}
\acmISBN{978-1-4503-XXXX-X/18/06}

\begin{document}
\title{Property-Driven Evaluation of GNN Expressiveness at Scale: \\ Datasets, Framework, and Study}

\settopmatter{authorsperrow=4}

\author{Sicong Che}
\email{sicongche@utexas.edu}
\affiliation{%
  \small\institution{University of Texas at Austin}
  \country{}
}

\author{Jiayi Yang}
\email{jiayiyang1997@utexas.edu}
\affiliation{%
  \small\institution{University of Texas at Austin}
  \country{}
}

\author{Sarfraz Khurshid}
\email{khurshid@ece.utexas.edu}
\affiliation{%
  \small\institution{University of Texas at Austin}
  \country{}
}

\author{Wenxi Wang}
\email{wenxiw@virginia.edu}
\affiliation{%
  \small\institution{University of Virginia}
  \country{}
}

\renewcommand{\shortauthors}{Che et al.}

\input{abstract}
\maketitle 

\input{1_introduction}

\input{2_background}
\input{3_related_work}

\input{4_dataset}

\input{5_metrics}
\input{6_study}

\input{7_exp}
\input{8_conclusion}



\bibliographystyle{ACM-Reference-Format}
\bibliography{reference}

\input{9_appendix}
\end{document}

%% file: abstract.tex
\begin{abstract}

Advancing trustworthy AI requires principled approaches to model evaluation. Graph Neural Networks (GNNs) excel at processing graph-structured data; however, their expressiveness, particularly in capturing fundamental graph properties, remains an open challenge. We address this by developing a property-driven evaluation methodology grounded in rigorous benchmarking using formal specification, systematic evaluation, and empirical study.

Leveraging Alloy, a widely used software specification language and analyzer, we introduce a configurable graph dataset generator that produces \emph{two dataset families}: GraphRandom, containing diverse graphs that either satisfy or violate specific properties, and GraphPerturb, introducing controlled graph structural variations to stress-test models. Together, these benchmarks encompass 352   datasets, each with at least 10,000 labeled graphs, covering 16 fundamental graph properties critical to both theoretical and real-world domains such as distributed systems, knowledge graphs, and biological networks.


Building on these datasets, we \emph{propose a general evaluation framework} that assesses three key aspects (generalizability, sensitivity, and robustness) and introduces two new quantitative metrics. Leveraging this framework, we conduct the \emph{first study} on the impact of global pooling methods on GNN expressiveness. Our findings provide critical insights into how pooling strategies influence graph-level representation learning, highlighting fundamental limitations of existing pooling strategies and opening several research directions. By embedding formal specification rigor into AI evaluation, this work establishes a principled foundation for developing GNN architectures that combine expressive power with reliability across a wide range of real-world applications.

\end{abstract}


%% file: 1_introduction.tex
\section{Introduction} 
Graph Neural Networks (GNNs) \cite{wu2020comprehensive, zhou2020graph} excel at processing graph-structured data, driving advancements in diverse applications such as social network analysis \cite{kipf2016semi, velivckovic2017graph}, molecular chemistry \cite{you2018graph, gilmer2017neural}, recommendation systems \cite{fan2019graph, wu2022graph}, and knowledge graphs \cite{dettmers2018convolutional}. The expressiveness of GNNs has been extensively studied~\cite{sato2020survey, morris2023weisfeiler, wang2024empirical}, with the Weisfeiler-Lehman (WL) test~\cite{Weisfeiler1968} commonly used to assess a model’s ability to distinguish graph structures. 
More recently, Zhang et al.~\cite{zhang2023rethinking} proposed a new perspective by systematically evaluating GNN expressiveness through the lens of a specific graph property, \texttt{biconnectivity}, and introduced evaluation metrics beyond the WL hierarchy to quantify expressive power. While their work provides a valuable framework, it is limited in scope: it focuses solely on a single property and lacks generality for evaluating expressiveness across a broader range of graph properties.




To address the limitation, we evaluate GNN expressiveness across 16 fundamental graph properties, as shown in Table~\ref{tab:relational_properties_natural}. We classified the 16 properties into three groups: basic properties, function-related properties, and combined properties, defined as follows.
\begin{itemize} [nosep, left=0pt]
    \item \textcolor{textorange}{\textbf{Basic properties}}: define fundamental constraints on directed relationships within graphs, governing self-loops, mutual edges, and pairwise connectivity. They include \texttt{antisymmetry}, \texttt{connex}, \texttt{reflexivity}, \texttt{irreflexivity}, and \texttt{transitivity}.
    \item \textcolor{textgreen}{\textbf{Function-related properties}}: describe how graph nodes map to one another, analogous to mathematical functions. They include \texttt{function}, \texttt{functionality}, \texttt{injectivity}, \texttt{surjectivity}, and \texttt{bijectivity}. 
    \item \textcolor{textpurple}{\textbf{Combined properties}}:   emerge from combining basic properties, forming well-known structures such as equivalence and orders. They include \texttt{equivalence}, \texttt{partial order}, \texttt{preorder}, \texttt{strict order}, \texttt{non-strict order}, and \texttt{total order}.
\end{itemize}

\input{property_table.tex}

Beyond the theoretical significance of these 16 properties, they play a crucial role in real-world applications across various domains. For instance, \texttt{total order} ensures consistency and coordination in distributed systems~\cite{lamport2019time}, while \texttt{antisymmetry} is fundamental in knowledge bases~\cite{fellbaum1998wordnet, manabe2018data}. \texttt{Reflexivity} models self-regulating genes in gene regulatory networks, playing a key role in homeostasis and environmental response~\cite{Shen-Orr2002}. Similarly, \texttt{transitivity} helps analyze group interactions in real-world hypergraphs, providing insights into social network dynamics~\cite{Kim_2023}. Moreover, a tournament graph, in which every pair of distinct nodes is connected by exactly one directed edge, inherently satisfies the \texttt{connex} property~\cite{moon1968topics}. 
Given the foundational importance of these properties and their widespread real-world applications, it is crucial for GNNs to exhibit strong expressive power in capturing them. 

To evaluate GNN expressiveness in this context, we first \textbf{aim to construct comprehensive datasets covering these 16 fundamental graph properties}. 
The goal is to efficiently generate class-balanced graphs that either satisfy (positive samples) or violate (negative samples) a given property. A straightforward approach is to randomly generate graphs and filter them using a property checker. However, this method is highly inefficient, as positive samples often represent a tiny portion of the graph space (e.g., for the \texttt{total order} property with graph size 13,  the proportion of positive graph samples is only $2.07 \times 10^{-47}$). 


To address this challenge, \textbf{we leverage Alloy}~\cite{jackson2002alloy}, which serves as both a lightweight software specification language and an Analyzer. Based on relational logic, Alloy specification language maps naturally to graph structures, enabling concise encoding of diverse graph properties. The Alloy Analyzer supports bounded, exhaustive enumeration of graphs that either satisfy or violate the specified property, eliminating the need for post-generation filtering or checking. Building on these capabilities, \textbf{we transform Alloy into a flexible, reproducible graph dataset generator tailored for evaluating GNN expressiveness.}

As a result, we apply our Alloy-based graph generator to produce two dataset families, \textbf{GraphRandom and GraphPerturb}. 
\textbf{GraphRandom comprises 176 datasets} across 16 graph properties. Each dataset includes positive samples (graphs satisfying the property) and an equal number of randomly generated negative samples (graphs that do not), spanning 11 different graph sizes per property. GraphPerturb builds on GraphRandom by introducing datasets in which each positive sample is paired with a structurally similar negative counterpart that differs by only one or two edges. As a result, \textbf{GraphPerturb contains 176 datasets}, enabling rigorous testing of GNNs in distinguishing structurally similar graphs with contrasting labels.

\textbf{Based on our generated datasets, we propose a general framework that systematically assesses three key aspects of GNN expressiveness: generalizability, sensitivity, and robustness}. Specifically, using different combinations of our datasets, we design structured training and testing strategies to examine how well GNNs capture graph properties, differentiate structurally similar graphs, and generalize to unseen variations. In addition, to quantify performance, we introduce two evaluation metrics: 1) Unified Score, a weighted accuracy measure prioritizing larger graphs for fair cross-dataset comparisons; and 2) Relative Score, which normalizes performance against peer GNN models to highlight strengths and weaknesses across graph properties and evaluation aspects. This framework provides a standardized and rigorous benchmark for GNN expressiveness.


Although our evaluation framework is general, capable of assessing GNN expressiveness across a variety of dimensions, in this paper we focus on evaluating \emph{the impact of global pooling methods} on GNN expressiveness. Global pooling plays a crucial role in graph-level tasks, yet its impact on GNN expressiveness remains largely unexplored. Existing research~\cite{xu2018powerful, you2021identity, barcelo2021graph, bouritsas2022improving} has primarily focused on enhancing GNN expressiveness through node-level feature augmentation and message-passing mechanisms, \emph{often overlooking the role of pooling in aggregating information for graph-level representations}. However, even if node embeddings capture rich local structures, an ineffective global pooling mechanism can lead to significant information loss, ultimately affecting downstream tasks such as graph classification. To address this gap, we \textbf{use our framework to conduct the \emph{first study} on how global pooling affects GNN expressiveness.}

Our study systematically evaluates nine state-of-the-art global pooling methods across three key aspects and 16 graph properties, revealing numerous insights. We find that while most pooling methods generalize well to larger graphs, they often struggle with sensitivity and robustness, particularly in distinguishing subtle structural differences or adapting to distributional shifts. No single pooling method consistently excels across all properties; attention-based approaches show strong generalizability and robustness, while second-order pooling offers better sensitivity. These findings point to key limitations in existing pooling strategies and highlight several \ul{promising research directions, including property-aware adaptive pooling, graph-size-aware architectures, robustness-oriented training, and hybrid designs that unify attention and second-order mechanisms}. Such directions may pave the way for more expressive and reliable GNN models in graph-level tasks.

Our contributions are summarized as follows:
\begin{itemize}[nosep, left=0pt]
\item \textbf{Datasets}: We transform Alloy into a flexible and reproducible graph dataset generator, producing \emph{two dataset families}, GraphRandom and GraphPerturb, totaling 352 generated datasets. Our datasets and code are publicly available at: 
\url{https://anonymous.4open.science/r/Property-Driven-GNN/}. 
\item \textbf{Evaluation Framework}: We propose 
a \emph{general framework} for evaluating GNN expressiveness across three key aspects and 16 graph properties, using two novel quantitative metrics. 
\item \textbf{Study}: We conduct the \emph{first study} on the impact of global pooling on GNN expressiveness, aiming to inform the development of more effective global pooling methods.

\end{itemize}

%% file: property_table.tex
\begin{table}[b]
\centering
    \vspace{-6ex}
  \caption{Definitions of 16 graph properties.} 
  \vspace{-3ex}
\scalebox{0.8}{
    \begin{tabular}{||m{2.1cm}|m{6.7cm}|>{\centering\arraybackslash}m{0.6cm}||}
    \hline
    \textbf{Property} & \textbf{Conceptual Definition} & \textbf{Base Size}  \\ \hline
    \rowcolor{lightorange}
    Antisymmetry     &  No two distinct nodes in the graph have edges in both directions between them. & 5\\ \hline
    \rowcolor{lightorange}
    Connex &For any two different nodes $u$ and 
$v$, either there is an edge from 
$u$ to $v$ or an edge from $v$ to $u$ (or both). &6 \\ \hline
    \rowcolor{lightorange}
    Reflexivity         &Every node in the graph has a self-loop.  & 5      \\ \hline
    \rowcolor{lightorange}
    Irreflexivity       &No node in the graph has a self-loop.  & 5  \\ \hline
    \rowcolor{lightorange}
    Transitivity    &If for all nodes \(u\), \(v\), \(w\), whenever there are edges \(u\) $\rightarrow$ \(v\) and \(v\) $\rightarrow$ \(w\), there is an edge \(u\) $\rightarrow$ \(w\).     & 6 \\ \hline

    \rowcolor{lightgreen}
    Function    & Each node has exactly one outgoing edge. & 8\\ \hline
    \rowcolor{lightgreen}
    Functionality  & Each node has at most one outgoing edge.  & 8    \\ \hline
    \rowcolor{lightgreen}
    Injectivity   & No two distinct nodes have edges directed to the same node. & 8\\ \hline
    \rowcolor{lightgreen}
    Surjectivity  & Each node has at least one incoming edge.  & 14  \\ \hline
    \rowcolor{lightgreen}
    Bijectivity    & Each node has exactly one incoming and one outgoing edge.  & 14      \\ \hline
    
    \rowcolor{lightpurple}
    Equivalence   & The graph is reflexive, symmetric, and transitive.  & 20   \\ \hline
    \rowcolor{lightpurple}
    Partial order   & The graph is reflexive, antisymmetric, and transitive. &6      \\ \hline
    \rowcolor{lightpurple}
    Preorder    & The graph is reflexive and transitive.     &  7   \\ \hline
    \rowcolor{lightpurple}
    Strict order  & The graph is irreflexive and transitive. & 7
    \\ \hline
    \rowcolor{lightpurple}
    Non-strict order  & The graph is reflexive, antisymmetric, and transitive. & 7      \\ \hline
    \rowcolor{lightpurple}
    Total order    & The graph is reflexive, antisymmetric, transitive, and connex.    & 13   \\ \hline
    \end{tabular}
     }

\label{tab:relational_properties_natural}
    \end{table}

%% file: 2_background.tex
\vspace{-2ex}
\section{Background: Alloy}

Alloy is a lightweight modeling language for specifying and analyzing relational structures in a concise and declarative manner. Its core tool, Alloy Analyzer, enables users to explore possible solutions or examples that satisfy specified properties, providing valuable insights into the behavior of complex systems. This makes Alloy particularly useful for tasks requiring precise generation or validation of data structures under constraints.

An Alloy specification comprises \emph{relations} (sets of objects) and \emph{constraints} (logical rules applied to these relations). For instance, a relation might represent connections between nodes in a graph, while constraints define properties such as connex or symmetry. With relational logic, Alloy allows users to express these rules using quantifiers like \textit{all} (universal), \textit{some} (existential), and \textit{no} (none), combined with logical operators such as \textit{and}, \textit{or}, and \textit{not}. This enables the creation of expressive specifications for describing and reasoning about interconnected data. The Alloy Analyzer translates specifications into Boolean logical formulas, which are then solved by SAT (Boolean satisfiable) solvers, specialized tools for efficiently finding SAT  instances. For example, the Analyzer can verify if a graph satisfies certain relational properties or generate graphs that meet specified constraints.

Figure~\ref{fig:random} shows an Alloy specification where \texttt{sig Node} defines a set of nodes, and \texttt{edge} represents a relation mapping nodes to one another. The predicate \texttt{Reflexivity()} enforces the reflexivity property, requiring that every node \(n\) has an edge pointing to itself. The Alloy Command executes the \texttt{Reflexivity()} predicate, invoking the Alloy Analyzer to generate graphs with five nodes.




%% file: 3_related_work.tex
\vspace{-2ex}
\section{Related Work}
The expressiveness of GNNs has been extensively studied~\cite{sato2020survey, morris2023weisfeiler, wang2024empirical} and is often evaluated using the Weisfeiler-Lehman (WL) test~\cite{Weisfeiler1968}, which measures a model’s ability to distinguish graph structures. Standard message-passing GNNs are at most as powerful as 1-WL test~\cite{xu2018powerful} and could struggle with distinguishing certain graph structures due to non-injective aggregation. The GIN model~\cite{xu2018powerful} matches the expressiveness of 1-WL test, but like other 1-GNNs, it inherits limitations of the 1-WL test in differentiating specific graph structures.

\begin{figure*}[t!]
    \centering
    \includegraphics[width=0.85\textwidth]{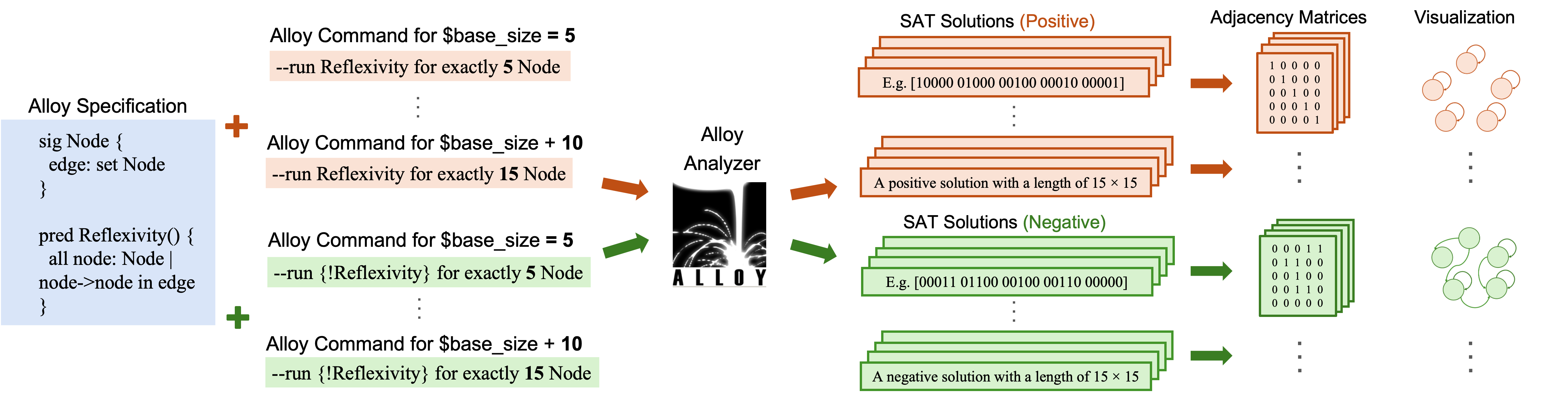} 
    \vspace{-3ex}
    \caption{GraphRandom dataset family generation. 
    } 
    \vspace{-4ex}
    \label{fig:random} 
\end{figure*}

To overcome the limitations of 1-GNNs, various approaches have been proposed. \emph{Higher-order GNNs}, inspired by the k-WL test~\cite{morris2023weisfeiler}, extend neighborhood aggregation to k-tuple subgraphs, allowing models to capture richer structural information \cite{morris2019weisfeiler, maron2019universality, keriven2019universal,azizian2020expressive}. \emph{Subgraph-based GNNs} enhance expressiveness by partitioning graphs into smaller subgraphs using strategies such as node deletion~\cite{cotta2021reconstruction}, edge deletion~\cite{bevilacquaequivariant}, and ego networks \cite{bevilacquaequivariant, you2021identity, zhang2021nested}, as well as the combination of these strategies~\cite{bevilacquaequivariant}. Meanwhile, \emph{substructure-based GNNs} encode counts of graph motifs (e.g., cycles, cliques) into feature representations to improve structural discrimination \cite{bouritsas2022improving, barcelo2021graph}. \emph{Transformer-based GNNs} integrate structural encodings and self-attention mechanisms to improve representation power beyond traditional message-passing models \cite{ying2021transformers, zhou2024theoretical}. \emph{Randomization-enhanced GNNs}, such as DropGNN \cite{papp2021dropgnn}, DropMessage~\cite{Fang_2023}, random node initialization~\cite{abboud2020surprising}, and random feature addition~\cite{sato2021random}, introduce stochastic perturbations during training to improve generalization and expressiveness.

Recent work explores fundamental graph properties to systematically improve GNN expressiveness. Zhang et al. \cite{zhang2023rethinking} introduced biconnectivity as an expressivity metric, revealing that most existing GNNs fail on biconnectivity tasks.  To address this, they proposed the Generalized Distance (GD) WL algorithm, incorporating a distance metric into WL aggregation, leading to Graphormer-GD, which outperforms prior models on biconnectivity tasks. 
This shift towards graph property-driven GNNs provides a structured approach to designing more expressive models. Inspired by this direction, our work systematically explores 16 additional graph properties as expressivity measures, introduces new datasets, and defines three key evaluation metrics: generalizability, sensitivity, and robustness. 

%% file: 4_dataset.tex
\vspace{-2ex}
\section{Graph Dataset Generation using Alloy}
\label{sec:dataset}
To investigate the expressive power of GNNs in capturing graph properties, we developed a comprehensive suite of graph-level classification datasets. Our design follows widely used benchmarks in real-world domains such as chemistry and social-network analysis, which typically involve small graphs. For example, MUTAG~\cite{debnath1991structure} and QM9~\cite{ramakrishnan2014quantum} average around 18 nodes per graph, PTC~\cite{toivonen2003statistical} around 14, and IMDB-BINARY and IMDB-MULTI~\cite{yanardag2015deep} between 13 and 20. Accordingly, we constrain our generated graphs to a comparable range of 5–30 nodes.

Beyond mirroring real-world graph scales, our datasets are explicitly designed to evaluate whether GNNs trained on smaller graphs can generalize to larger ones.
For each property, we identify a \emph{base graph size}, the largest size for which positive instances can still be exhaustively generated within reasonable compute. 
Table~\ref{tab:relational_properties_natural} shows the base size of each property. We then synthesize additional larger graphs for each property. This controlled variation enables us to assess whether a model can transfer its reasoning to larger graphs. To further increase the difficulty, we introduce an alternative dataset design in which each positive graph is paired with a structurally similar negative example that differs by only one or two edges. This setup challenges the model to rely on deeper structural understanding rather than superficial pattern matching.

To this end, we introduce two dataset families: \textbf{GraphRandom} and \textbf{GraphPerturb}. The remainder of this section describes their construction in detail.

\vspace{-3ex}
\subsection{GraphRandom Dataset Family}
To increase graph size diversity, we construct the GraphRandom dataset family by extending each property’s base graph size to include graphs with sizes ranging from \( \text{base size} + 1 \) to \( \text{base size} + 10 \). For each size, we systematically generate graphs using the Alloy Analyzer. As shown in Figure~\ref{fig:random}, we execute the Alloy command for graph sizes \( |V| = \text{base size} + 1 \) to \( |V| = \text{base size} + 10 \), and interpret the resulting SAT solutions as adjacency matrices representing graphs of varying sizes. For example, with a base size of 5 for the \texttt{reflexivity} property, we generated datasets for \( |V| = 5 \) to \( |V| = 15 \), resulting in 11 datasets. 

Across 16 properties, this process produces a total of 176 datasets, collectively referred to as the GraphRandom dataset family.
For datasets with \( |V| \) ranging from \( \text{base size} + 1 \) to \( \text{base size} + 10 \), the number of positive samples was excessively large. As a result, we randomly selected 5000 positive samples and matched them with 5000 negative samples for each graph size, creating balanced datasets with 10,000 samples each.

\vspace{-4ex}
\subsection{GraphPerturb Dataset Family}

\begin{figure*}[t]
    \centering
    \includegraphics[width=0.85\textwidth]{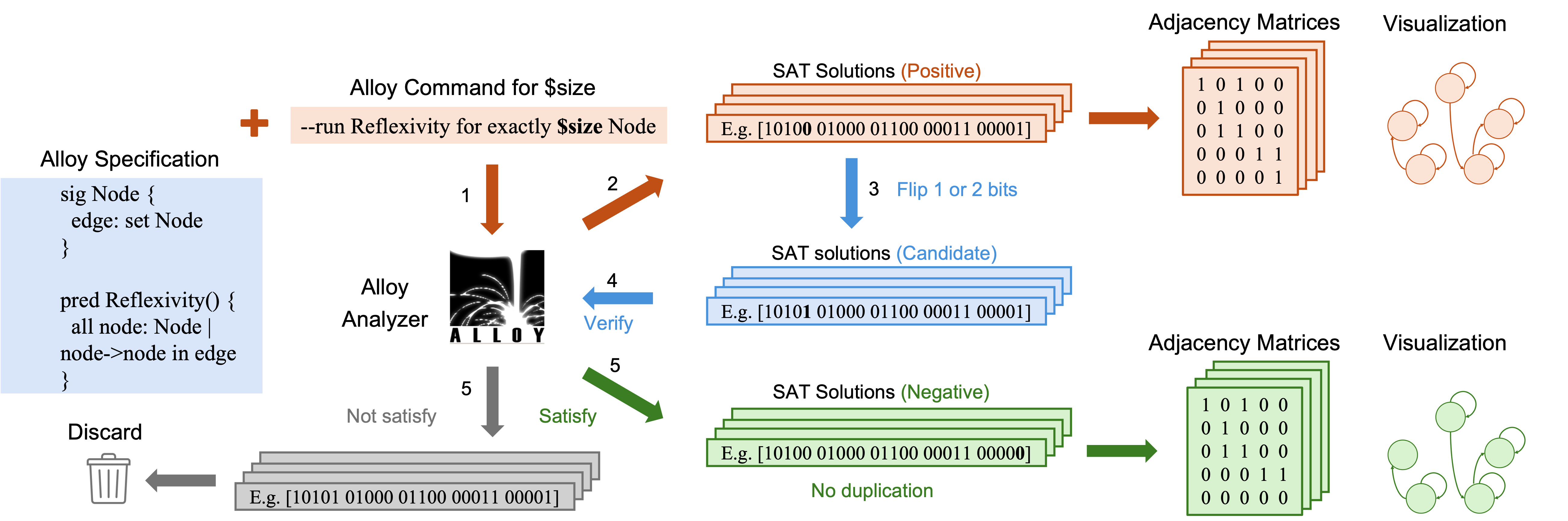} 
    \vspace{-3ex}
    \caption{GraphPerturb dataset family generation; take graph size equal to 5 as an example. 
    } 
    \vspace{-3ex}
    \label{fig:perturb} 
\end{figure*}

\noindent\textbf{Overview.}
In the GraphRandom datasets, all samples, except for positive samples with $|V|$ equal to the base size, are randomly generated. This random sample selection often creates significant structural differences between positive and negative graphs. However, tasks involving graphs with highly similar structures (e.g., differing by only one or two edges) but completely different labels for a given property are often more challenging for GNNs. To stress-test GNNs under such conditions, we developed the GraphPerturb dataset family.

\begin{algorithm}[t]
    \begin{algorithmic}[1]
 \STATE \textbf{Input:} $property$: dataset property \par  
\hspace*{3em}$M$: number of positive/negative samples \\
\hspace*{3em}$max\_fbits$: maximum number of bits to flip  \\
\hspace*{3em}$size$: graph size  
\STATE \textbf{Output:} $PSet$: positive SAT solutions \\
\hspace*{3.5em}$NSet$: negative SAT solutions  

\STATE $PSet = \{\}, \; NSet = \{\}$

\WHILE {$PSet.size() < M$}
    \STATE $s \gets alloy.generate(property)$ \hfill // generate pos sample
    \STATE $PSet.add(s)$
\ENDWHILE

\FOR{each $s \in PSet$}
    \STATE $neg\_generated \gets \texttt{false}$
    \FOR{$f = 1$ to $max\_fbits$}
        \IF{$neg\_generated$} \STATE \textbf{break} \ENDIF
        \STATE $len \gets size \cdot size$ \hfill // SAT solution length
        \STATE $N \gets C_{len}^{f}$ \hfill // number of flip combinations
        \FOR{$i = 1$ to $N$}
            \STATE $t \gets rand\_flip\_bits(f, s)$
            \IF{$alloy.verify(t, !property)$}
                \IF{$NSet.add(t)$}
                    \STATE {// add unique neg sample}  
                    \STATE neg\_generated $\gets$ \texttt{true}
                    \STATE \textbf{break}
                \ENDIF
            \ENDIF
        \ENDFOR
    \ENDFOR
\ENDFOR

    \end{algorithmic}
    \caption{Generating each dataset in the GraphPerturb family}
    \label{alg:perturb}
    \end{algorithm}

The key distinction of GraphPerturb lies in its generation of negative samples. Instead of random generation, we ensure that each positive graph has a structurally similar negative counterpart. As shown in Figure~\ref{fig:perturb}, we use the Alloy Analyzer to generate positive samples satisfying the property. For each positive sample, a negative sample is created by flipping one or two bits in the corresponding SAT solution and verifying with the Alloy Analyzer that the modified graph qualifies as a valid negative sample.

\noindent\textbf{Algorithm.}
The algorithm for generating each dataset in the GraphPerturb family is detailed in Algorithm~\ref{alg:perturb}. It takes four inputs: the property defining the dataset (\( property \)), the number of positive samples (\( M \)), the maximum number of bits to flip (\( max\_fbits \)), and graph size (\( size \)), which is the number of graph nodes. The outputs are the sets of positive and negative SAT solutions (\( PSet \) and \( NSet \)).

The algorithm begins by generating positive samples (\( PSet \)). For each positive sample \( s \), it attempts to create a structurally similar negative sample by flipping one bit and checking if the modified solution \( t \) satisfies the negative property using the Alloy Analyzer. If unsuccessful, the number of flipped bits is incrementally increased up to \( max\_fbits \). In our GraphPerturb dataset family generation, \( max\_fbits = 2 \) is sufficient, as negative samples already dominate the state space. This ensures that each positive sample in our dataset has at least one structurally similar negative counterpart differing by one or two bits (i.e., graph edges).

For each number of flipped bits, the algorithm allows \( C_{len}^f \) attempts to perturb a positive sample into a valid negative sample, where \( C_{len}^f \) represents all possible combinations of \( f \) flipped bits. Each random flip produces a candidate \( t \), which is verified as a valid negative sample. If valid, \( t \) is added to \( NSet \), and the process moves to the next positive sample. To avoid duplication, all samples are stored in a set.

\noindent\textbf{Statistics.}
In summary, the GraphPerturb dataset family consists of \( 11 \times 16 = 176 \) datasets. For \( |V| = \text{base size} \), we run the Alloy command to obtain all positive samples and generate an equal number of perturbed negative samples. For \( |V| \) ranging from \( \text{base size} + 1 \) to \( \text{base size} + 10 \), we run the Alloy command, randomly select 5000 positive samples, and use Algorithm~\ref{alg:perturb} to generate an equal number of perturbed negative samples.

%% file: 5_metrics.tex
\vspace{-2ex}
\section{GNN Expressiveness Evaluation Framework}
We propose an evaluation framework for evaluating the expressive power of GNNs in terms of three key aspects (i.e., generalizability, sensitivity, and robustness) using our generated datasets and proposed evaluation metrics. 
\vspace{-2ex}
\subsection{Training and Testing Sets}
To facilitate this evaluation, we split our generated datasets into training and testing sets. Each GNN is trained and validated on a dataset where the graph size \( |V| = \text{base size} \), as the dataset contains all possible positive samples of that size, allowing for sufficient learning. The trained model is then tested on the remaining datasets with varying graph sizes. For each property, we define four dataset categories:

\begin{itemize} [nosep, left=0pt]
    \item \textbf{\texttt{GraphRandom-Train}}: GraphRandom dataset with graph size equal to the base size.
    \item \textbf{\texttt{GraphRandom-Test}}: GraphRandom datasets with graph sizes ranging from \( \text{base size} + 1 \) to \( \text{base size} + 10 \).
    \item \textbf{\texttt{GraphPerturb-Train}}: GraphPerturb dataset with graph size equal to the base size.
    \item \textbf{\texttt{GraphPerturb-Test}}: GraphPerturb datasets with graph sizes ranging from \( \text{base size} + 1 \) to \( \text{base size} + 10 \).
\end{itemize}

\vspace{-2ex}
\subsection{Three Key Aspects}
We assess each aspect of expressiveness using different combinations of our generated datasets:  \\
\noindent\textbf{Generalizability.}
We define a GNN as generalizable if it maintains stable accuracy across varying graph sizes. To evaluate this property, we train models on \texttt{GraphRandom-Train} and test them on \texttt{GraphRandom-Test}, assessing their ability to generalize to graphs of different sizes and structural variations.

\noindent\textbf{Sensitivity.}
Sensitivity measures whether a GNN can distinguish between highly similar graphs with different labels. We assess this by training the model on \texttt{GraphPerturb-Train} and testing on \texttt{GraphPerturb-Test}, where small structural differences determine classification outcomes.

\noindent\textbf{Robustness.}
A robust GNN should differentiate between highly similar, unseen graph structures, especially when trained on simpler graphs. We evaluate robustness by training on \texttt{GraphRandom-Train} and testing on \texttt{GraphPerturb-Test} to determine if the model can generalize to more challenging graph variations.

\vspace{-2ex}
\subsection{Evaluation Metrics}
\noindent\textbf{Unified Score.}
To evaluate three aspects, generalizability, sensitivity, and robustness, we propose a unified quantitative metric:
\vspace{-2ex}
\begin{equation}
\text{U\_score} = \sum_{j=1}^{10}\frac{\text{accuracy}_j \times \text{gsize}_j}{\sum_{j=1}^{10}\text{gsize}_j}
\vspace{-1.5ex}
\end{equation}
where \( j \) corresponds to the index of testing datasets (\( j = 1, 2, \dots, 10 \)). Here, \( \text{gsize}_j \) is the graph size of the \( j \)-th testing dataset, and \( \text{accuracy}_j \) is the model's accuracy on that dataset. 

Our unified score, ranging from 0.0 to 1.0, reflects a model’s effectiveness across diverse datasets. Higher scores indicate better accuracy across varying graph sizes. Since larger graphs typically exhibit greater structural complexity and pose more learning challenges, we weight each accuracy score by its graph size. This ensures that performance on more complex graphs contributes proportionally more to the overall score. Normalizing by total graph size makes the score independent of individual dataset sizes, enabling standardized comparisons across models or experiments. In addition, the metric is uniformly applicable to generalizability, sensitivity, and robustness, ensuring consistent performance evaluation. 

\noindent\textbf{Relative Score.}
To access the relative capability of a GNN model comparing its peers, we define the \emph{Relative Score} ($R\_score$) of a model $i$ for a graph property $p$ on an aspect $a$  as:
\vspace{-1.5ex}
\begin{equation}
  R\_score_{a,p,i} = \frac{U\_score_{a, p, i}}{mean_{a,p}}
  \vspace{-1.5ex}
\end{equation}
where
\vspace{-1ex}
\begin{equation}
  mean_{a,p} = \frac{\sum_i U\_score_{a, p, i}}{N_G}
  \vspace{-1ex}
\end{equation}
Here, $N_G$ refers to the total number of GNN models under comparison; $U\_score_{a, p, i}$ represents the unified score of a model $i$ for a property $p$ on an aspect $a$; and $mean_{a,p}$ is the average score across all GNN models for property $p$ on an aspect $a$.

To investigate the relative performance of a GNN model $i$ on an aspect $a$, we define $R\_score_{a, i}$:
\vspace{-1.5ex}
\begin{equation}
  R\_score_{a, i} = \frac{1}{N_p}\sum_p R\_score_{a, p, i}
  \vspace{-1.5ex}
\end{equation}
, where $N_p$ is the total number of properties.

To evaluate the relative performance of a model $i$ on a property $p$, we define $R\_score_{p, i}$:
\vspace{-1.5ex}
\begin{equation}
  R\_score_{p, i} = \frac{1}{N_a} \sum_a R\_score_{a, p, i}
  \vspace{-1.5ex}
\end{equation}
, where $N_a$ is the total number of aspects.

To obtain a view of the overall performance of a model $i$, we define $R\_score_i$:
\vspace{-1.5ex}
\begin{equation}
  R\_score_{i} = \frac{1}{N_p} \frac{1}{N_a}\sum_p \sum_a R\_score_{a, p, i}
  \vspace{-1.5ex}
\end{equation} 

The Relative Score is a normalized metric that compares a GNN model’s performance to the average across properties and aspects. A score of 1.0 indicates average performance, while values above or below 1.0 signify overperformance or underperformance to its peers, respectively. It enables direct comparison and ranking of different GNN models. Moreover, by aggregating these scores across properties and aspects, we can identify specific strengths and weaknesses of each GNN model, thereby guiding improvements in GNN expressiveness.

%% file: 6_study.tex
\vspace{-2ex}
\section{Study: Global Pooling for GNN Expressiveness}
Using our proposed evaluation framework, we conduct the first  comprehensive study on how different global pooling methods affect GNN expressiveness.

\noindent\textbf{Subjects.}
To the best of our knowledge, our study includes all state-of-the-art, open-source global pooling methods across four main categories, totaling nine subjects. 
\begin{itemize}[nosep, left=0pt]
    \item \emph{Basic approaches}: mean pooling and sum pooling~\cite{xu2018powerful}.
    \item  \emph{Neural network approaches}: DeepSets~\cite{navarin2019universal, buterez2022graph} and Set2Set~\cite{vinyals2015order}. 
    \item \emph{Attention approaches}: soft attention pooling~\cite{li2015gated, li2019graph}, Set Transformer~\cite{lee2019set, buterez2022graph}, and Graph Multiset Transformer (GMT)\cite{baek2021accurate}. 
    \item \emph{Second-order approaches}: Bilinear Mapping Second-Order Pooling (SoPool-BiMap) and Attentional Second-Order Pooling (SoPool-Attentional)\cite{wang2020second}.
\end{itemize}

\begin{figure*}[t!]
    \centering
    \begin{minipage}[t]{0.13\textwidth}
        \centering
        \vspace{1.3ex}
        \includegraphics[width=\textwidth]{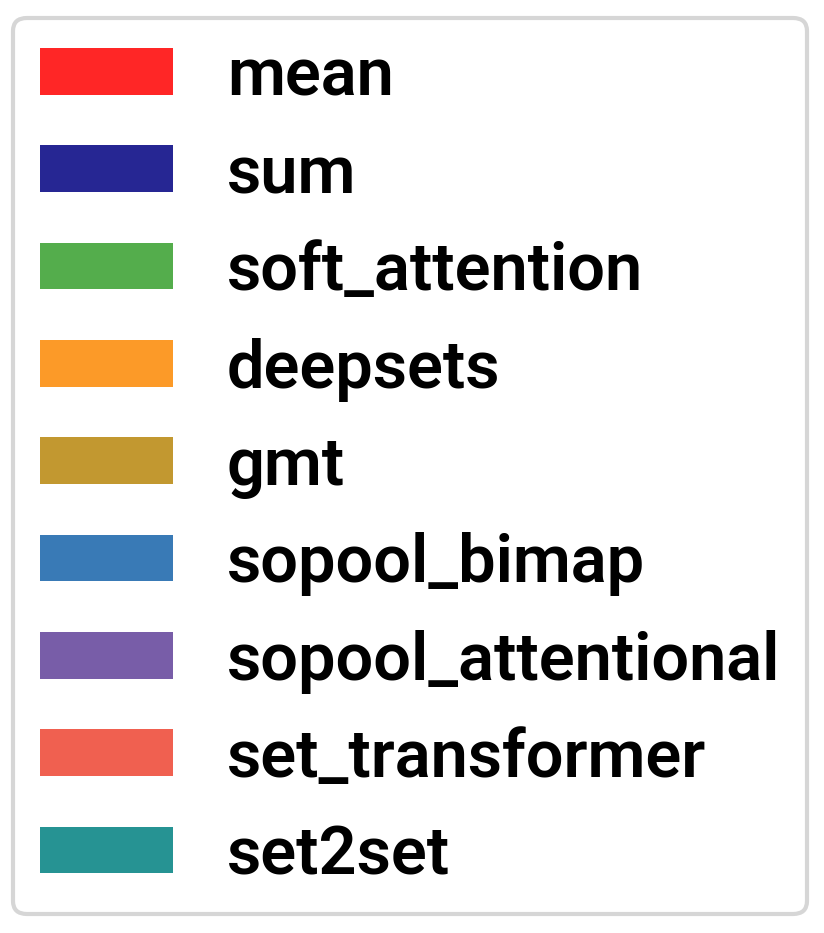} 
        \vfill
        \label{fig:bar_legend}
    \end{minipage}%
    \hspace{-0.01\textwidth} 
    \begin{minipage}[t]{0.85\textwidth}
        \centering
        \begin{subfigure}{\textwidth}
            \centering
            \includegraphics[width=0.95\textwidth]{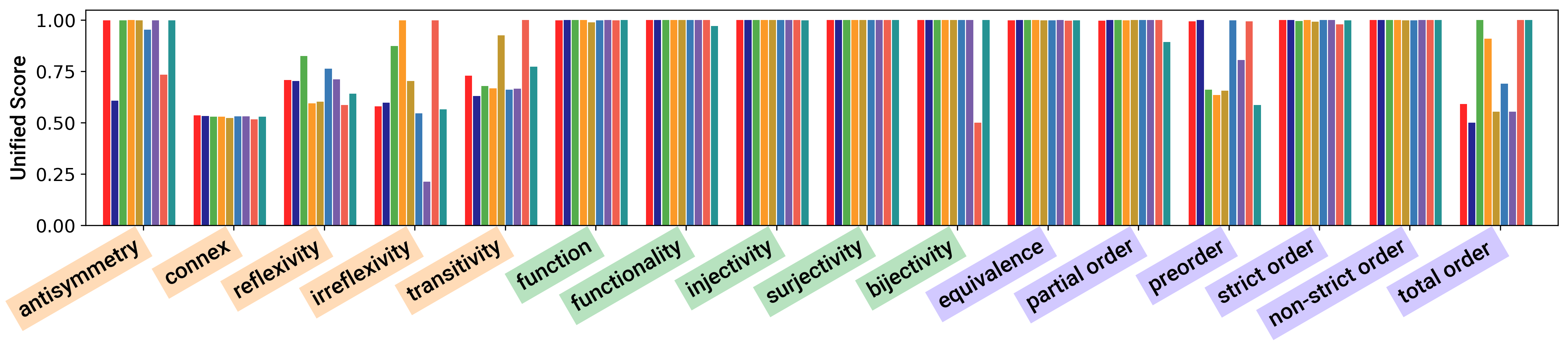}
            \vspace{-3ex}
            \caption{Generalizability}
            \vspace{-1ex}
            \label{fig:generalizability_bar}
        \end{subfigure}

        \vspace{1ex}
        \begin{subfigure}{\textwidth}
            \centering
            \includegraphics[width=0.95\textwidth]{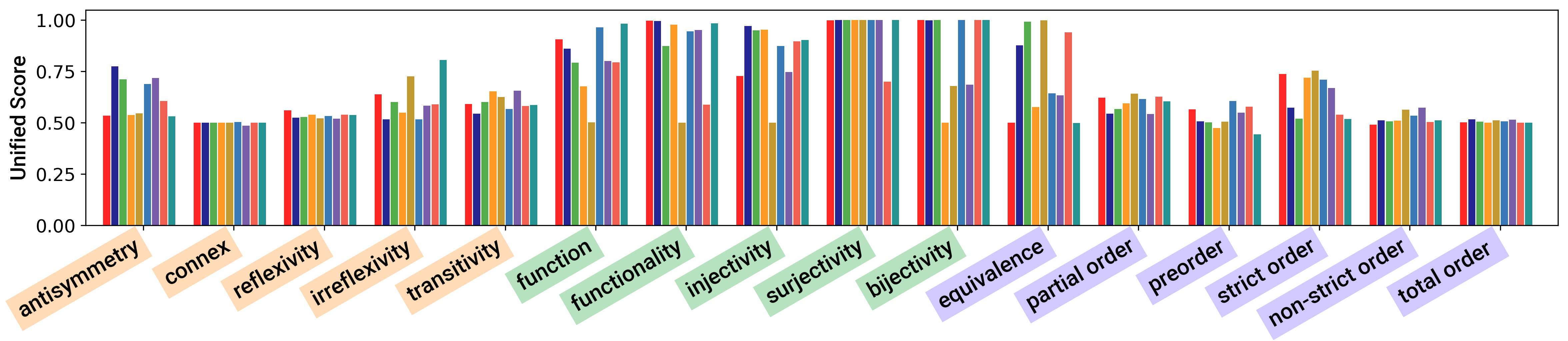}
            \vspace{-3ex}
            \caption{Sensitivity}
            \vspace{-1ex}
            \label{fig:sensitivity_bar}
        \end{subfigure}

        \vspace{1ex}
        \begin{subfigure}{\textwidth}
            \centering
            \includegraphics[width=0.95\textwidth]{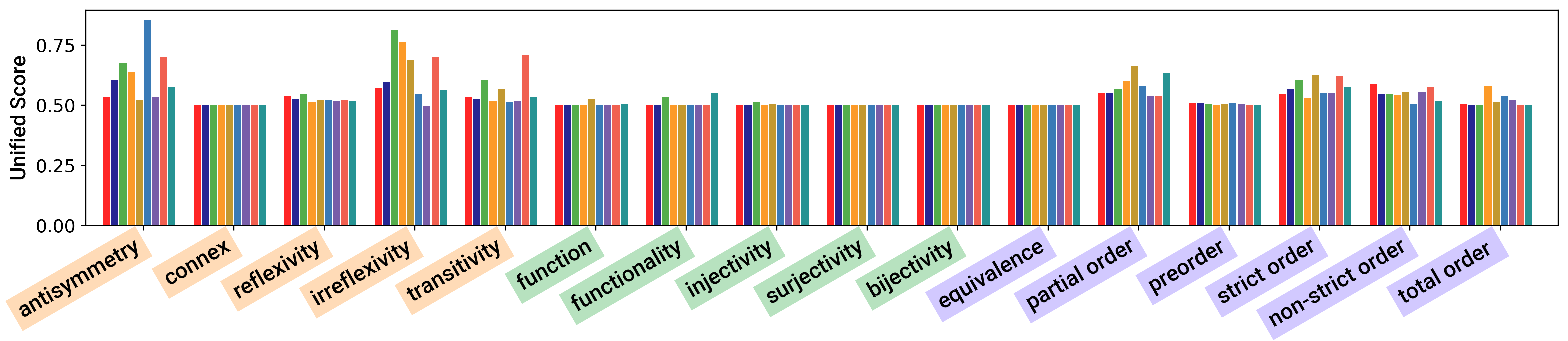}
            \vspace{-3ex}
            \caption{Robustness}
            \vspace{-3ex}\label{fig:robustness_bar}
        \end{subfigure}
    \end{minipage}

    \caption{Unified scores of GNNs with different pooling methods across three aspects on 16 properties. }
    \vspace{-3ex}\label{fig:3_aspects_3}
\end{figure*}

\noindent\textbf{Node Embedding Generation.}
To isolate the effect of global pooling, we use a fixed GNN architecture to generate node embeddings while varying only the pooling method. We adopt Identity-aware Graph Neural Networks (ID-GNN) \cite{you2021identity} with GIN as the neighborhood aggregation and update function, as this combination provides strong expressiveness and effectively captures local graph structure. GIN employs an injective aggregation scheme \cite{xu2018powerful}, assigning identical embeddings to two nodes only when their subtree structures and features match. ID-GNN further enhances expressiveness by incorporating node-level cycle counts.

Among the variants, ID-GNN-Fast achieves the best performance on graph classification tasks \cite{you2021identity} and is therefore used in our study. We configure the model with five GIN layers and concatenate the outputs of all layers to form the final node embeddings. Training uses the AdamW optimizer \cite{loshchilov2017decoupled} with binary cross-entropy loss, a batch size of 64, a learning rate of 0.001, and 20 epochs. For each dataset, 95\% of GraphRandom-Train or GraphPerturb-Train is used for training and the remaining 5\% for validation. Model selection is based on validation performance, and final results are reported on the held-out GraphRandom-Test and GraphPerturb-Test sets.

%% file: 7_exp.tex
\vspace{-3ex}
\subsection{Impact of Global Pooling}
This section evaluates the performance of global pooling methods in GNNs across three key aspects: generalizability, sensitivity, and robustness. 
Figure~\ref{fig:3_aspects_3} summarizes the unified score of different global pooling methods across three aspects and 16 properties. A detailed analysis of each aspect follows.

\noindent\textbf{Generalizability.}
\ul{Most global pooling methods achieve high accuracy in generalizability.} This can be attributed to the nature of the datasets, as both the training and testing datasets consist of graphs with less structural similarity, making the task relatively easier.
However, performance varies across different property categories. 
\ul{Function-related properties consistently yield near-perfect scores across methods}, showing that global pooling can effectively learn node mappings and function-like structures even when graph size changes. \ul{For basic properties, performance is mixed.} While \texttt{antisymmetry} is learned well (e.g., DeepSets and soft attention perform strongly), properties like \texttt{transitivity}, and \texttt{irreflexivity} show notable variability, suggesting that global pooling struggles with certain relational constraints. \ul{For most combined properties, models surprisingly perform well}, likely because structural dependencies among component properties reinforce the signal (e.g., \texttt{partial order} and \texttt{non-strict order} benefit from \texttt{transitivity} and \texttt{antisymmetry}). In addition, attention-based methods often lead in the performance on combined properties.

\noindent\textbf{Sensitivity.}
\ul{Compared to generalizability, sensitivity reveals a noticeable drop in performance across all three property categories}, underscoring the difficulty of detecting fine-grained structural differences in graphs. 
Among the three categories, \ul{function-related properties remain the most tractable}: most pooling methods achieve relatively high scores. For example, on the \texttt{surjectivity} property, nearly all pooling methods attain near-perfect performance (close to 1.0), indicating their ability to detect subtle violations in node-to-node mappings.
In contrast, \ul{basic properties display considerable variation}. Some properties can be partially captured; for instance, \texttt{antisymmetry} reaches a peak accuracy of approximately 0.77 with sum pooling, while \texttt{transitivity} tops out around 0.65. However, other properties remain elusive: \texttt{connex}, in particular, consistently hovers around 0.50 across all pooling methods, suggesting random-level performance and an inability to capture pairwise comparability under minor perturbations.
\ul{Combined properties present the greatest challenge.} Many pooling methods perform no better than chance on tasks such as \texttt{total order} and \texttt{non-strict order}—all methods score around 0.5 for the former, and the best score for the latter is only 0.57. These results suggest that models struggle to retain the nuanced logical combinations that define these complex relations when only minor graph changes are introduced.
Nevertheless, there are notable exceptions: attention-based pooling mechanisms demonstrate strong sensitivity to structural patterns in some combined properties. For instance, \texttt{equivalence} relations are captured with high fidelity by soft attention and GMT, both achieving scores close to 1.0.

\noindent\textbf{Robustness.} \ul{Robustness is the most challenging aspect} for global pooling methods, with their unified score dropping significantly across all property types, compare to both generalizability and sensitivity. Pooling methods that perform well on function-related properties in both generalizability and sensitivity see their robustness decline to around 0.5. Similarly, pooling methods exhibit poor robustness on combined properties and basic properties, as subtle structural variations in unseen, highly similar graphs overwhelm their ability to differentiate effectively. Nonetheless, \ul{a few exceptions highlight the potential of specialized architectures}: certain basic properties, such as \texttt{antisymmetry} and \texttt{irreflexivity}, are captured relatively well by tailored pooling strategies. For instance, SoPool-BiMap achieves a robustness score of 0.85 on \texttt{antisymmetry}, and soft attention reaches 0.81 on \texttt{irreflexivity}, demonstrating that architectural choices can make a significant difference in robustness for select properties.


\begin{table*}[t!]
\centering
\scalebox{0.83}{
\begin{tabular}{||c||c|c|c|c|c|c|c|c|c||}
\hline
\textbf{} & \textbf{mean} & \textbf{sum} & \textbf{soft\_attention} & \textbf{deepsets} & \textbf{gmt} & \textbf{set\_transformer} & \textbf{set2set} & \textbf{sopool\_bimap} & \textbf{sopool\_attentional} \\
\hline
\textbf{Generalization} & 1.003 & 0.962 & \textbf{1.044} & 1.025 & 0.990 & 1.033 & 0.992 & 1.004 & 0.946 \\
\hline
\textbf{Sensitivity} & 1.008 & 1.030 & 1.027 & 0.958 & 0.959 & 0.983 & 1.004 & \textbf{1.037} & 0.994 \\
\hline
\textbf{Robustness} & 0.978 & 0.983 & \textbf{1.033} & 1.009 & 1.012 & 1.030 & 0.989 & 1.002 & 0.964 \\
\hline
\end{tabular}
}
\caption{Relative pooling scores on different aspects.}
\vspace{-6ex}
\label{tab:relative_ai}
\end{table*}

\begin{table*}[t!]
\centering
\scalebox{0.82}{
\begin{tabular}{||c||c|c|c|c|c|c|c|c|c||}
\hline
\textbf{} & \textbf{mean} & \textbf{sum} & \textbf{soft\_attention} & \textbf{deepsets} & \textbf{gmt} & \textbf{set\_transformer} & \textbf{set2set} & \textbf{sopool\_bimap} & \textbf{sopool\_attentional} \\
\hline
\rowcolor{lightorange}
\textbf{Antisymmetry} & 0.929 & 0.953 & 1.098 & 0.986 & 0.930 & 0.961 & 0.950 & \textbf{1.165} & 1.027 \\
\hline
\rowcolor{lightorange}
\textbf{Connex} & \textbf{1.005} & 1.004 & 1.001 & 1.001 & 0.997 & 0.993 & 1.001 & \textbf{1.005} & 0.993 \\
\hline
\rowcolor{lightorange}
\textbf{Reflexivity} & 1.036 & 1.005 & \textbf{1.081} & 0.954 & 0.951 & 0.956 & 0.979 & 1.037 & 1.001 \\
\hline
\rowcolor{lightorange}
\textbf{Irreflexivity} & 0.932 & 0.887 & 1.183 & \textbf{1.189} & 1.100 & 1.180 & 1.012 & 0.835 & 0.681 \\
\hline
\rowcolor{lightorange}
\textbf{Transitivity} & 0.973 & 0.898 & 0.997 & 0.969 & 1.097 & \textbf{1.190} & 0.989 & 0.916 & 0.971 \\
\hline

\rowcolor{lightgreen}
\textbf{Function} & 1.038 & 1.020 & 0.993 & 0.944 & 0.885 & 0.992 & \textbf{1.071} & 1.062 & 0.996 \\
\hline
\rowcolor{lightgreen}
\textbf{Functionality} & 1.045 & 1.044 & 1.018 & 1.037 & 0.855 & 0.888 & \textbf{1.063} & 1.024 & 1.027 \\
\hline
\rowcolor{lightgreen}
\textbf{Injectivity} & 0.955 & \textbf{1.052} & \textbf{1.052} & 1.045 & 0.868 & 1.023 & 1.026 & 1.014 & 0.963 \\
\hline
\rowcolor{lightgreen}
\textbf{Surjectivity} & \textbf{1.011} & \textbf{1.011} & \textbf{1.011} & \textbf{1.011} & \textbf{1.011} & 0.908 & \textbf{1.011} & \textbf{1.011} & \textbf{1.011} \\
\hline
\rowcolor{lightgreen}
\textbf{Bijectivity} & \textbf{1.068} & 1.067 & \textbf{1.068} & 0.877 & 0.945 & 0.891 & \textbf{1.068} & \textbf{1.068} & 0.948 \\
\hline

\rowcolor{lightpurple}
\textbf{Equivalence} & 0.892 & 1.062 & 1.114 & 0.927 & \textbf{1.117} & 1.090 & 0.892 & 0.956 & 0.952 \\
\hline
\rowcolor{lightpurple}
\textbf{Partial order} & 1.002 & 0.959 & 0.981 & 1.015 & \textbf{1.077} & 0.997 & 1.004 & 1.016 & 0.949 \\
\hline
\rowcolor{lightpurple}
\textbf{Preorder} & 1.100 & 1.066 & 0.922 & 0.893 & 0.921 & 1.106 & 0.853 & \textbf{1.130} & 1.010 \\
\hline
\rowcolor{lightpurple}
\textbf{Strict order} & 1.037 & 0.964 & 0.955 & 1.017 & \textbf{1.089} & 0.970 & 0.939 & 1.026 & 1.003 \\
\hline
\rowcolor{lightpurple}
\textbf{Non-strict order} & 1.003 & 0.992 & 0.988 & 0.989 & 1.031 & 1.005 & 0.974 & 0.981 & \textbf{1.036} \\
\hline
\rowcolor{lightpurple}
\textbf{Total order} & 0.915 & 0.883 & 1.096 & \textbf{1.103} & 0.913 & 1.092 & 1.093 & 0.986 & 0.920 \\
\hline
\end{tabular}
}
\caption{Relative scores on different properties. }
\vspace{-6ex}

\label{tab:relative_pi}
\end{table*}

\begin{table*}[t!]
\centering
\scalebox{0.89}{
\begin{tabular}{||c||c|c|c|c|c|c|c|c|c||}
\hline
\textbf{} & \textbf{mean} & \textbf{sum} & \textbf{soft\_attention} & \textbf{deepsets} & \textbf{gmt} & \textbf{set\_transformer} & \textbf{set2set} & \textbf{sopool\_bimap} & \textbf{sopool\_attentional} \\
\hline
\textbf{Overall} & 0.996 & 0.992 & \textbf{1.035} & 0.997 & 0.987 & 1.015 & 0.995 & 1.014 & 0.968 \\
\hline
\end{tabular}
}
\caption{Overall relative pooling scores.}
\vspace{-7ex}
\label{tab:relative_i}
\end{table*}

\vspace{-2ex}
\subsection{Global Pooling Comparisons}

To compare global pooling methods, we evaluate their relative scores across aspects, properties, and overall.

\noindent\textbf{Relative Performance on Different Aspects.} Table~\ref{tab:relative_ai} shows the relative performance of global pooling methods on three aspects. \ul{In terms of generalizability and robustness, soft attention pooling achieves the best performance}, with relative scores of 1.044 and 1.033, respectively. \ul{Regarding sensitivity, SoPool-BiMap achieves the best performance}. In addition, most attention-based global pooling approaches achieve relatively higher scores for generalizability and robustness but show a slight drop in sensitivity. The opposite pattern is observed with SoPool-BiMap.

\noindent\textbf{Relative Performance on Different Properties} 
Table~\ref{tab:relative_pi} illustrates the relative performance of global pooling methods on 16 properties. Overall, \ul{no single pooling method excels across all properties}. Instead, different methods perform best under specific properties, highlighting trade-offs in global pooling design.
Despite their simplicity, mean and sum poolings often achieve scores near 1.0, suggesting simple pooling methods can perform comparably to more complex ones.
Among neural network-based methods, Set2Set consistently ranks among the top for most function-related properties (e.g., \texttt{bijectivity}, \texttt{function}, \texttt{functionality}, \texttt{surjectivity}), while deepsets occasionally leads, particularly for \texttt{irreflexivity}, \texttt{total order} and \texttt{surjectivity}. 
For attention-based methods, soft attention pooling ties for the best performance on \texttt{bijectivity}, \texttt{injectivity}, and \texttt{surjectivity}, while leading on \texttt{reflexivity} (1.081). Meanwhile, Set Transformer excels in \texttt{transitivity} (1.190), and Graph Multiset Transformer (GMT) performs generally well on combined properties like \texttt{equivalence} (1.117), \texttt{partial order} (1.077) and \texttt{strict order} (1.089).
Among second-order pooling methods, SoPool-BiMap leads in five properties, including \texttt{antisymmetry} (1.165) and \texttt{preorder} (1.130). In contrast, Attentional Second-Order Pooling shows mixed results, excelling in \texttt{non-strict order} (1.036) but significantly underperforming in \texttt{irreflexivity} (0.681), indicating that SoPool-BiMap is generally more effective.
Some properties see pooling methods performing nearly identically. \texttt{surjectivity} shows an almost universal tie (1.011), while \texttt{bijectivity} also trends toward similar performance.

\noindent\textbf{Overall Relative Performance} 
Table~\ref{tab:relative_i} shows the overall relative performance of the nine global pooling methods evaluated. Overall, \ul{the two attention-based methods (soft attention pooling and Set Transformer) achieve the highest average rankings}. SoPool-BiMap performs comparably to Set Transformer, while Attentional Second-Order Pooling ranks the lowest. Basic methods (mean, sum) and neural network-based approaches (DeepSets, Set2Set) cluster just below 1.0, showing a slight gap from the top performers.

\vspace{-2ex}
\subsection{Global Pooling across Graph Sizes}

Figure~\ref{fig:line_generalizability},~\ref{fig:line_sensitivity} and ~\ref{fig:line_robustness} in the Appendix illustrate how the performance of global pooling methods changes with graph size. 
We observe that the effects of increasing graph size differ across the three aspects. \ul{In terms of generalizability, most performance remains consistently high}, often near 1.0, even as graph size increases. This suggests that global pooling methods can generalize well to larger graphs when the underlying distribution remains similar.
In contrast, \ul{sensitivity typically starts with high performance on small graphs but shows a sharp decline} as graph size grows. This steeper drop indicates that while the graphs are structurally similar, increasing complexity makes subtle distinctions harder to capture. Finally, \ul{robustness begins at a much lower performance level and tends to degrade slowly} with increasing graph size. This reflects the challenge of handling distributional shifts: pooling methods already struggle at small sizes, leaving little room for further decline.

More interestingly, performance also varies across property categories. \ul{Basic properties tend to be the most fragile as graph size increases}, likely due to their reliance on local, detail-sensitive cues—such as self-loops, edge directionality, or multi-hop transitivity—that pooling operators may struggle to preserve in larger graphs. In contrast, \ul{function-related properties generally remain more stable}, as their constraints resemble node-to-node mapping patterns that pooling methods appear better able to capture across scales. \ul{Combined properties fall in between}, with degradation rates varying: in some cases overlapping constraints seem to provide reinforcing signals, while in others small errors in one component can cascade and undermine overall performance.

\vspace{-2ex}
\section{Future Directions for Global Pooling}

Our results expose a clear gap between the capabilities of existing global pooling methods and the demands of fine-grained structural reasoning, motivating several future research directions toward more expressive and reliable GNN architectures.

\noindent\textbf{Property-Aware Adaptive Pooling.}
No single pooling method consistently performs best across all 16 properties (Table~\ref{tab:relative_pi}). Attention pooling excels in robustness and generalization, whereas bilinear second-order pooling achieves the highest sensitivity. This suggests that pooling should be adaptive rather than fixed. A promising direction is dynamic pooling that learns to select or combine multiple primitives (e.g., mean, attention, Set2Set, and second-order pooling) based on graph-level signals, enabling task- and property-specific aggregation.

\noindent\textbf{Graph-Size-Aware Architectures.}
Generalization, sensitivity, and robustness all degrade as graph size increases (Figures~\ref{fig:line_generalizability}–\ref{fig:line_robustness}). Future pooling layers could incorporate explicit size encodings, hierarchical coarsening, or scale-adaptive parameters to preserve structural information across sizes, yielding more consistent performance on large molecular, social, and communication graphs.

\noindent\textbf{Robustness-Oriented Training.}
Robustness evaluations show substantial performance drops (often exceeding 35\%,  Figure~\ref{fig:3_aspects_3}), indicating that current pooling methods lack mechanisms to ensure stability under structural noise. Incorporating robustness-driven objectives—such as adversarial perturbations, contrastive learning, or stability regularization—could significantly improve reliability, particularly in high-stakes domains.

\noindent\textbf{Unified Attention and Second-Order Pooling.}
Our results highlight a clear trade-off: attention-based pooling offers robustness and stability, while second-order methods (e.g., SoPool-BiMap) provide fine-grained sensitivity. Hybrid designs that combine both—such as attention over bilinear interactions or covariance-aware attention—may achieve expressive yet resilient pooling, mitigating the trade-offs observed in Figure~\ref{fig:3_aspects_3} and Tables~\ref{tab:relative_ai} and~\ref{tab:relative_pi}.

\noindent\textbf{Theory-Guided Expressiveness Analysis.}
Empirically, existing pooling methods fail to distinguish certain relational properties (e.g., \texttt{total order} and \texttt{connex}). Developing theoretical characterizations of pooling expressiveness—through logic-based distinguishability or subgraph counting power—would provide formal grounding to guide principled pooling design.

\noindent\textbf{Benchmark Expansion.}
Extending benchmarks to include edge attributes, temporal dynamics, and multi-relational graphs would better reflect real-world settings. These dimensions are currently missing but critical for applications such as knowledge graphs and dynamic systems. Richer benchmarks can expose new limitations and drive the development of more versatile pooling operators.

%% file: 8_conclusion.tex
\vspace{-3ex}
\section{Conclusion}

In this work, we introduce a property-driven methodology for evaluating GNN expressiveness at scale, grounded in formal specification, systematic benchmarking, and rigorous analysis. Using Alloy, we construct two dataset families—GraphRandom and GraphPerturb—comprising 352 balanced datasets across 16 fundamental graph properties. Building on these datasets, we propose a unified evaluation framework with quantitative metrics that assess expressiveness along three dimensions: generalizability, sensitivity, and robustness, and use it to conduct the first systematic study of global pooling methods.

Our results show that no single pooling design performs well across all properties and expressiveness dimensions, underscoring the need for property-aware model selection and architectural design. This perspective motivates future directions including adaptive, task-aware pooling mechanisms, scale-aware architectures that preserve structural cues in large graphs, and robustness-oriented training strategies. It also calls for theory-guided analyses of pooling expressiveness and richer benchmarks incorporating temporal, attributed, and multi-relational graphs. By integrating formal specification rigor into GNN evaluation, this work lays a foundation for graph learning systems that combine expressive power with reliability across diverse applications.


%% file: 9_appendix.tex
\appendix
\section{Performance across graph sizes}

\begin{figure*}[t]
  \centering

    \begin{subfigure}{\textwidth}
        \centering
        \includegraphics[width=0.5\textwidth]{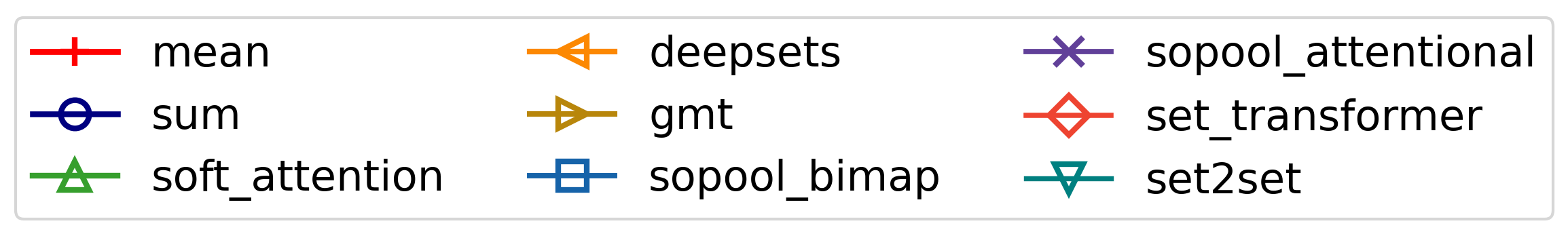}
        \label{fig:line_legend}
    \end{subfigure}

  \begin{subfigure}[b]{0.24\textwidth}
    \includegraphics[width=\textwidth]{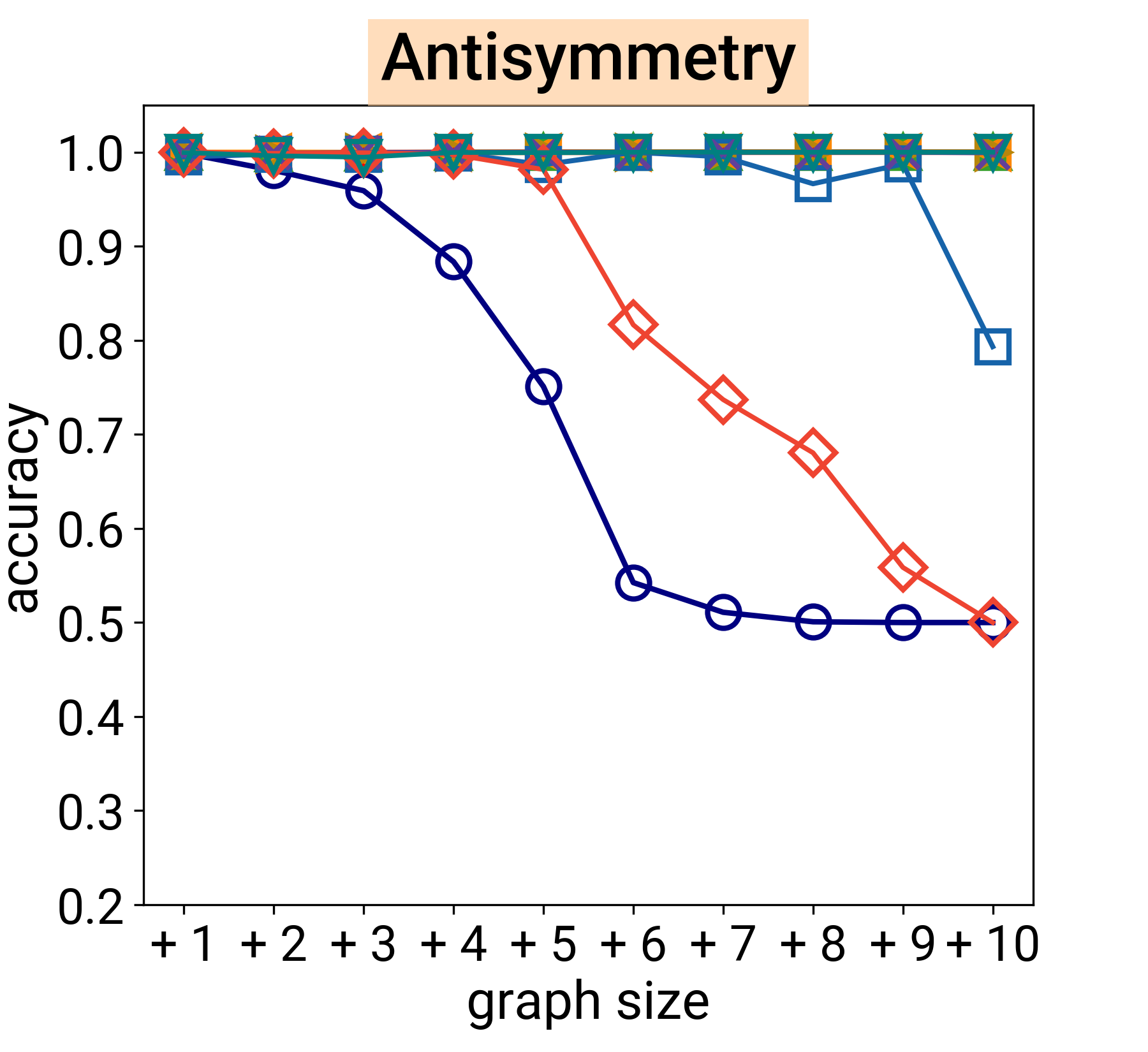}
  \end{subfigure}
    \begin{subfigure}[b]{0.24\textwidth}
    \includegraphics[width=\textwidth]{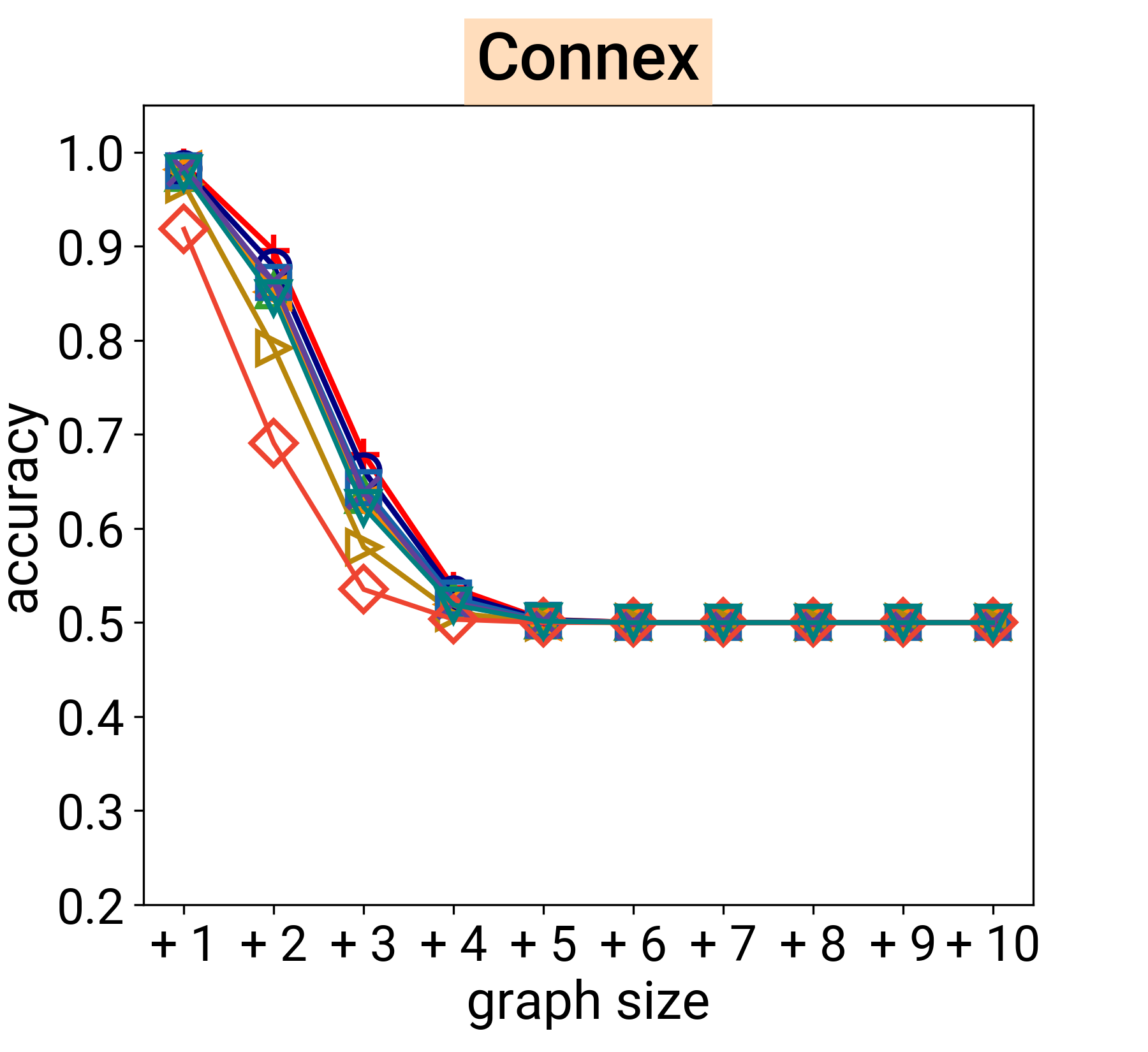}
  \end{subfigure}
    \begin{subfigure}[b]{0.24\textwidth}
    \includegraphics[width=\textwidth]{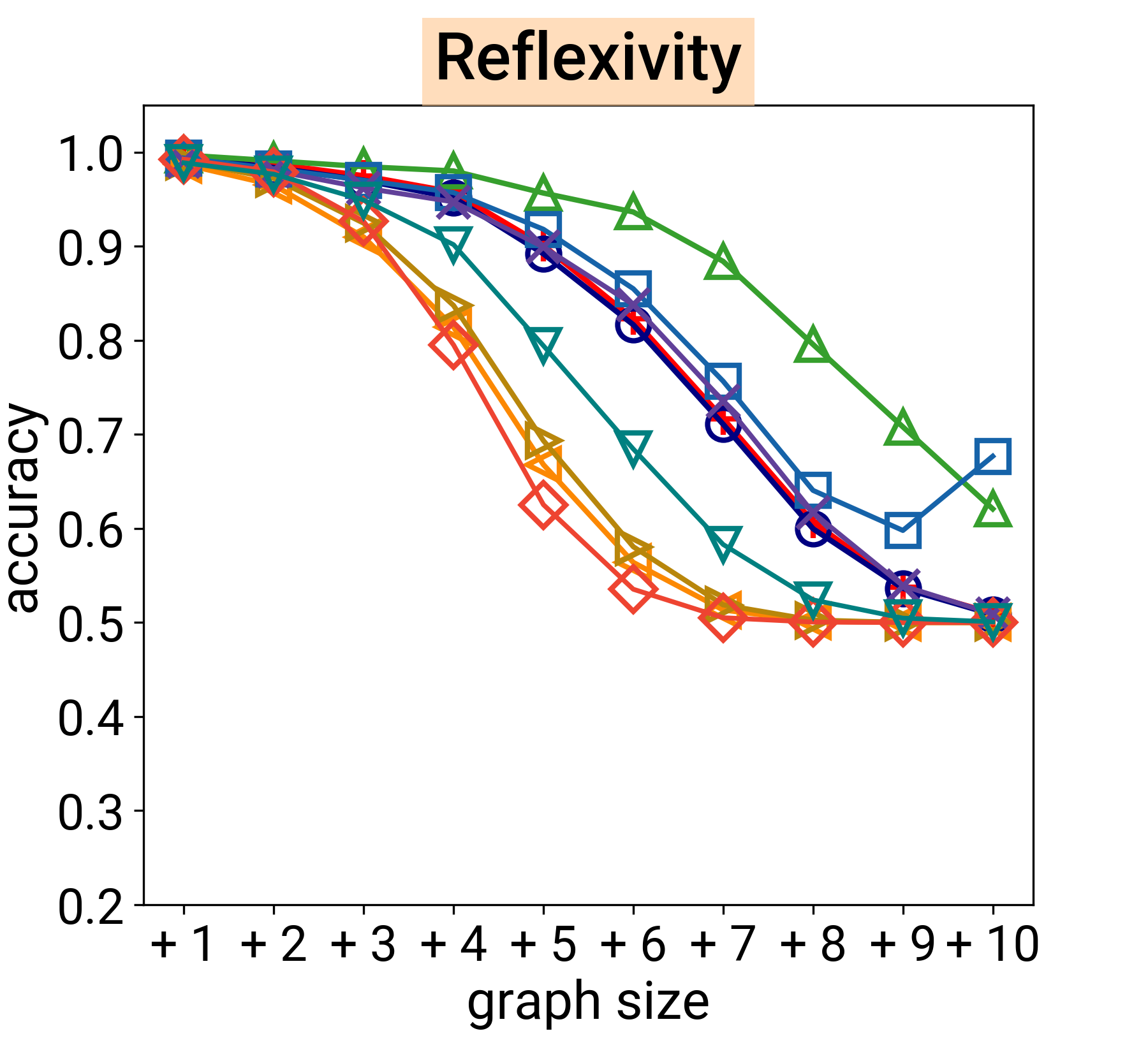}
  \end{subfigure}
    \begin{subfigure}[b]{0.24\textwidth}
    \includegraphics[width=\textwidth]{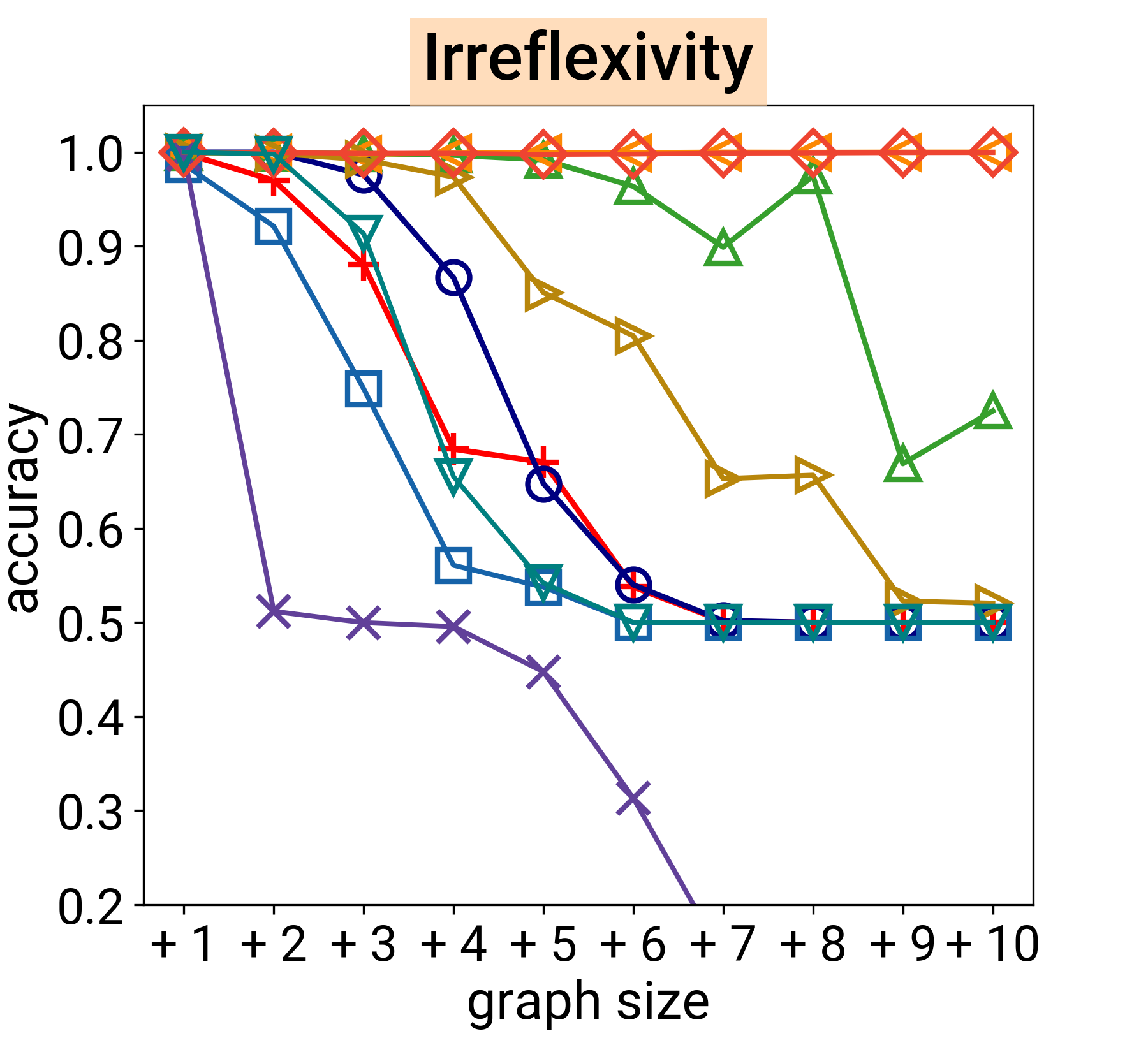}
  \end{subfigure}

  \par\smallskip
    \begin{subfigure}[b]{0.24\textwidth}
    \includegraphics[width=\textwidth]{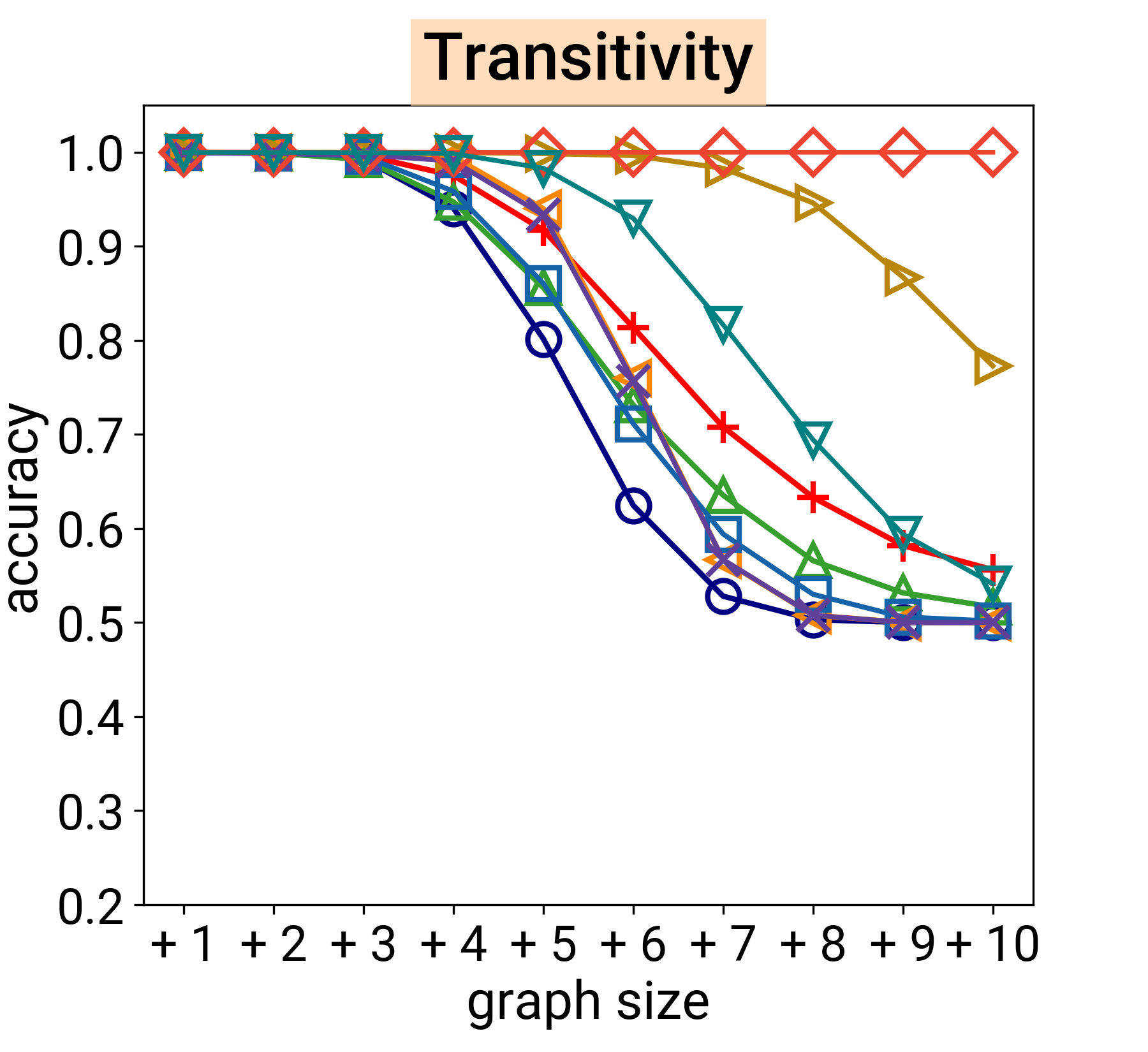}
  \end{subfigure}
    \begin{subfigure}[b]{0.24\textwidth}
    \includegraphics[width=\textwidth]{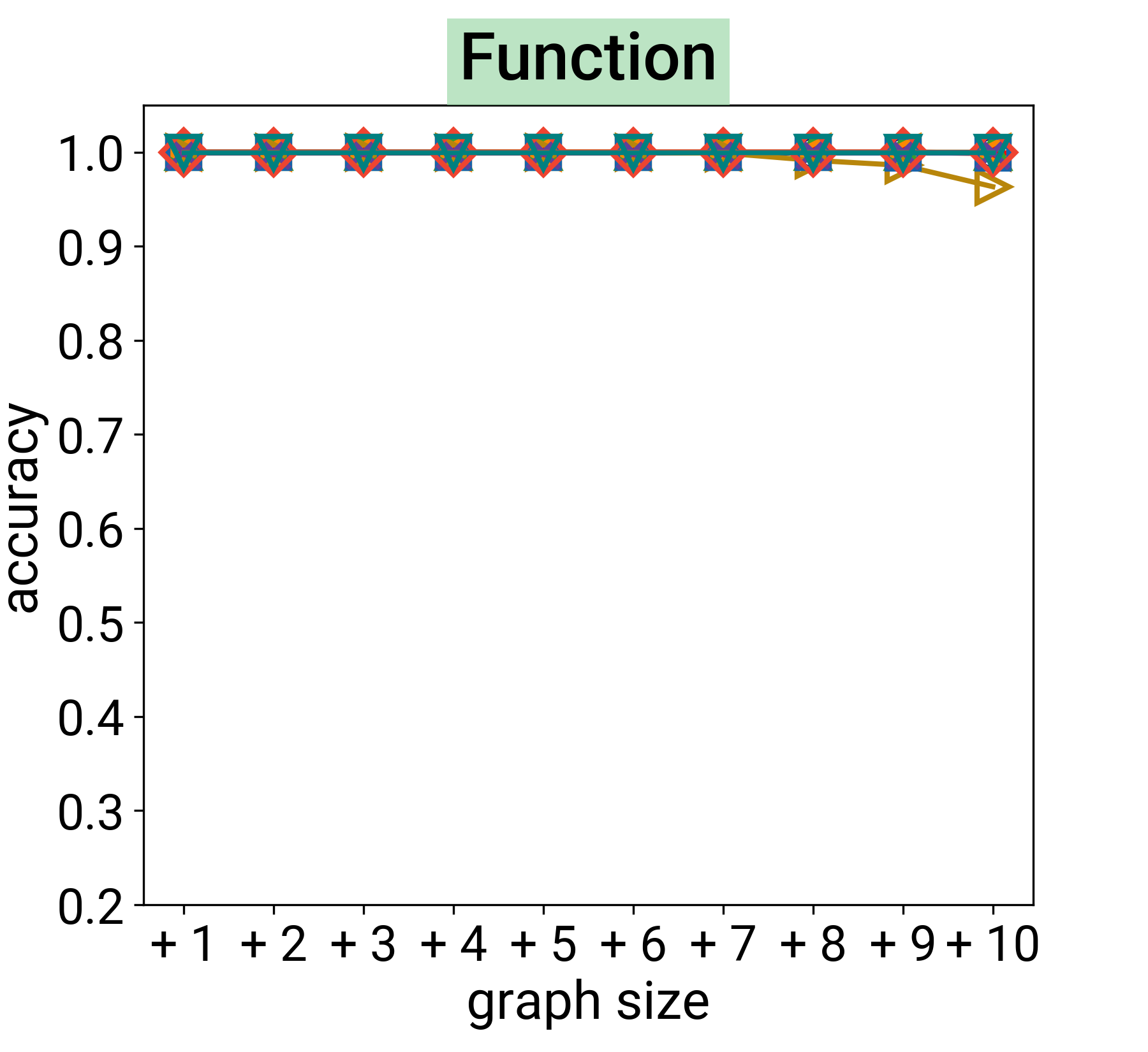}
  \end{subfigure}
    \begin{subfigure}[b]{0.24\textwidth}
    \includegraphics[width=\textwidth]{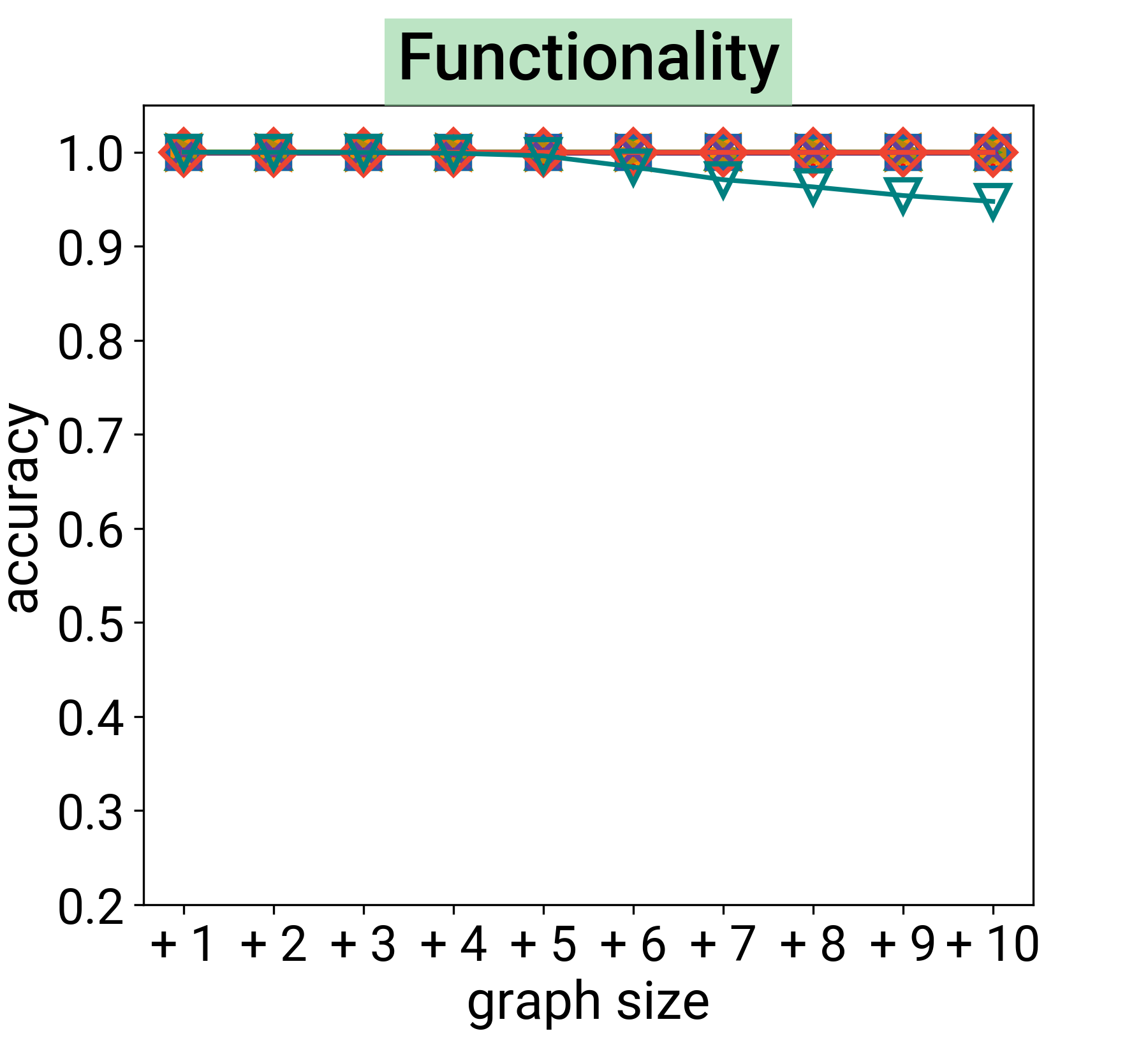}
  \end{subfigure}
    \begin{subfigure}[b]{0.24\textwidth}
    \includegraphics[width=\textwidth]{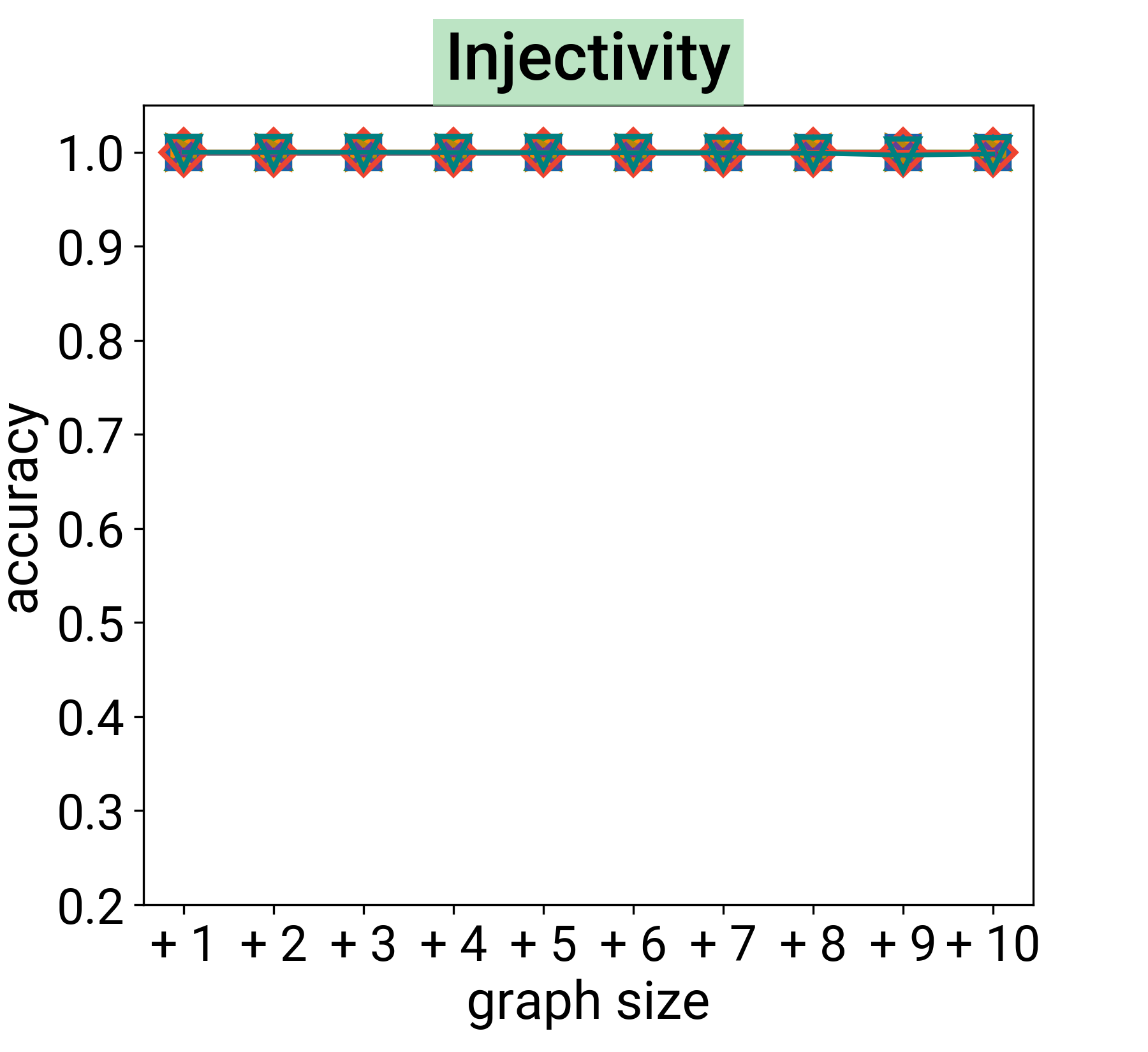}
  \end{subfigure}

   \par\smallskip
    \begin{subfigure}[b]{0.24\textwidth}
    \includegraphics[width=\textwidth]{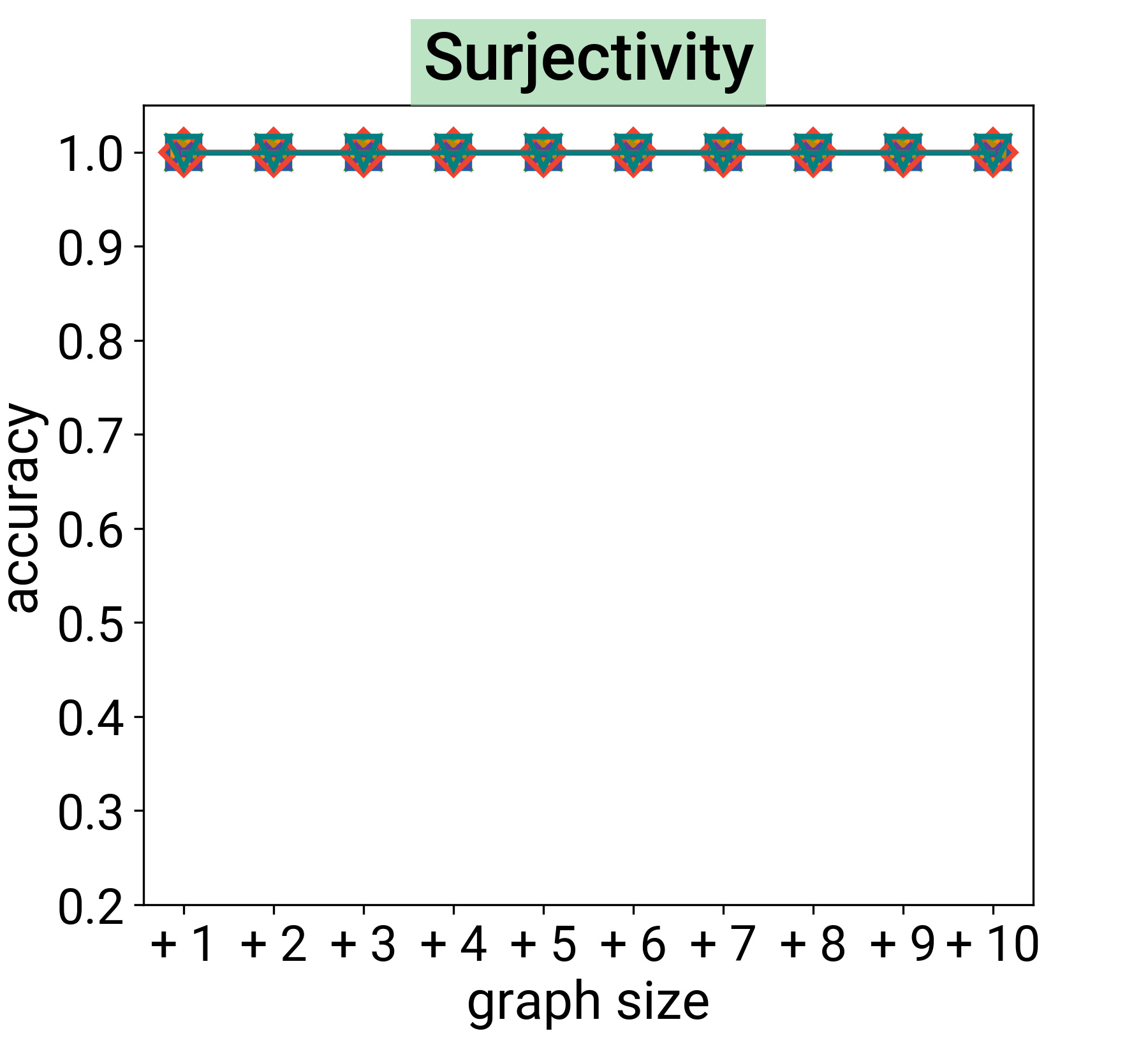}
  \end{subfigure}
  \begin{subfigure}[b]{0.24\textwidth}
    \includegraphics[width=\textwidth]{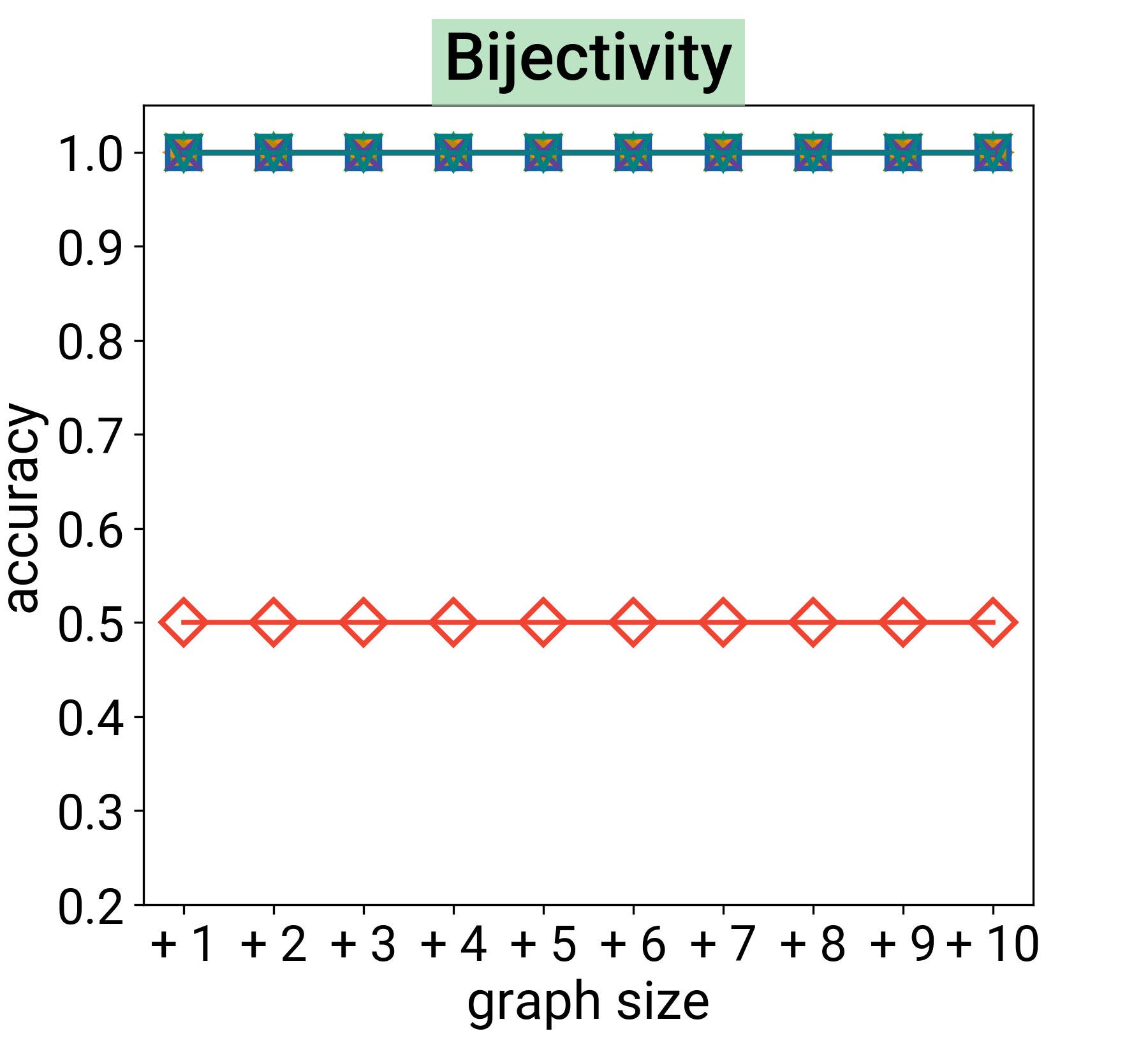}
  \end{subfigure}
  \begin{subfigure}[b]{0.24\textwidth}
    \includegraphics[width=\textwidth]{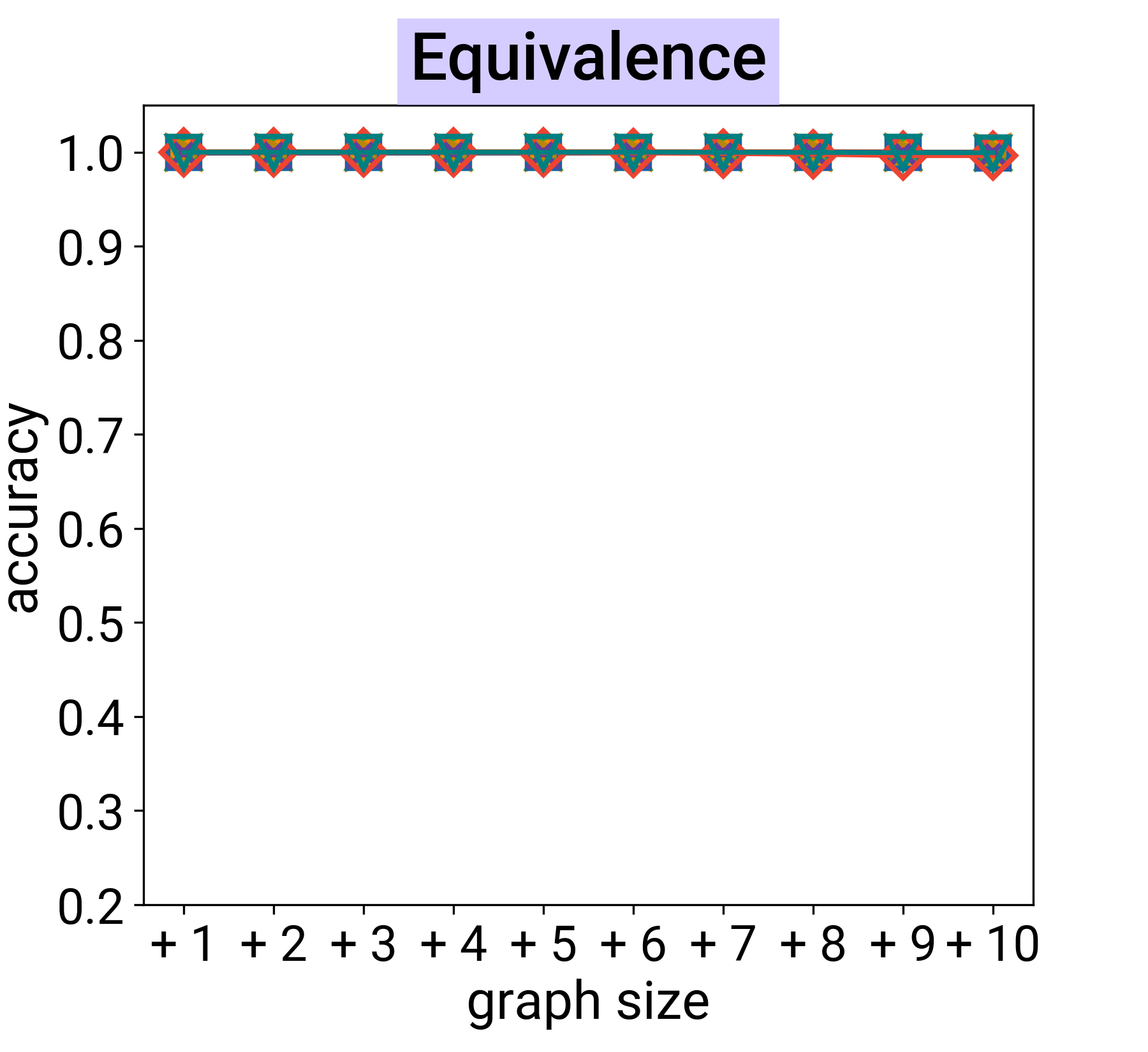}
  \end{subfigure}
    \begin{subfigure}[b]{0.24\textwidth}
    \includegraphics[width=\textwidth]{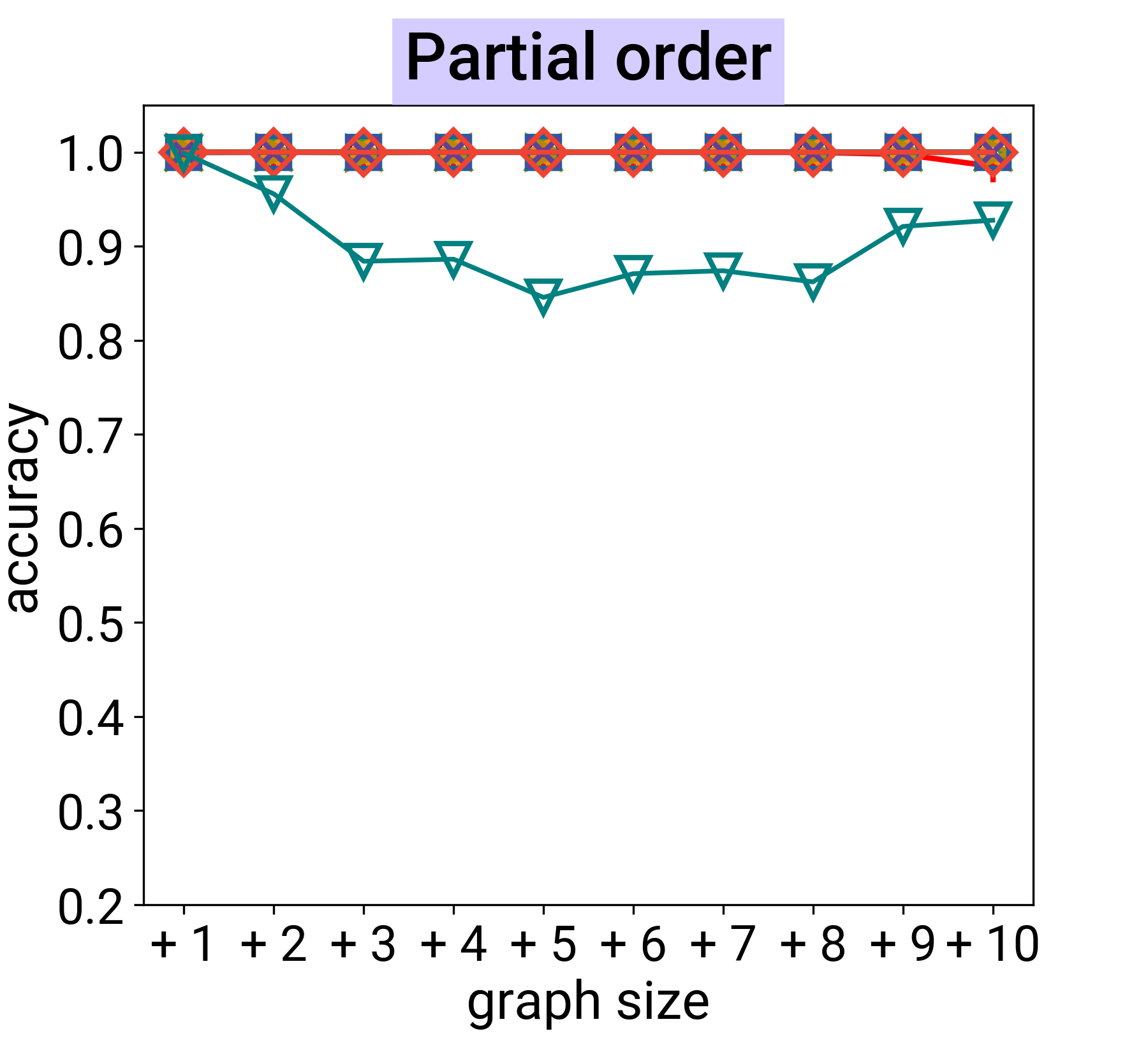}
  \end{subfigure}

  \par\smallskip
    \begin{subfigure}[b]{0.24\textwidth}
    \includegraphics[width=\textwidth]{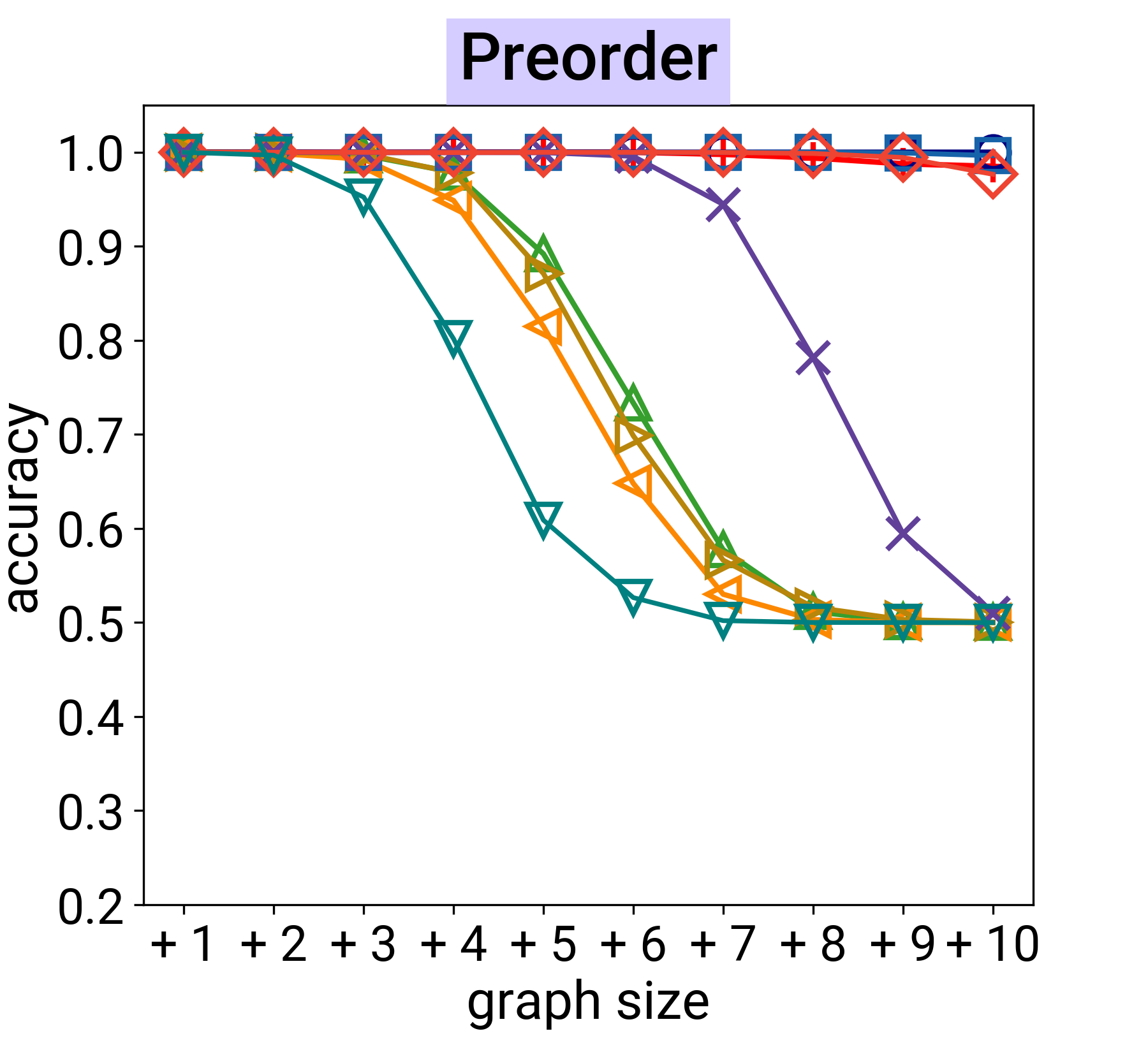}
  \end{subfigure}
    \begin{subfigure}[b]{0.24\textwidth}
    \includegraphics[width=\textwidth]{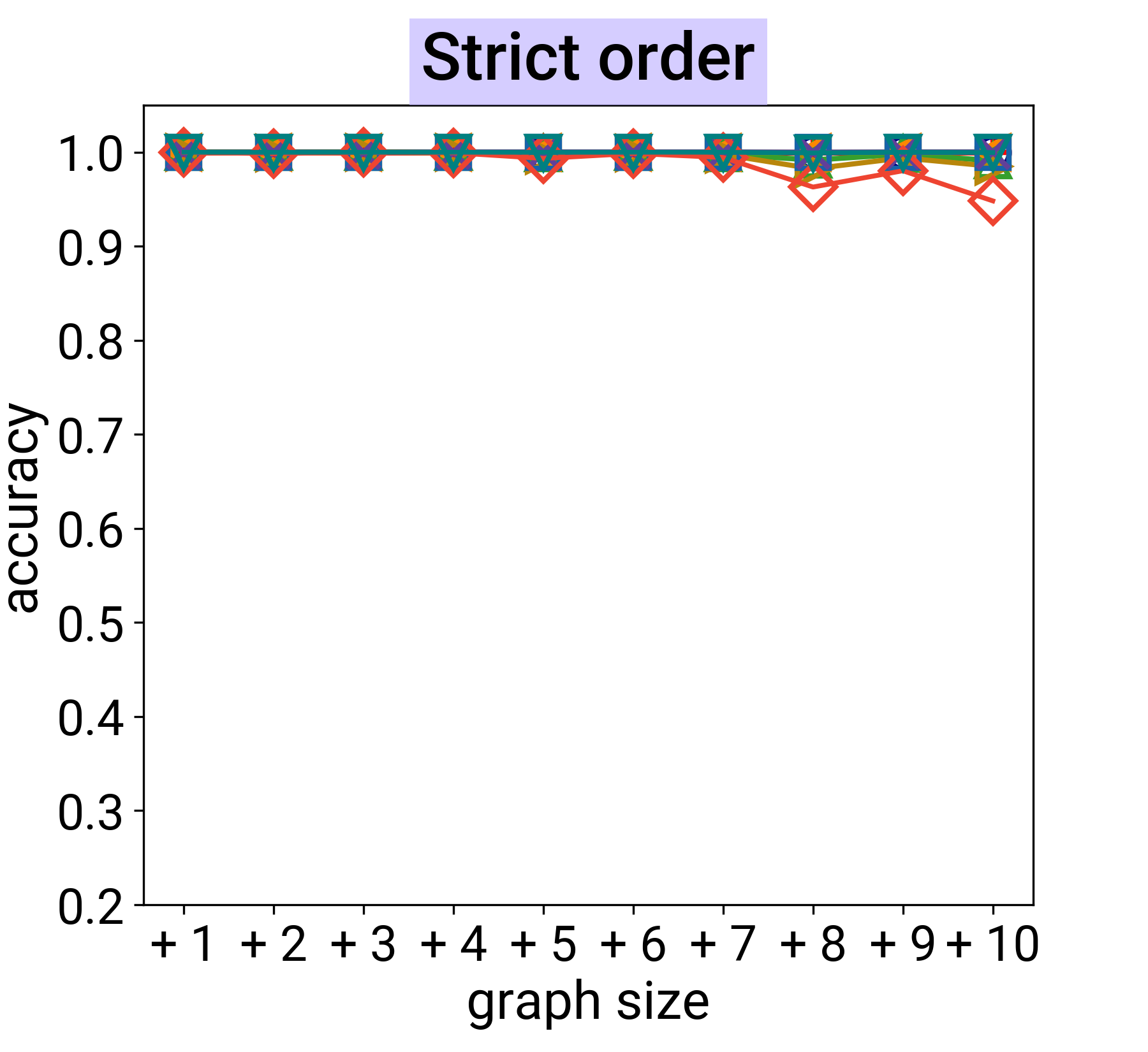}
  \end{subfigure}
  \begin{subfigure}[b]{0.24\textwidth}
    \includegraphics[width=\textwidth]{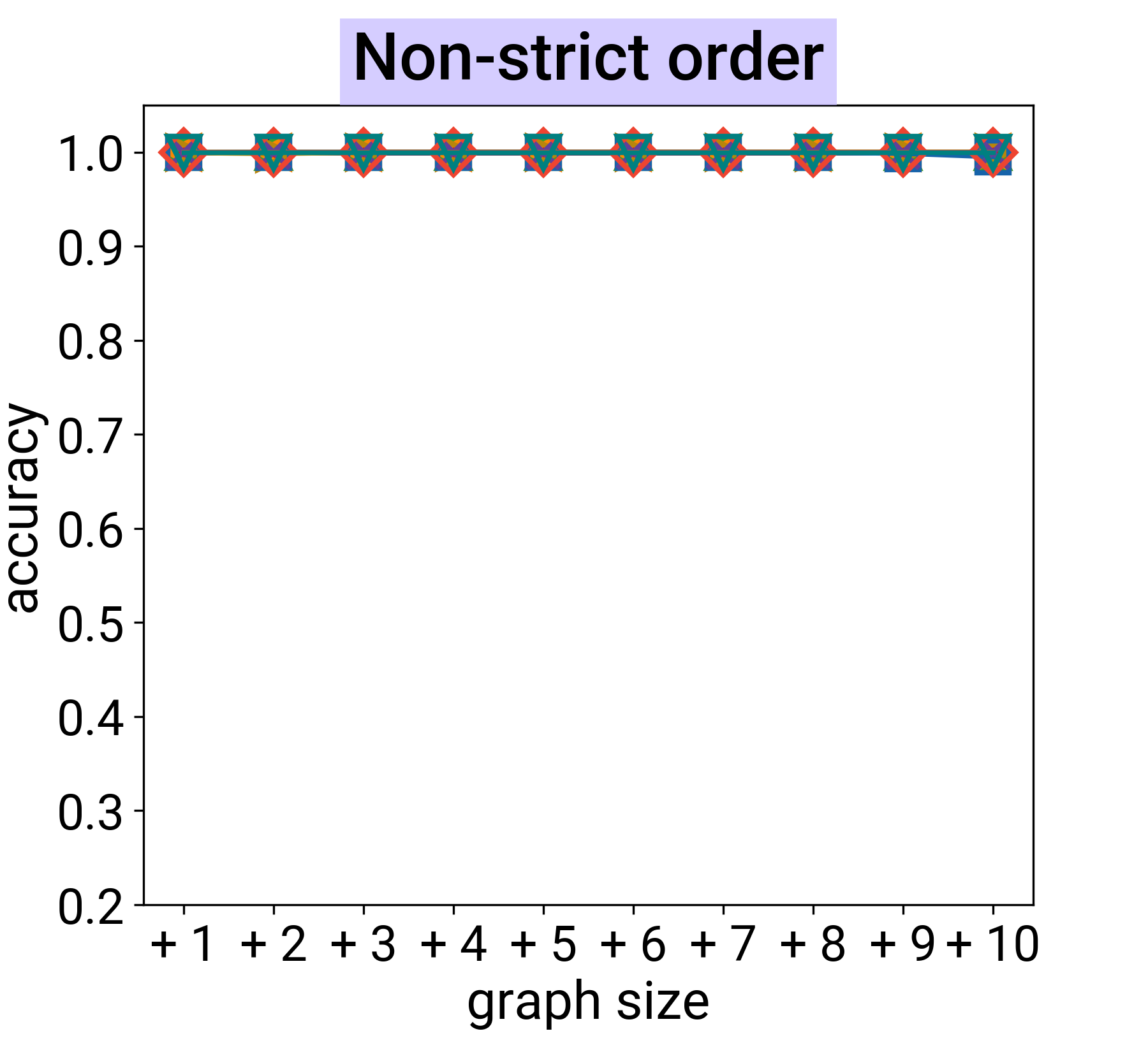}
  \end{subfigure}
    \begin{subfigure}[b]{0.24\textwidth}
    \includegraphics[width=\textwidth]{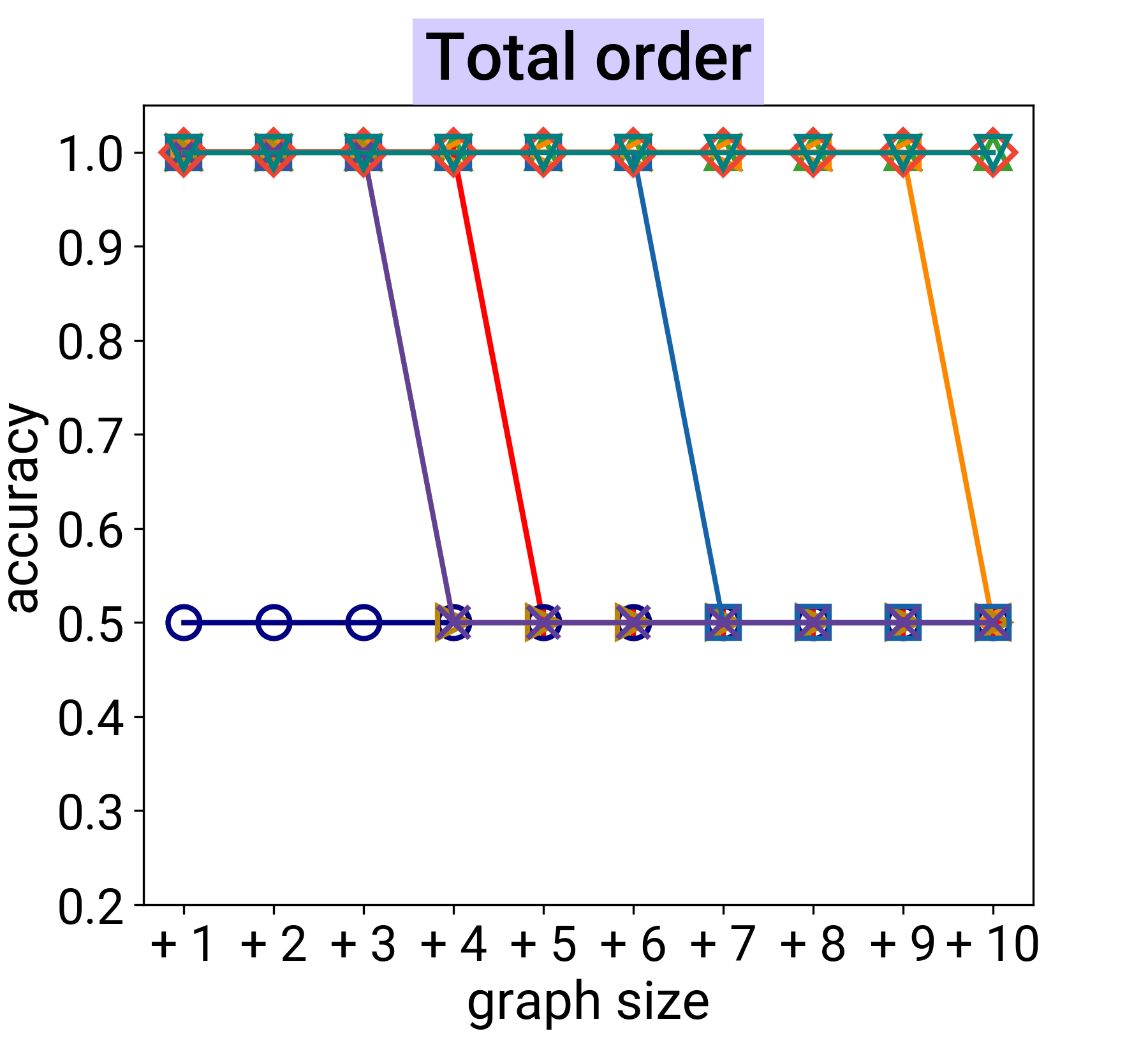}
  \end{subfigure}
\vspace{-2ex}
  \caption{ Global pooling performance across ten graph sizes under generalizability aspect.}
  \vspace{-4ex}
  \label{fig:line_generalizability}
\end{figure*}

\begin{figure*}[t]
  \centering

    \begin{subfigure}{\textwidth}
        \centering
        \includegraphics[width=0.5\textwidth]{Pics/line_legend.png}
        \label{fig:line_legend}
    \end{subfigure}

  \begin{subfigure}[b]{0.24\textwidth}
    \includegraphics[width=\textwidth]{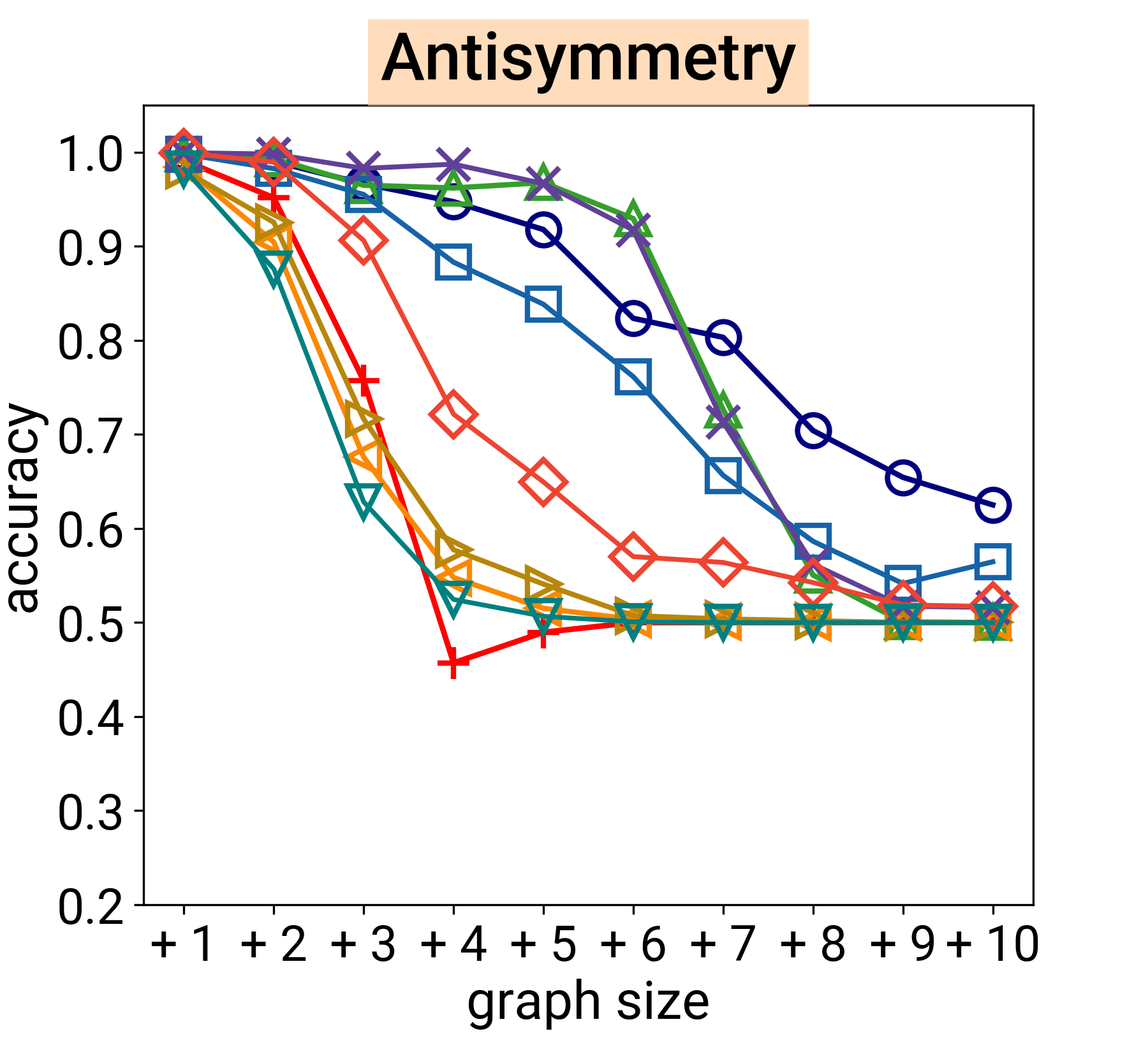}
  \end{subfigure}
    \begin{subfigure}[b]{0.24\textwidth}
    \includegraphics[width=\textwidth]{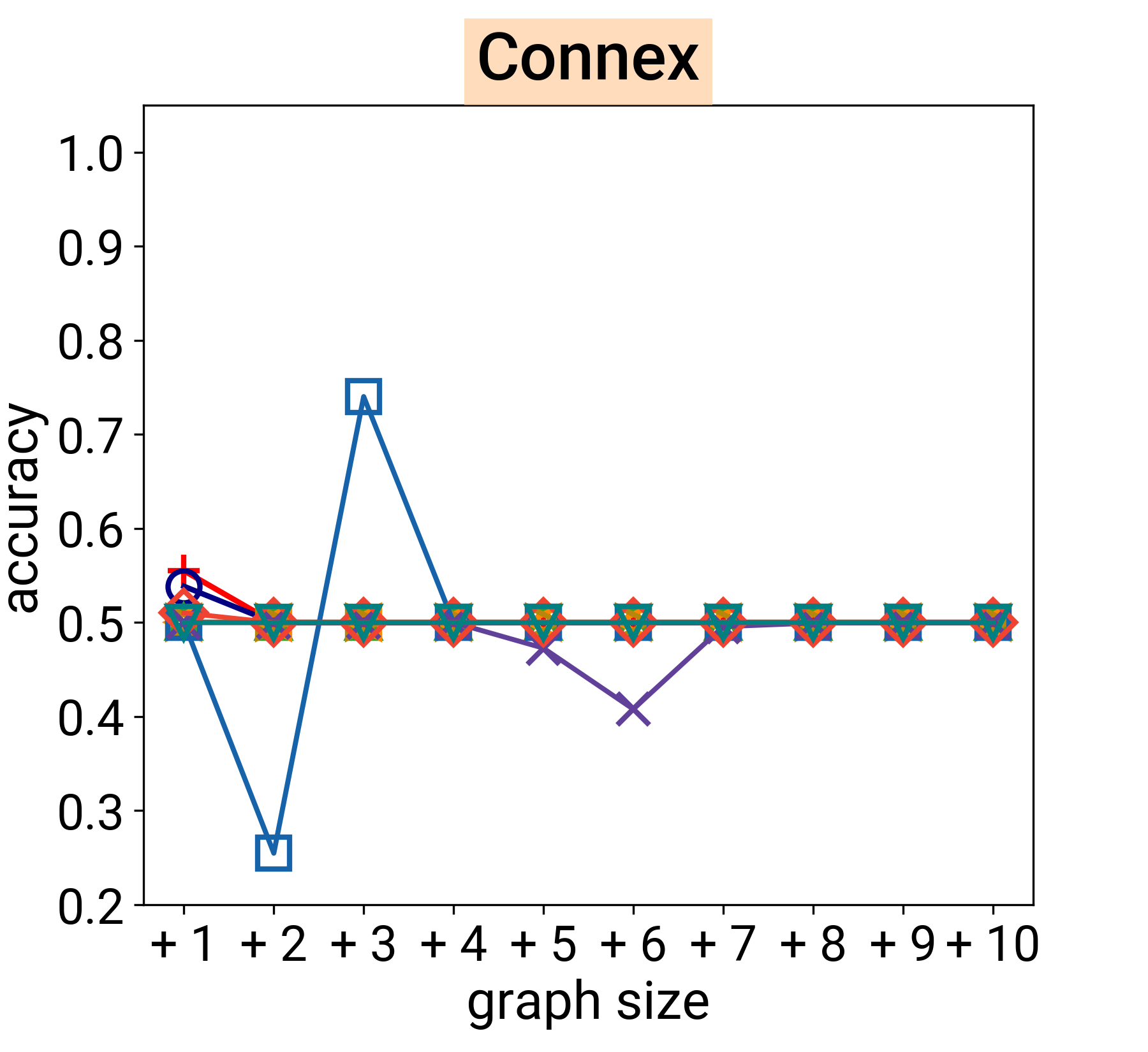}
  \end{subfigure}
    \begin{subfigure}[b]{0.24\textwidth}
    \includegraphics[width=\textwidth]{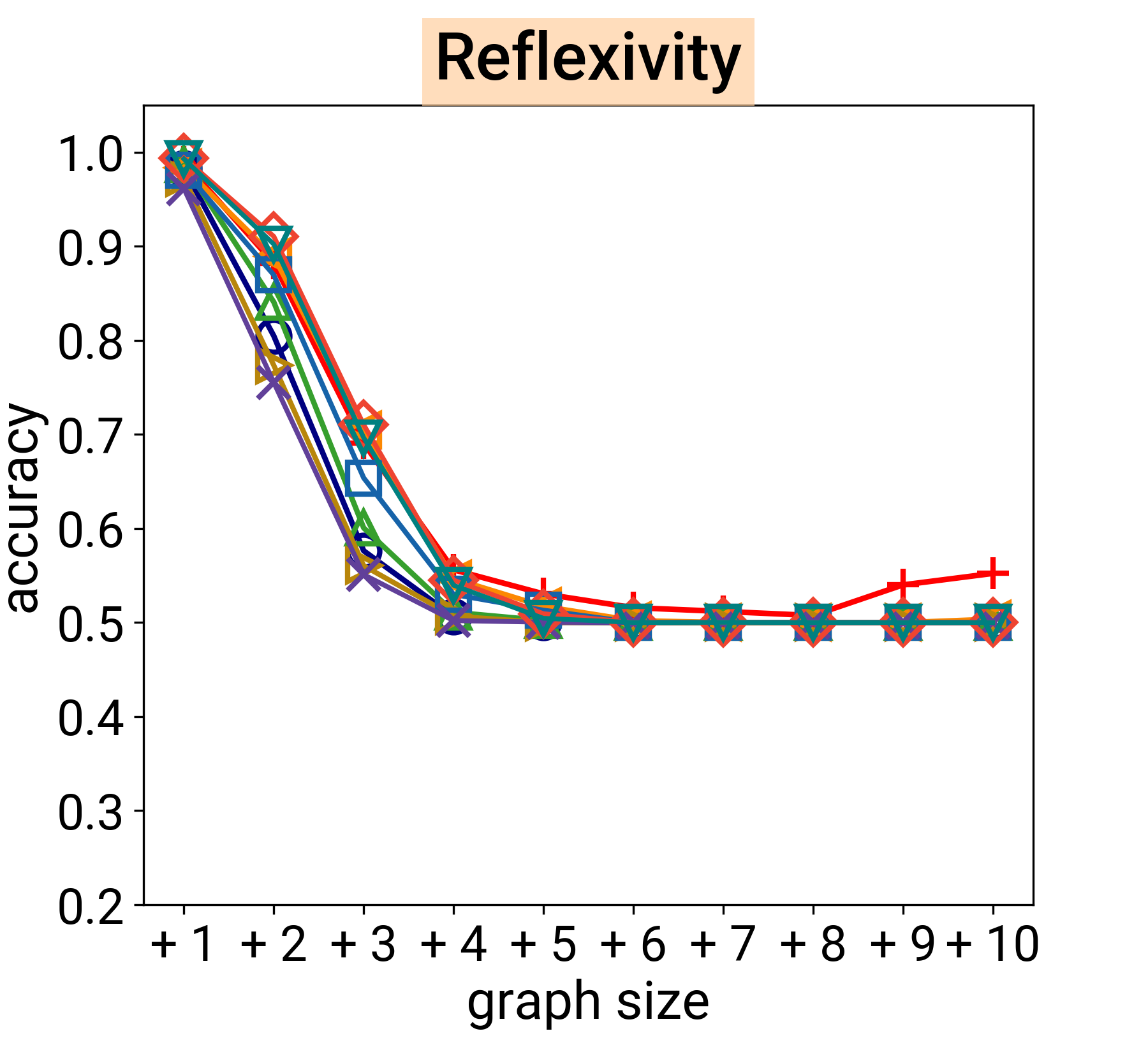}
  \end{subfigure}
    \begin{subfigure}[b]{0.24\textwidth}
    \includegraphics[width=\textwidth]{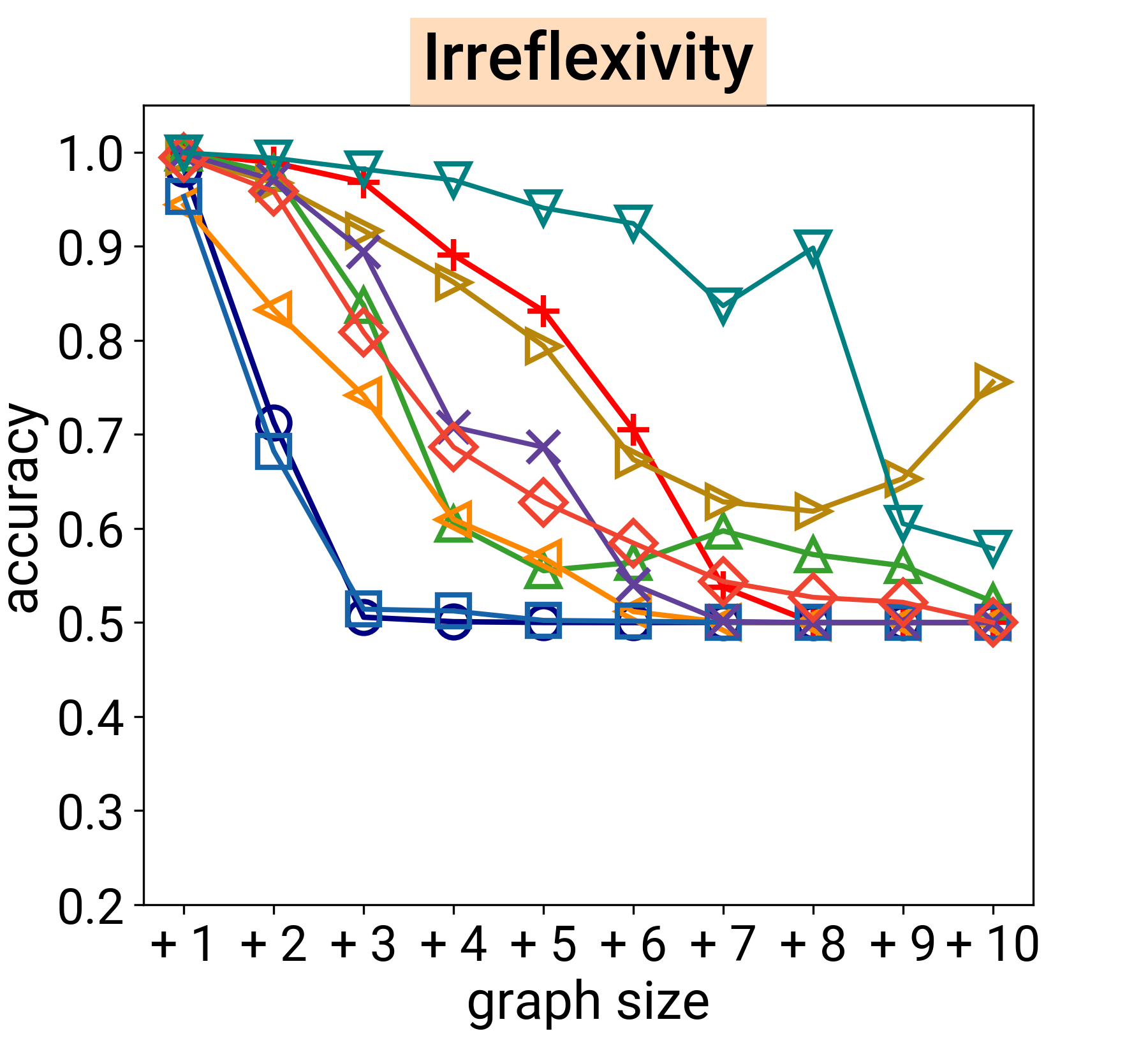}
  \end{subfigure}

  \par\smallskip
    \begin{subfigure}[b]{0.24\textwidth}
    \includegraphics[width=\textwidth]{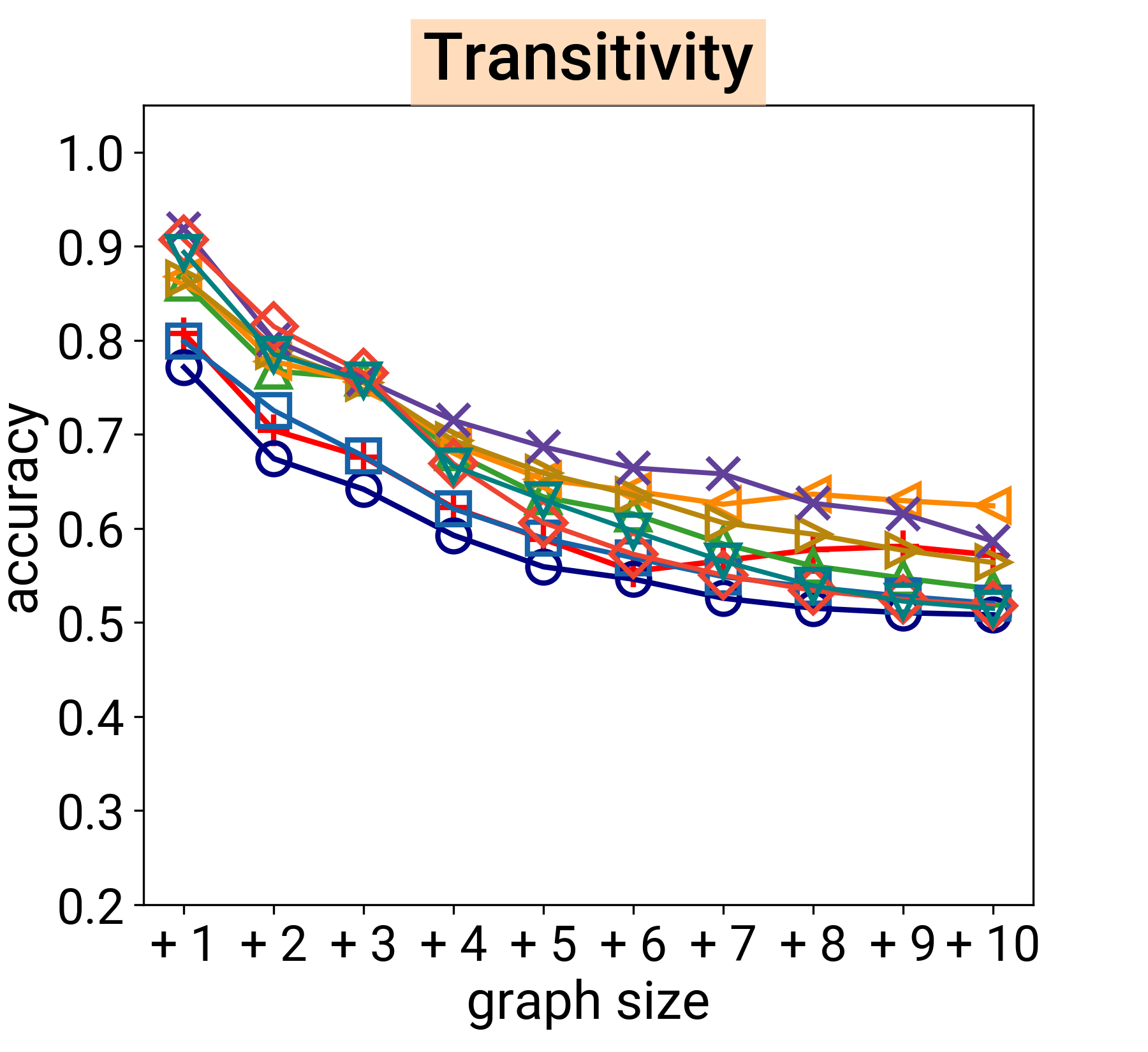}
  \end{subfigure}
    \begin{subfigure}[b]{0.24\textwidth}
    \includegraphics[width=\textwidth]{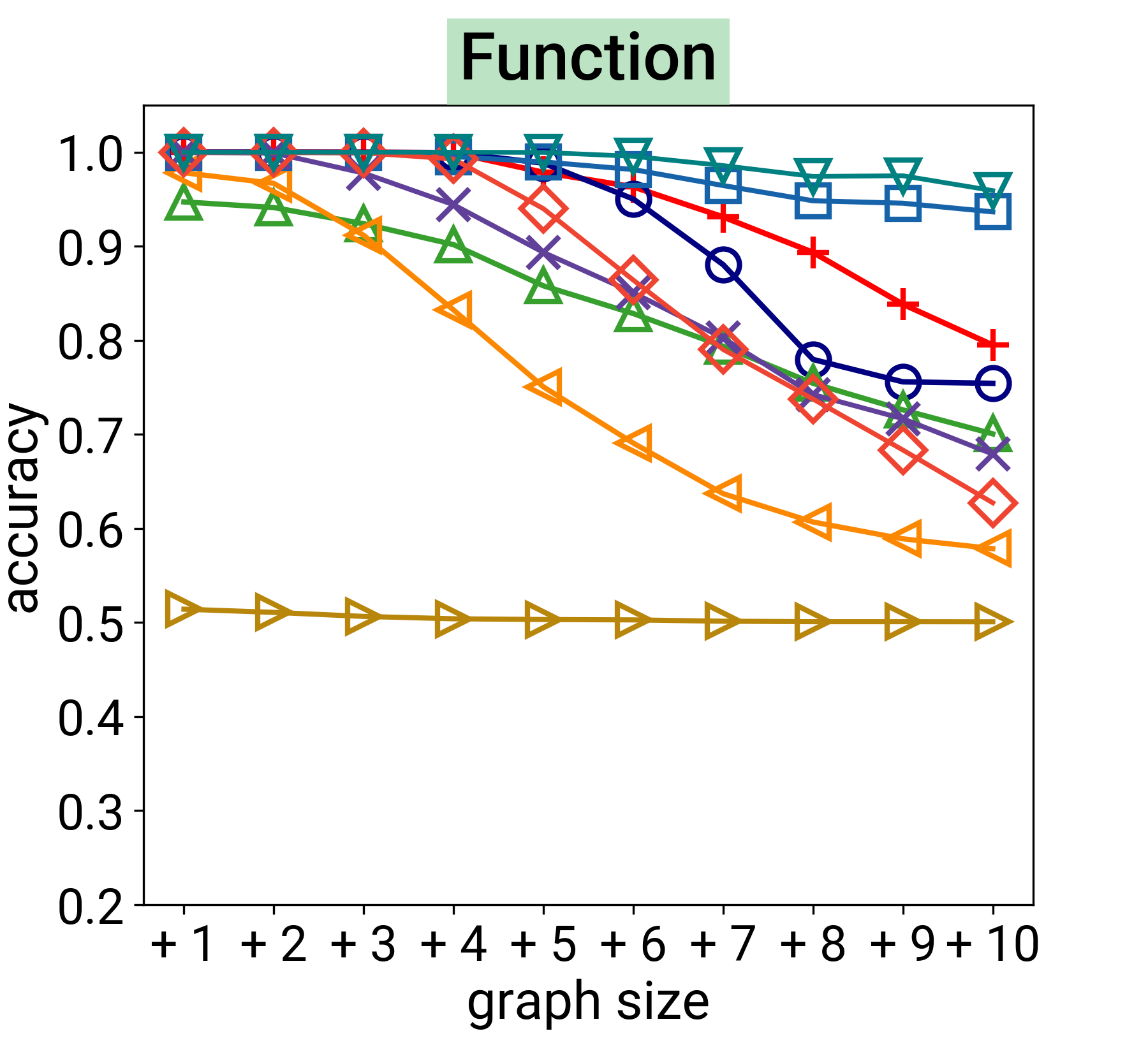}
  \end{subfigure}
    \begin{subfigure}[b]{0.24\textwidth}
    \includegraphics[width=\textwidth]{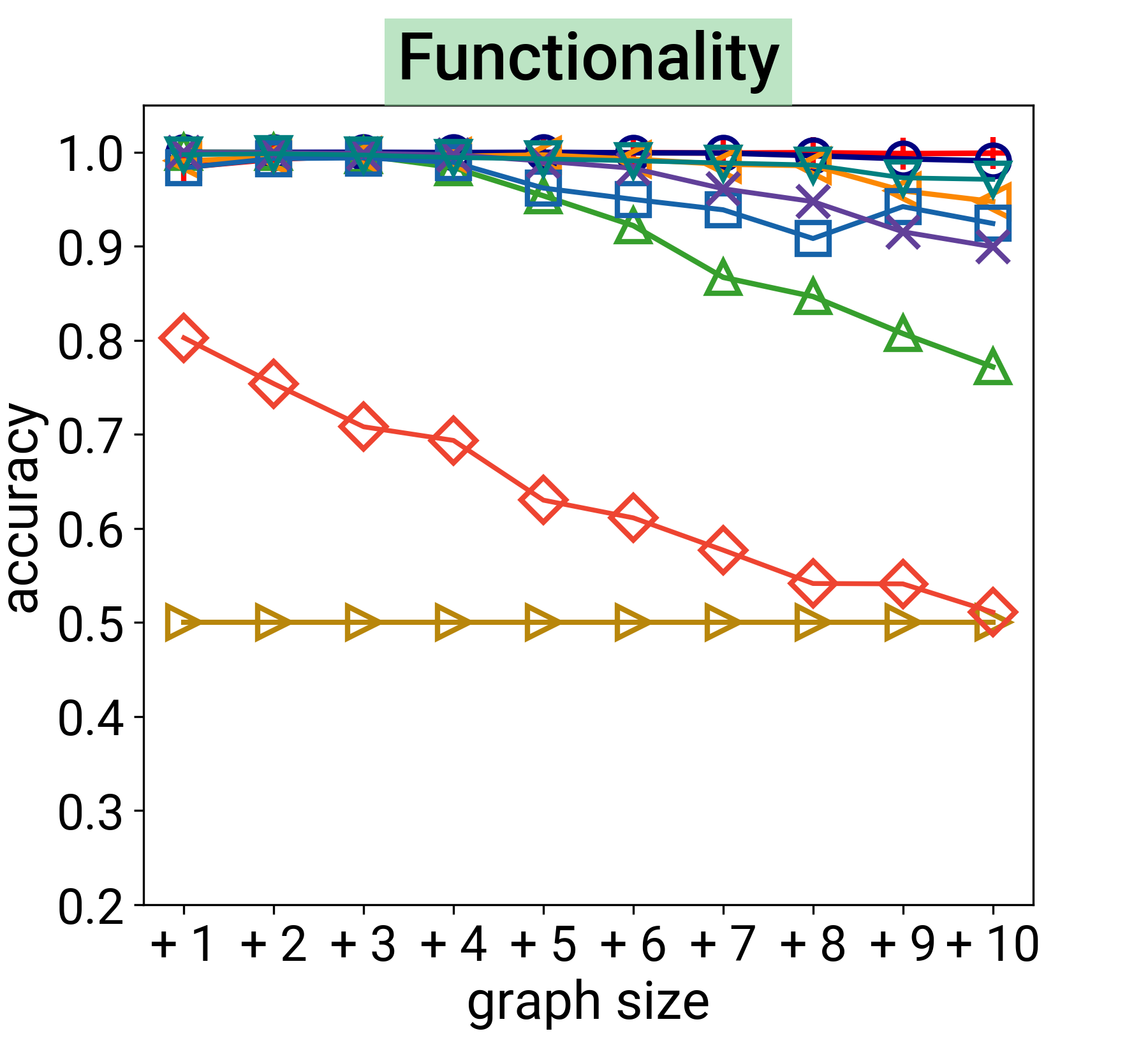}
  \end{subfigure}
    \begin{subfigure}[b]{0.24\textwidth}
    \includegraphics[width=\textwidth]{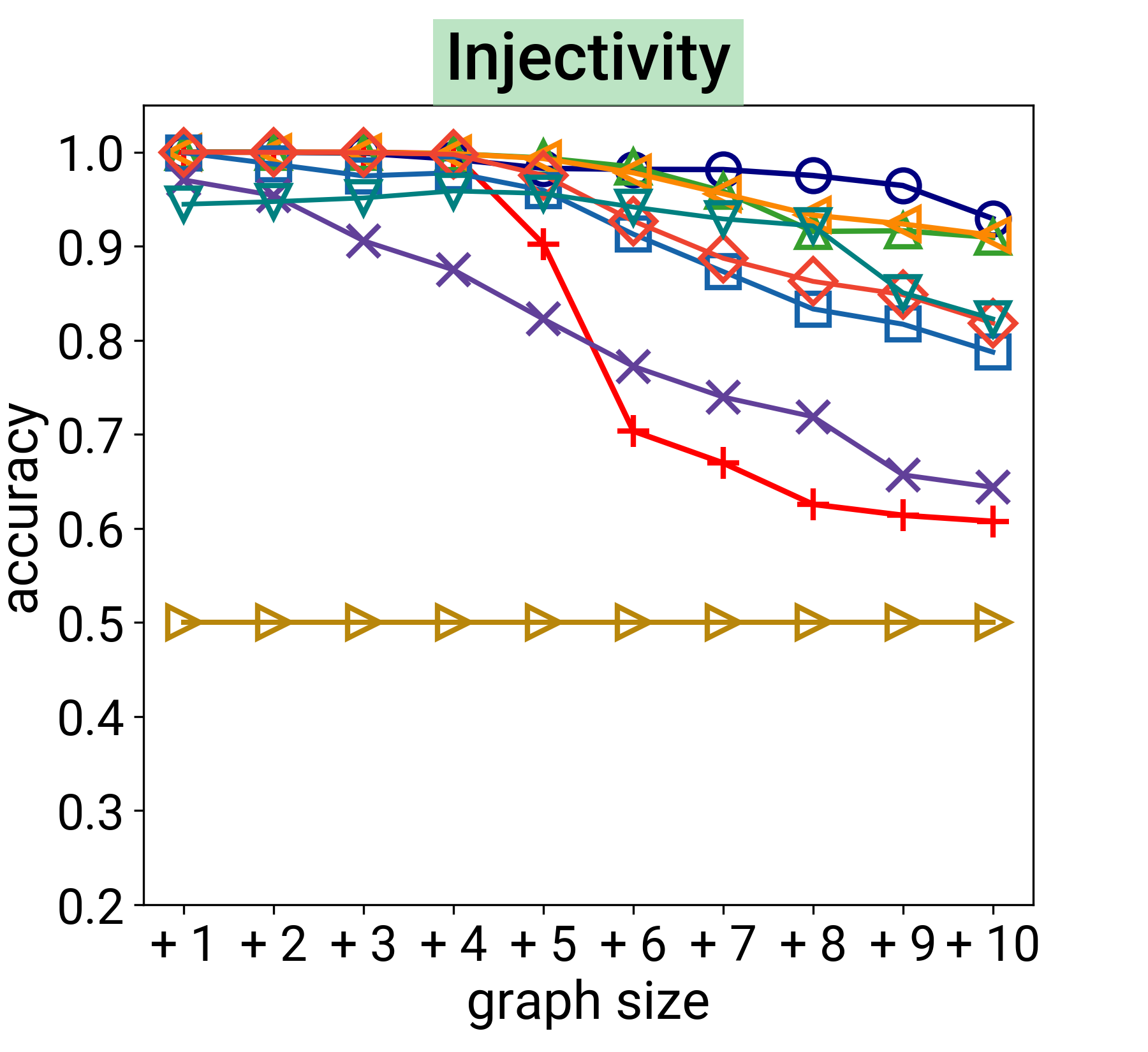}
  \end{subfigure}

   \par\smallskip
    \begin{subfigure}[b]{0.24\textwidth}
    \includegraphics[width=\textwidth]{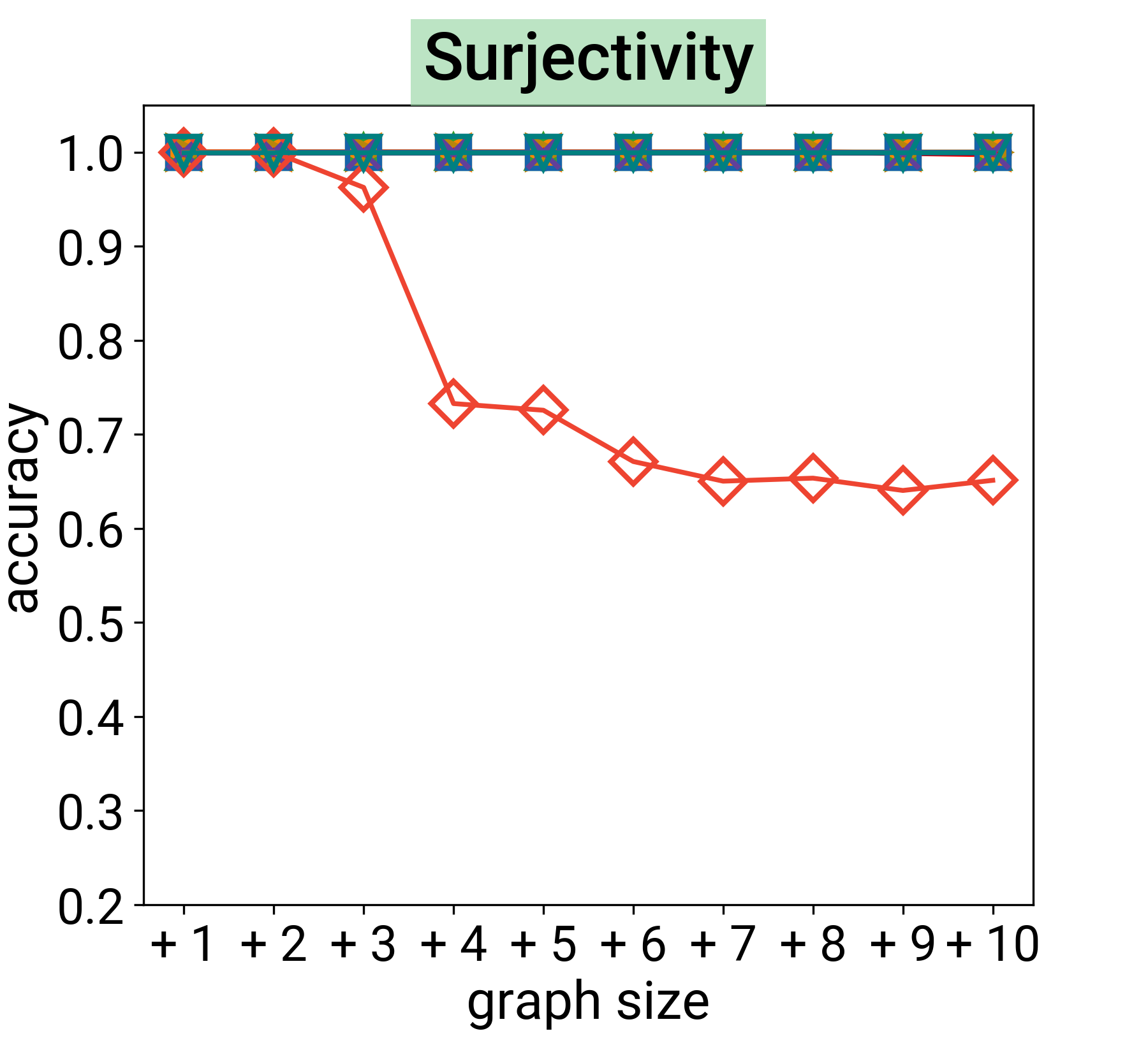}
  \end{subfigure}
  \begin{subfigure}[b]{0.24\textwidth}
    \includegraphics[width=\textwidth]{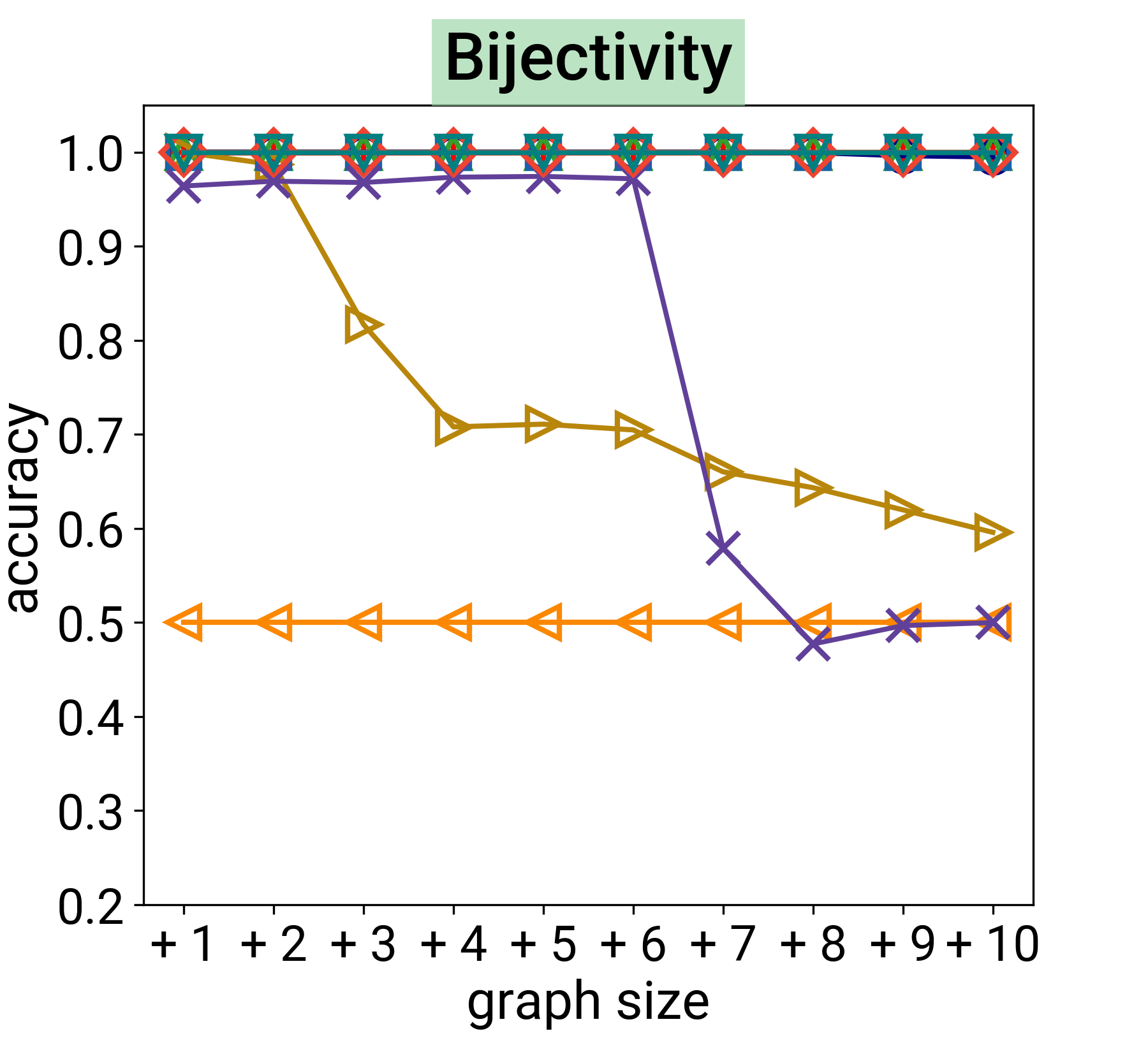}
  \end{subfigure}
  \begin{subfigure}[b]{0.24\textwidth}
    \includegraphics[width=\textwidth]{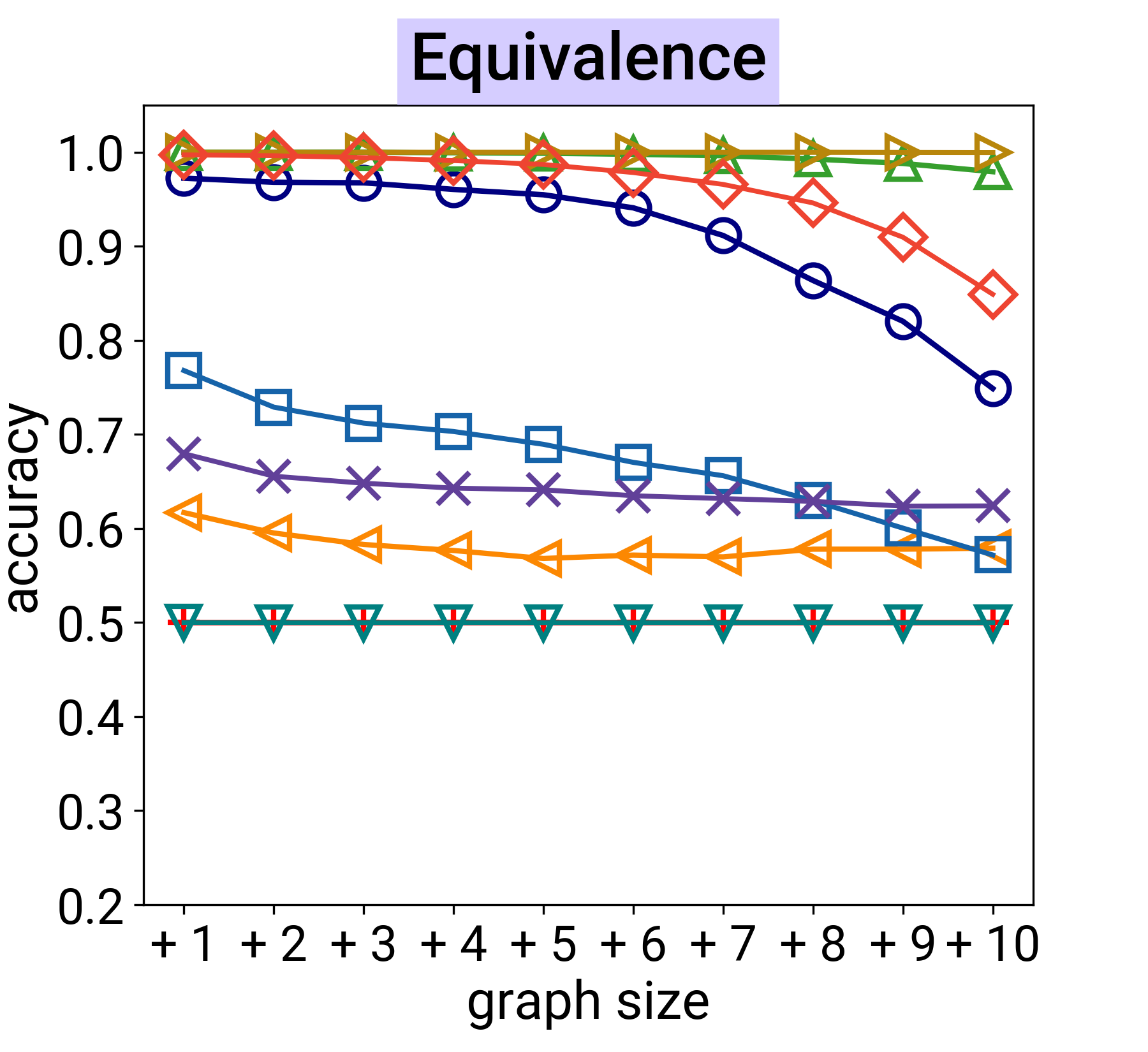}
  \end{subfigure}
    \begin{subfigure}[b]{0.24\textwidth}
    \includegraphics[width=\textwidth]{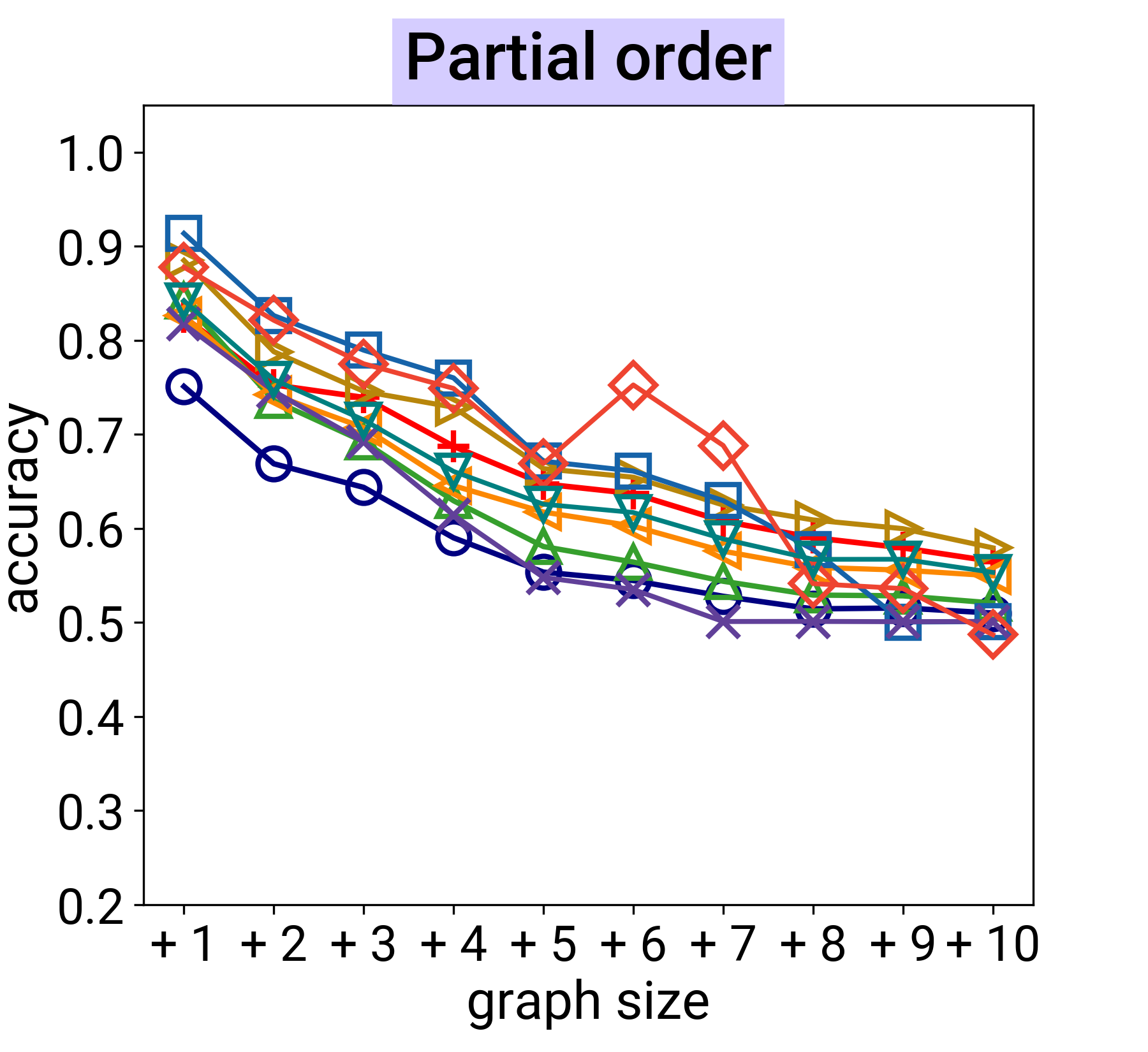}
  \end{subfigure}

  \par\smallskip
    \begin{subfigure}[b]{0.24\textwidth}
    \includegraphics[width=\textwidth]{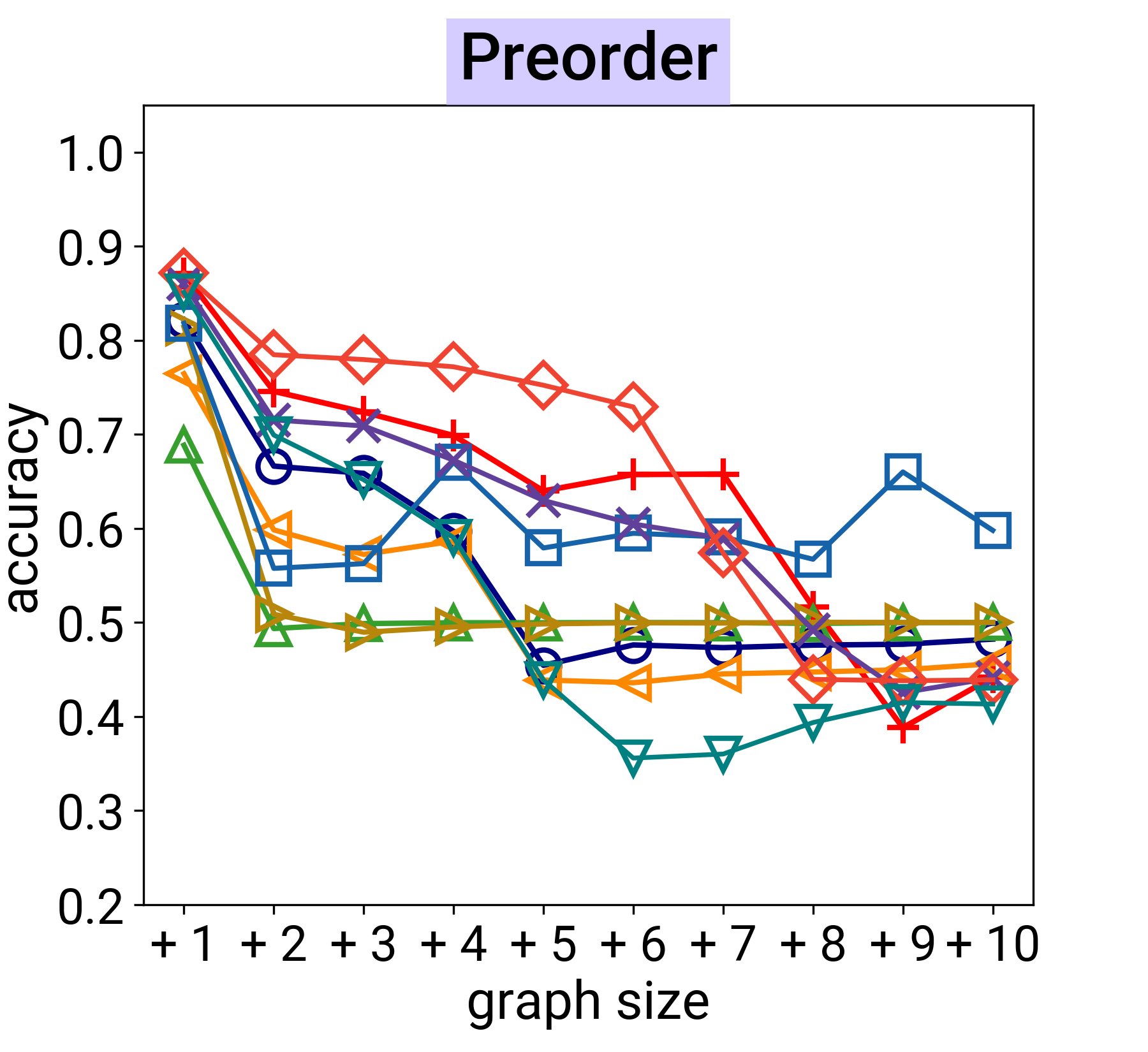}
  \end{subfigure}
    \begin{subfigure}[b]{0.24\textwidth}
    \includegraphics[width=\textwidth]{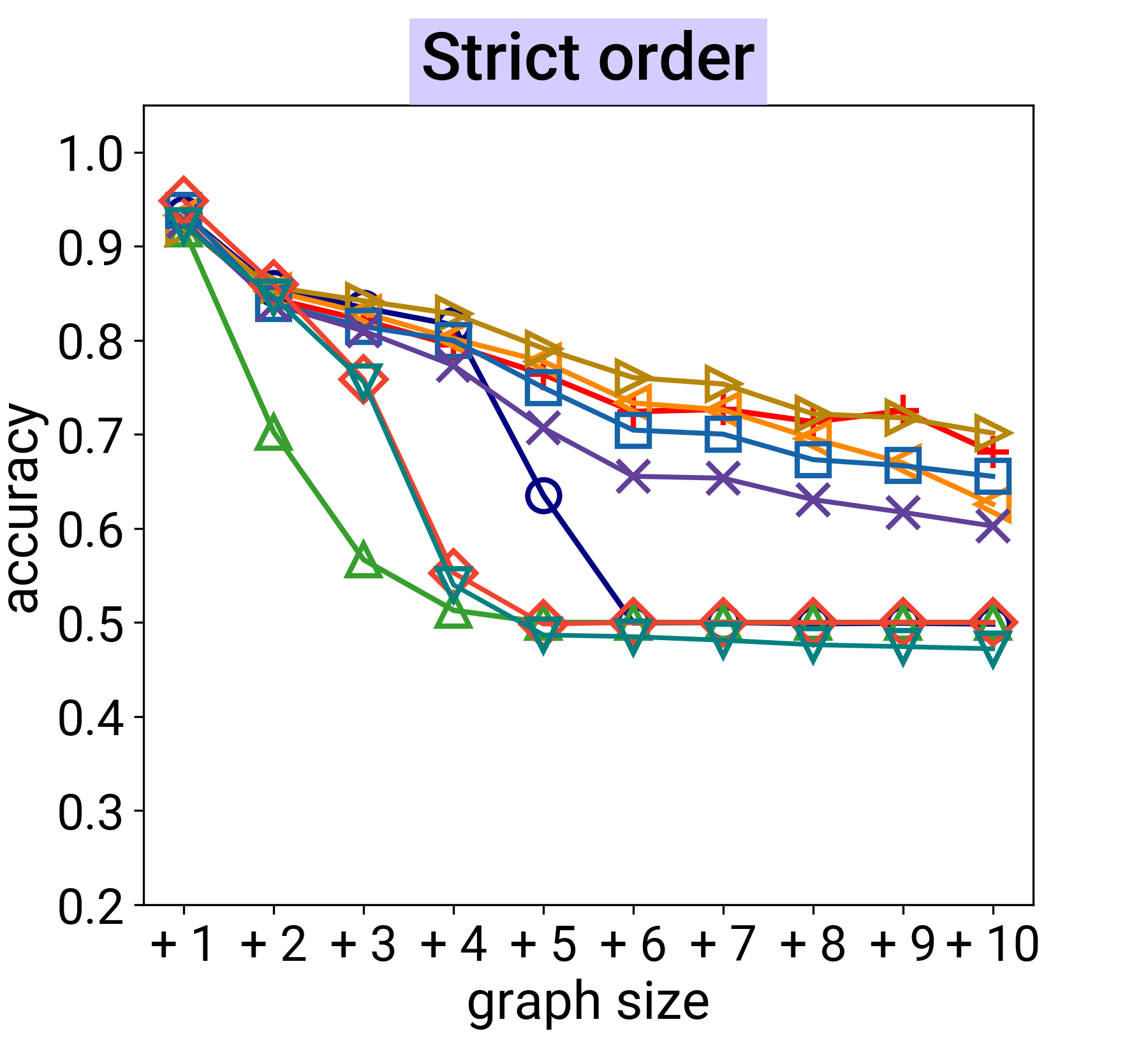}
  \end{subfigure}
  \begin{subfigure}[b]{0.24\textwidth}
    \includegraphics[width=\textwidth]{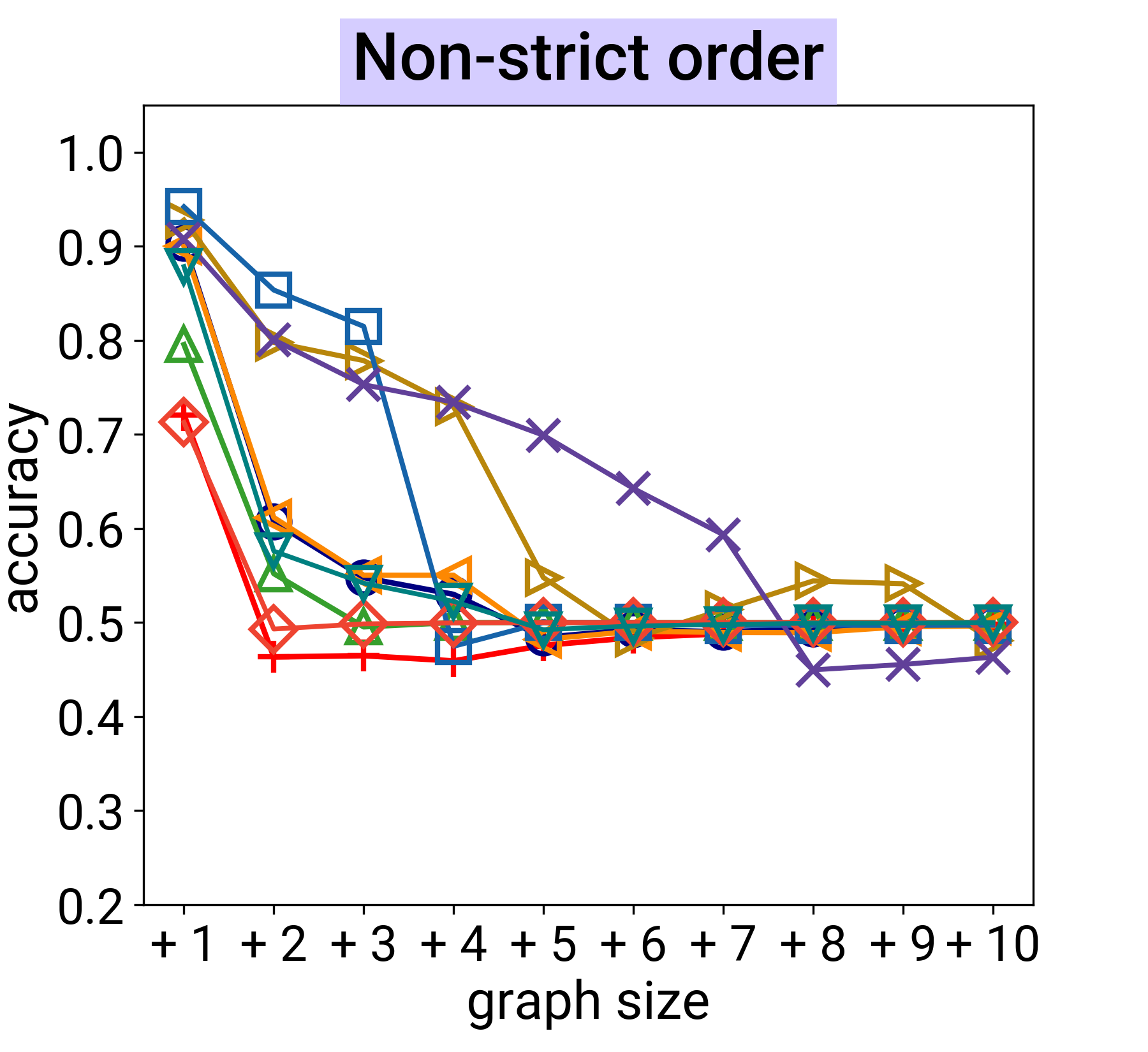}
  \end{subfigure}
    \begin{subfigure}[b]{0.24\textwidth}
    \includegraphics[width=\textwidth]{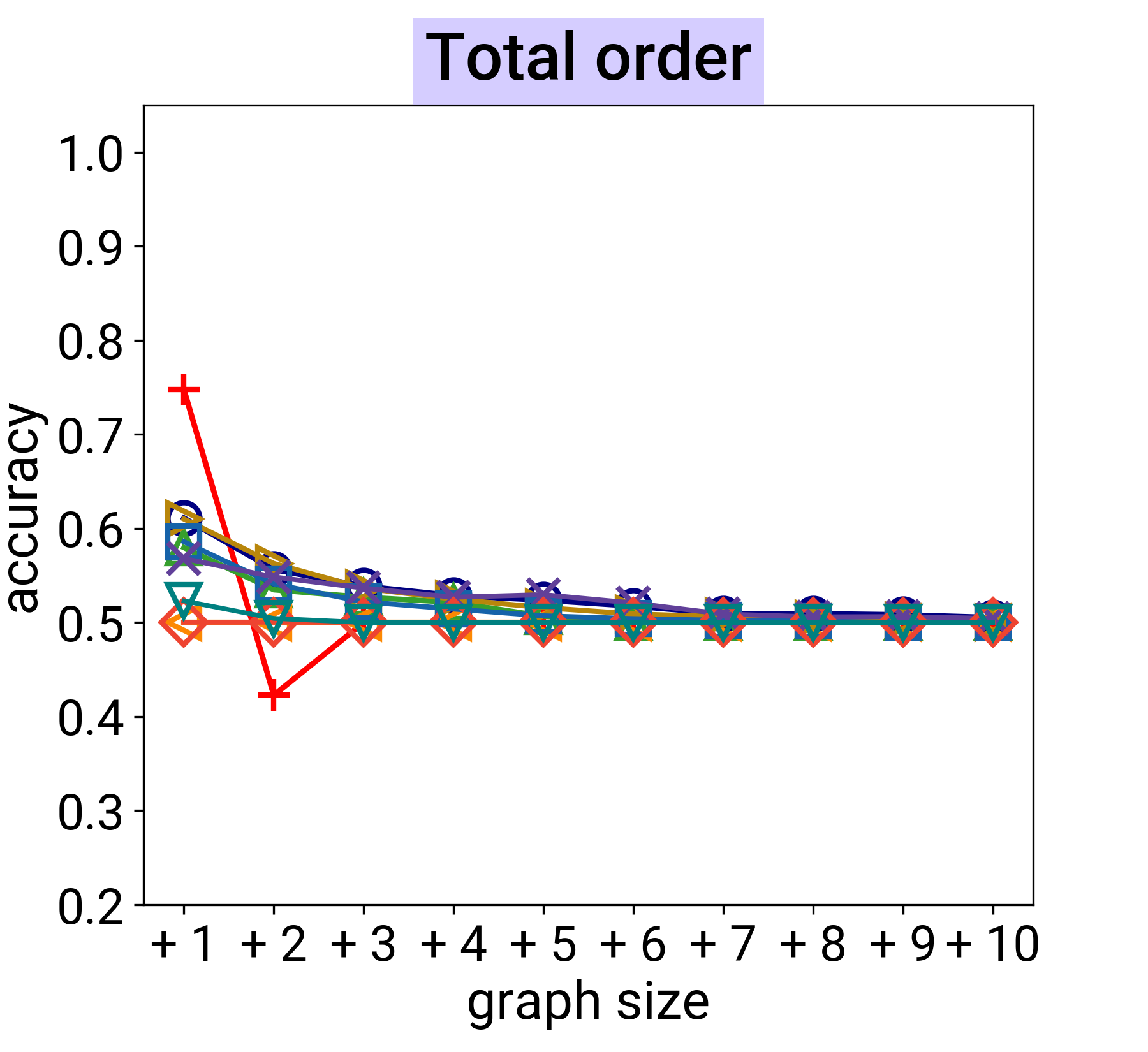}
  \end{subfigure}
\vspace{-2ex}
  \caption{ Global pooling performance across ten graph sizes under sensitivity aspect.}
  \vspace{-4ex}
  \label{fig:line_sensitivity}
\end{figure*}

\begin{figure*}[t]
  \centering

    \begin{subfigure}{\textwidth}
        \centering
        \includegraphics[width=0.5\textwidth]{Pics/line_legend.png}
        \label{fig:line_legend}
    \end{subfigure}
    
  \begin{subfigure}[b]{0.24\textwidth}
    \includegraphics[width=\textwidth]{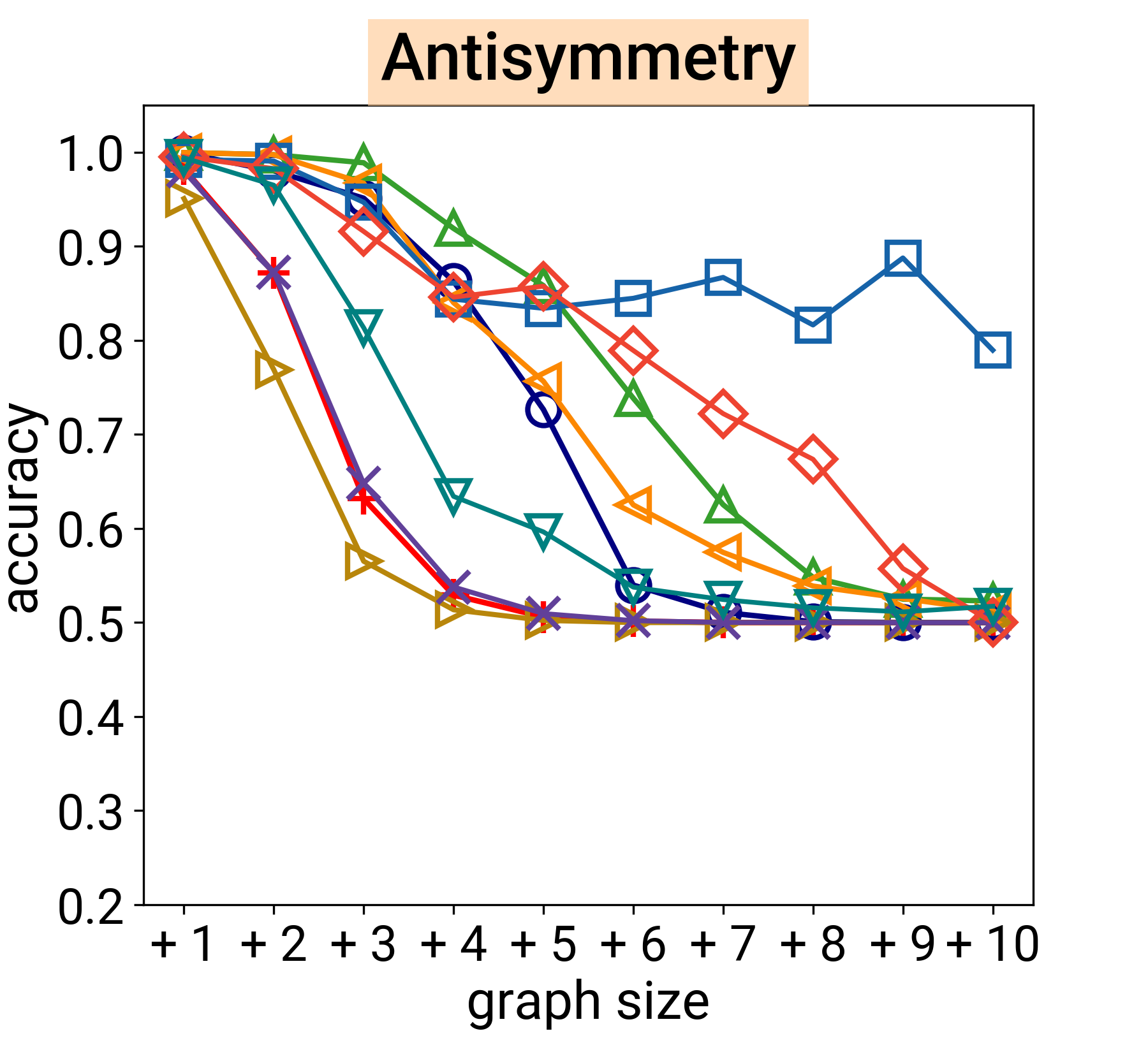}
  \end{subfigure}
    \begin{subfigure}[b]{0.24\textwidth}
    \includegraphics[width=\textwidth]{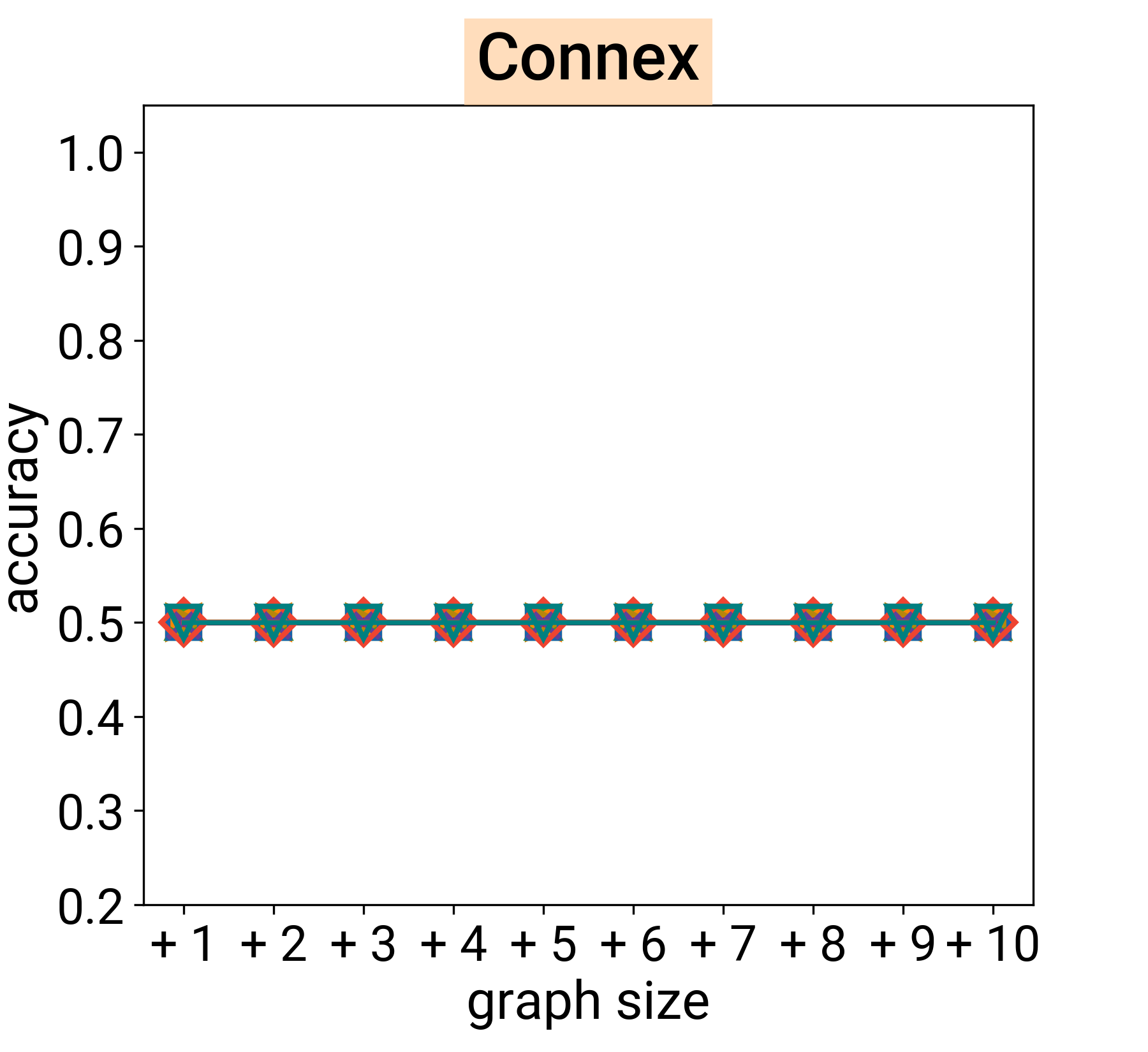}
  \end{subfigure}
    \begin{subfigure}[b]{0.24\textwidth}
    \includegraphics[width=\textwidth]{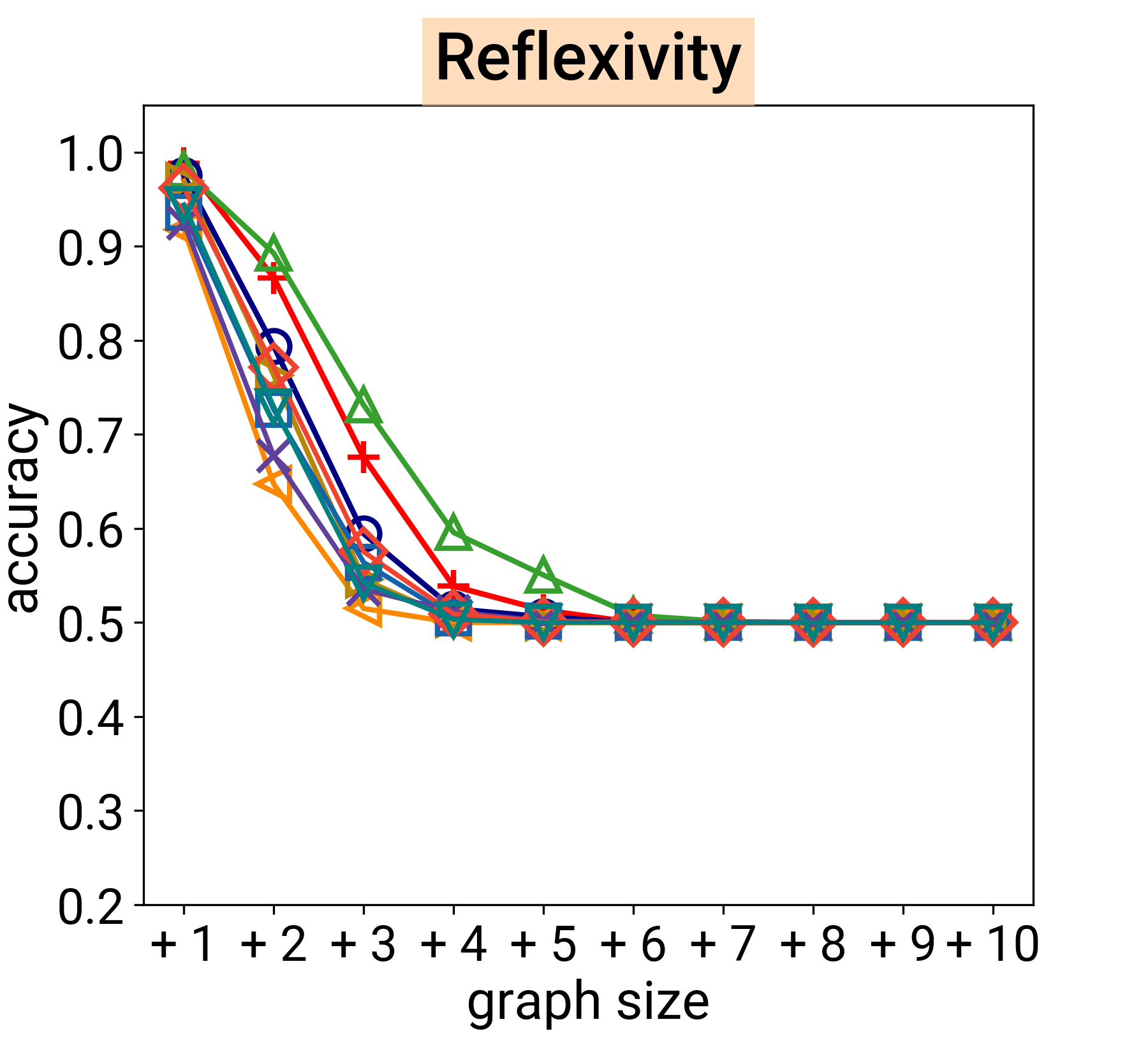}
  \end{subfigure}
    \begin{subfigure}[b]{0.24\textwidth}
    \includegraphics[width=\textwidth]{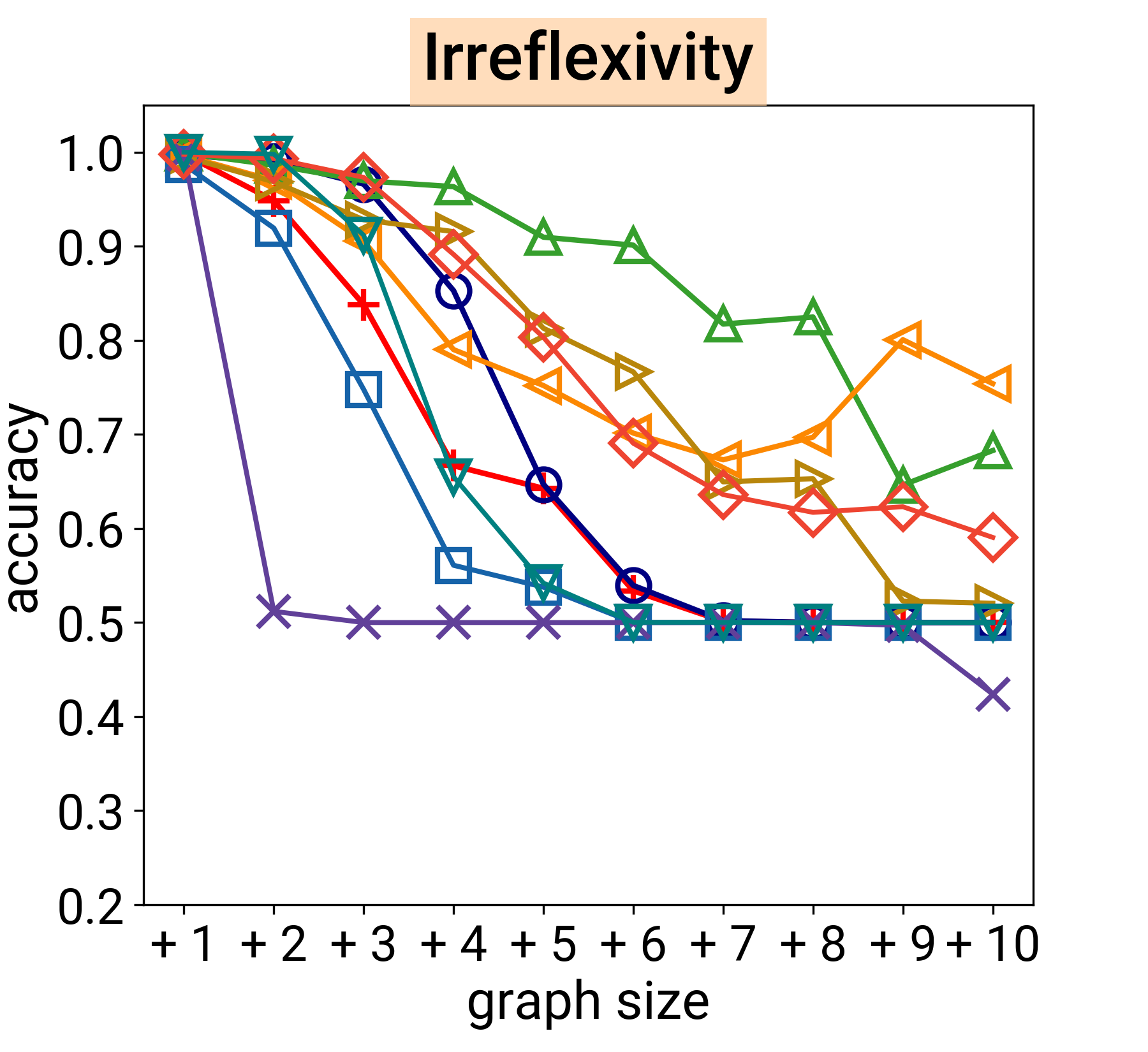}
  \end{subfigure}

  \par\smallskip
    \begin{subfigure}[b]{0.24\textwidth}
    \includegraphics[width=\textwidth]{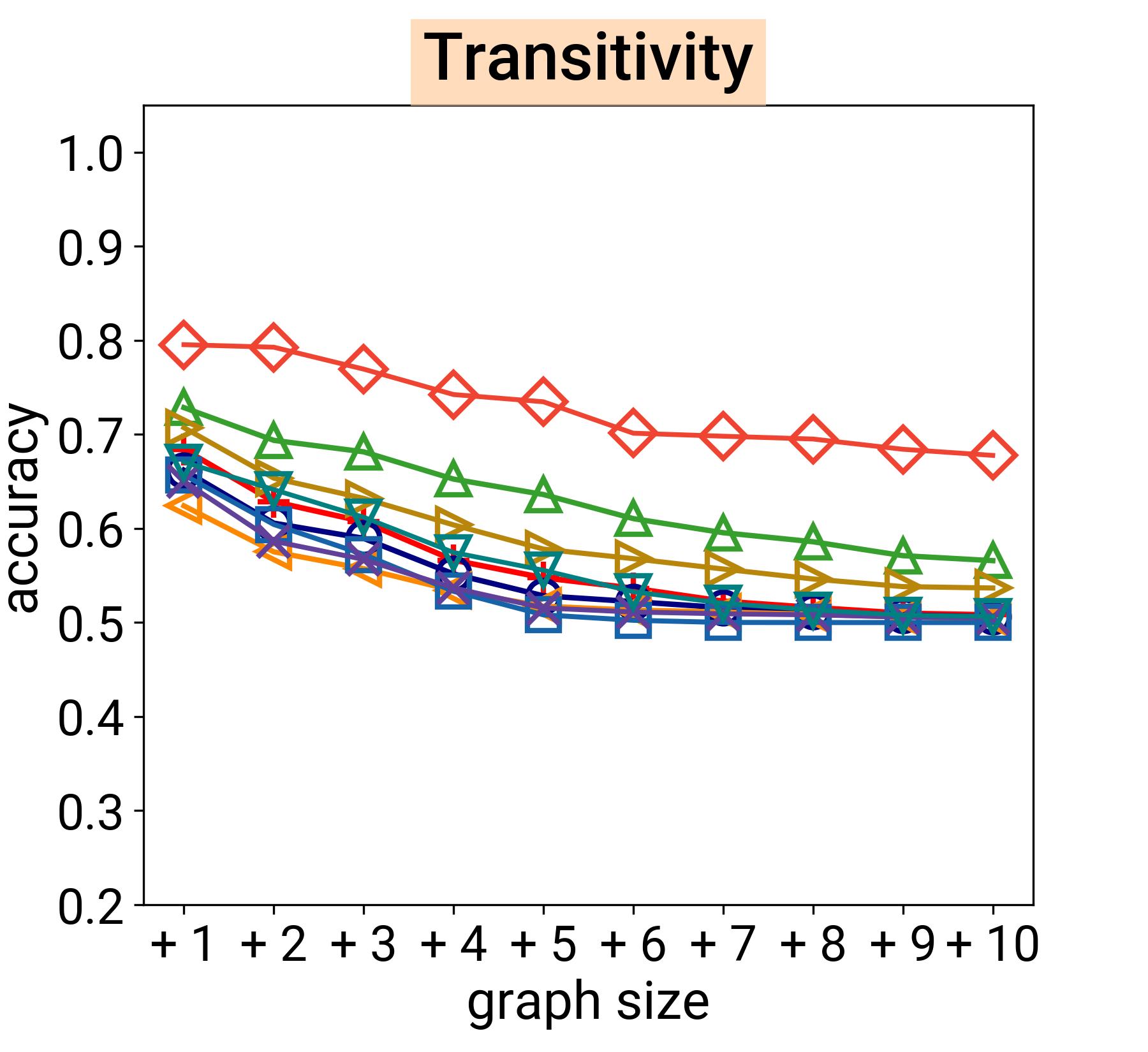}
  \end{subfigure}
    \begin{subfigure}[b]{0.24\textwidth}
    \includegraphics[width=\textwidth]{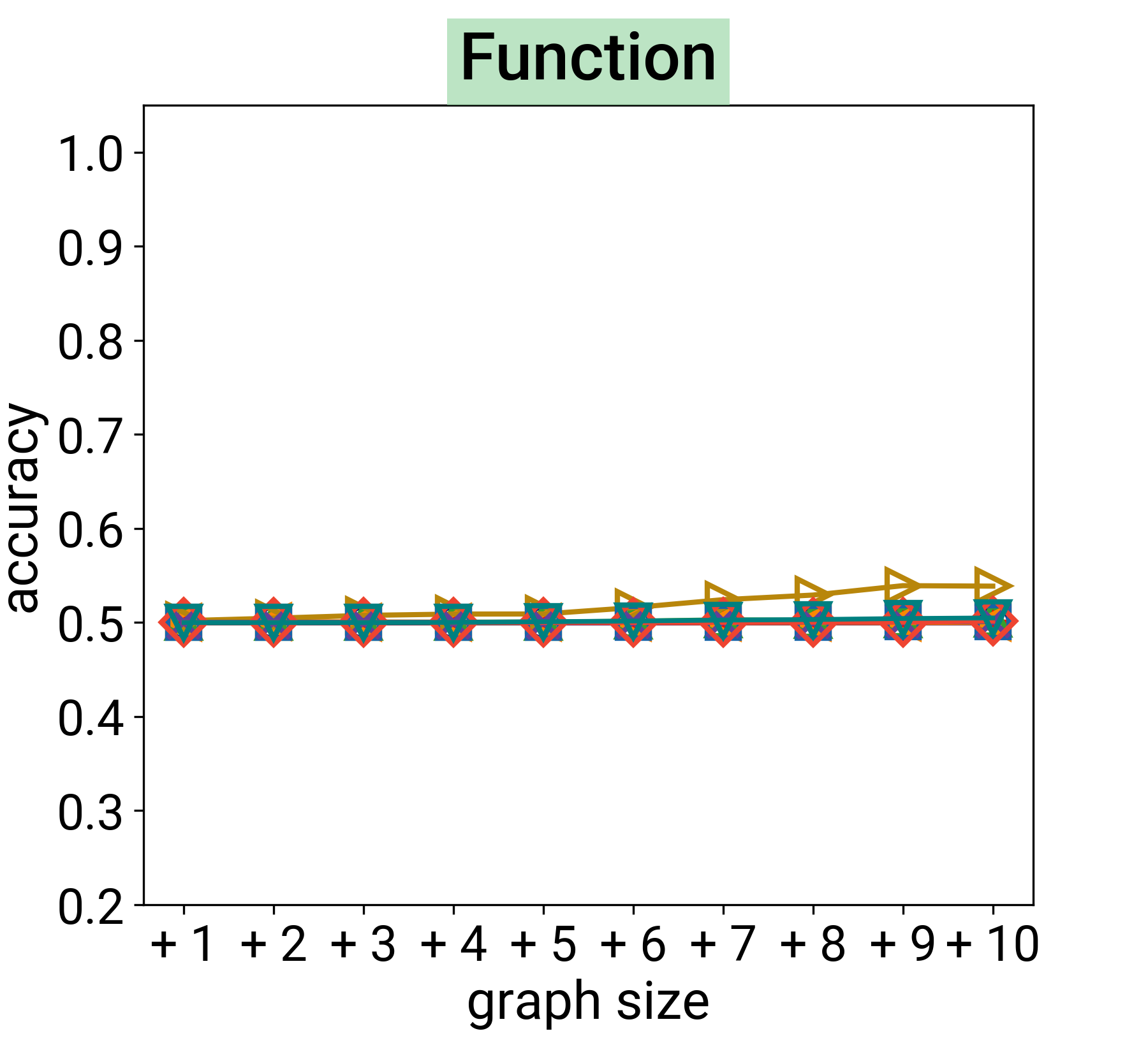}
  \end{subfigure}
    \begin{subfigure}[b]{0.24\textwidth}
    \includegraphics[width=\textwidth]{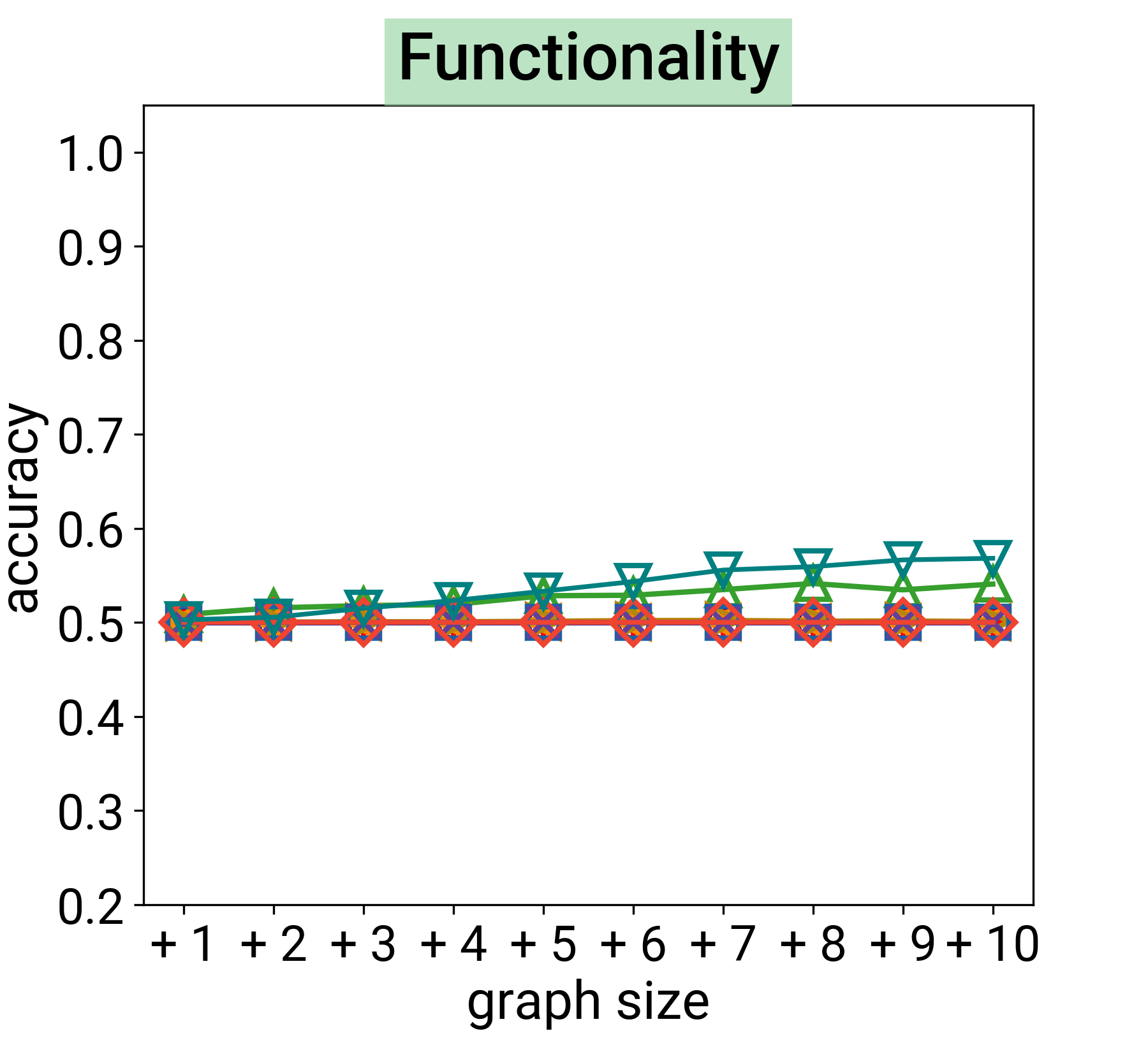}
  \end{subfigure}
    \begin{subfigure}[b]{0.24\textwidth}
    \includegraphics[width=\textwidth]{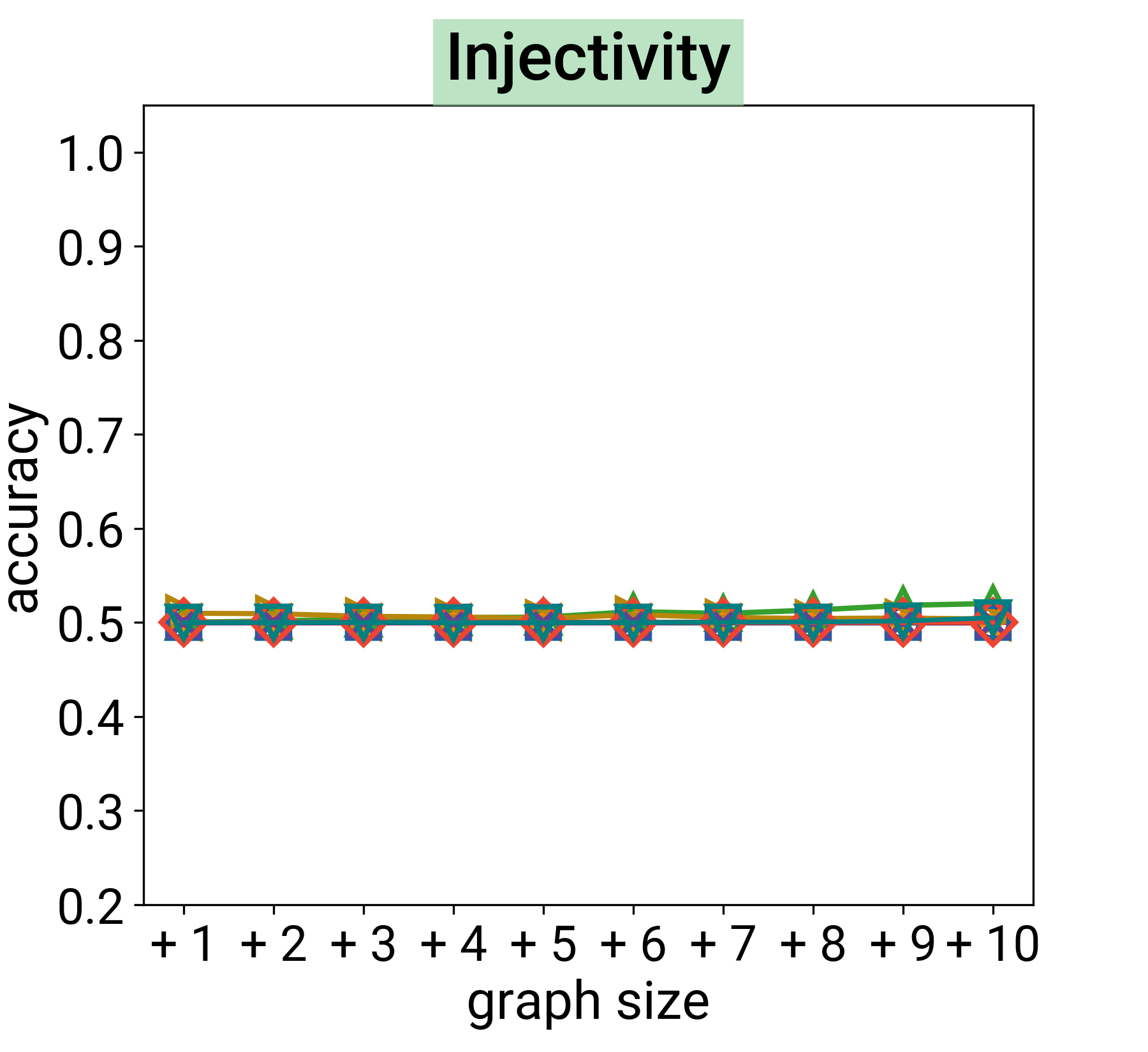}
  \end{subfigure}

   \par\smallskip
    \begin{subfigure}[b]{0.24\textwidth}
    \includegraphics[width=\textwidth]{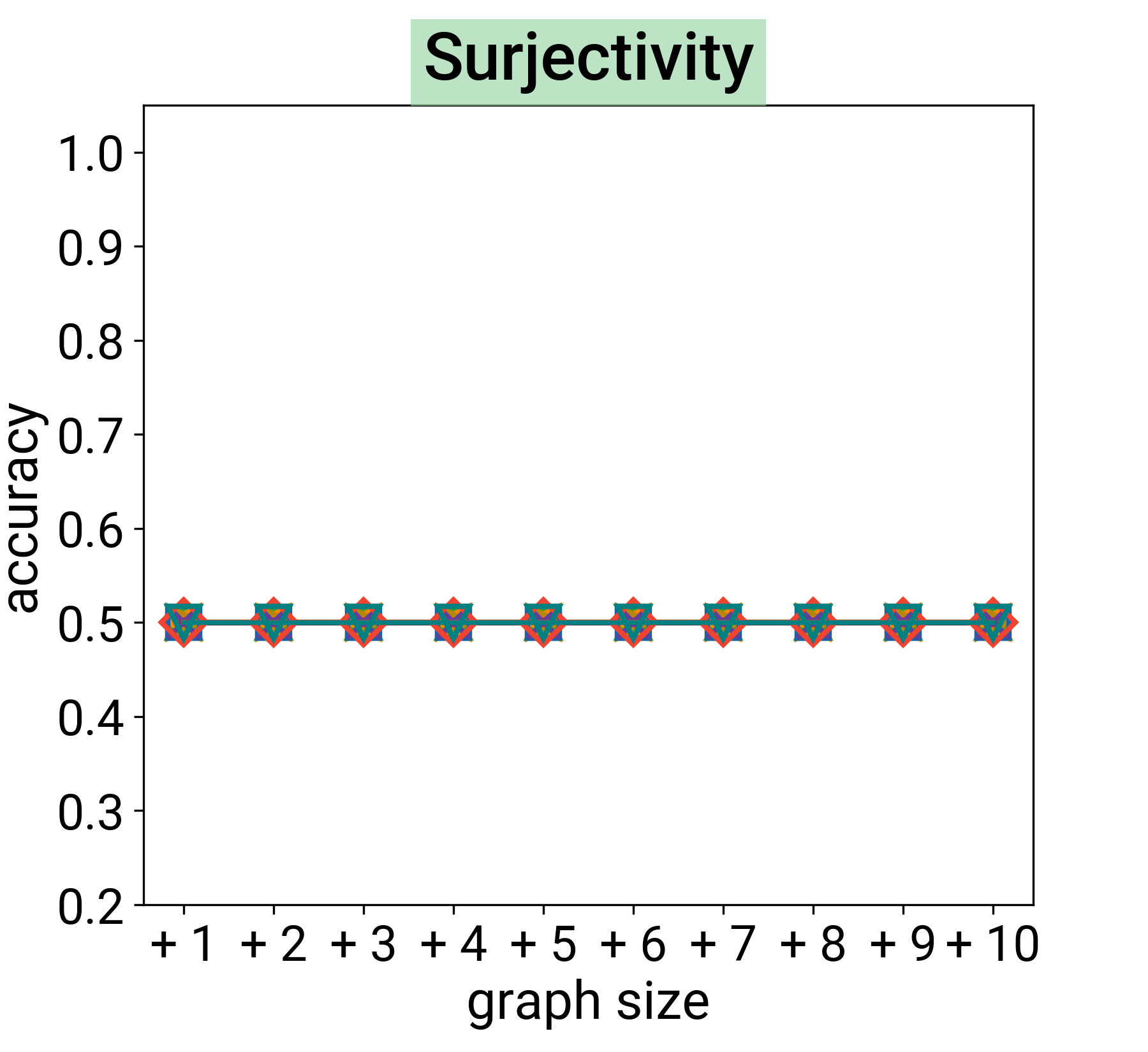}
  \end{subfigure}
  \begin{subfigure}[b]{0.24\textwidth}
    \includegraphics[width=\textwidth]{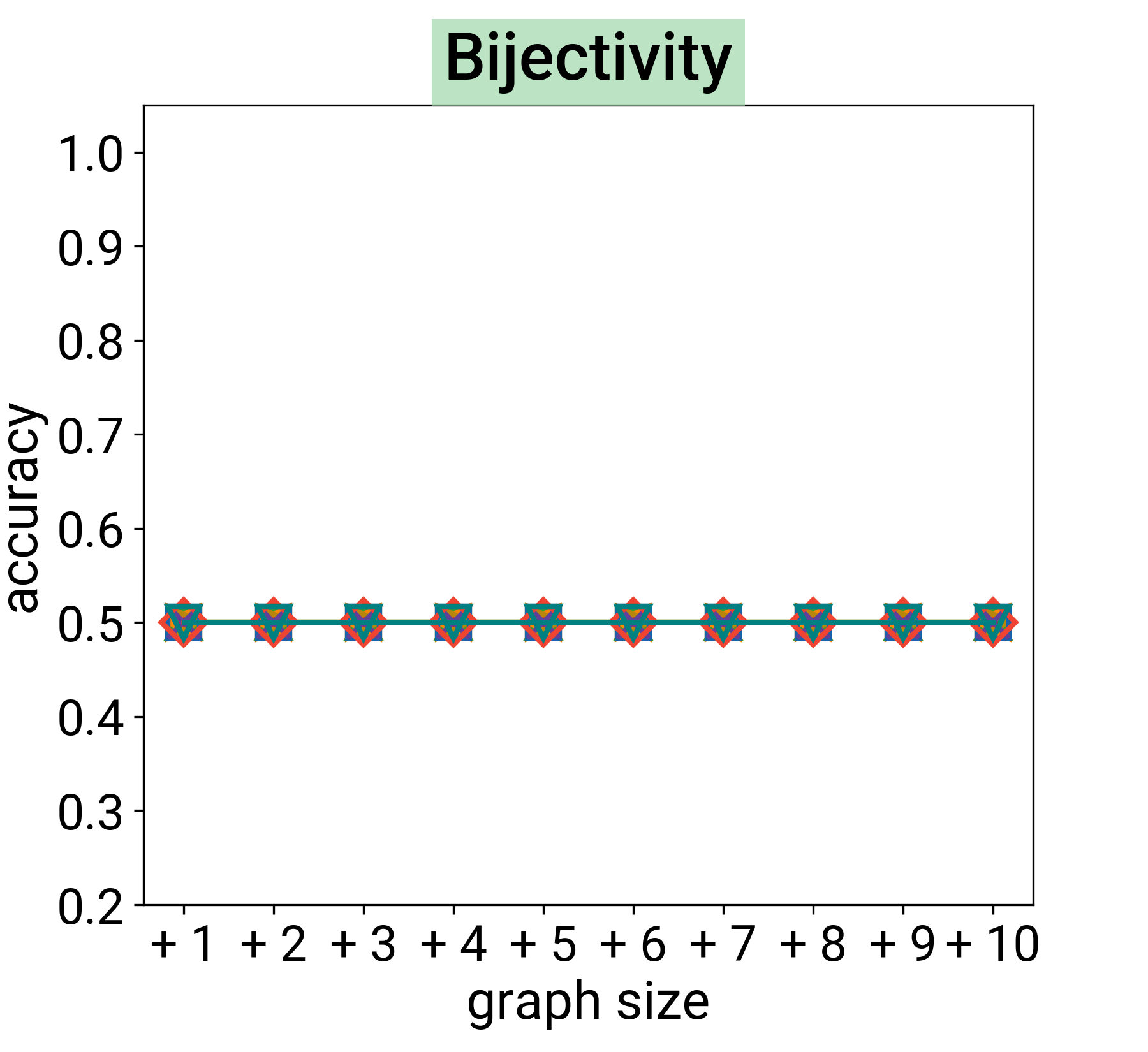}
  \end{subfigure}
  \begin{subfigure}[b]{0.24\textwidth}
    \includegraphics[width=\textwidth]{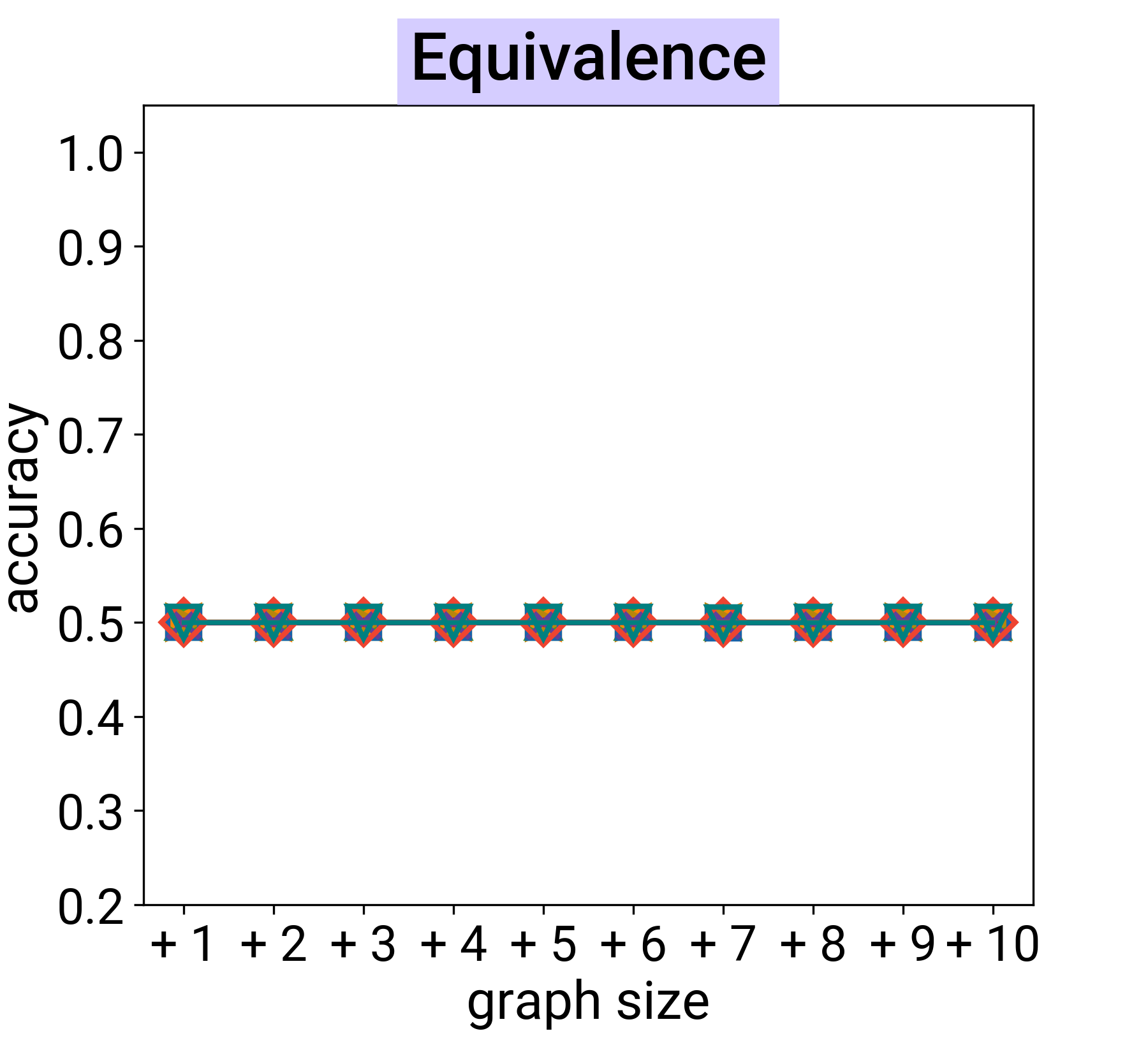}
  \end{subfigure}
    \begin{subfigure}[b]{0.24\textwidth}
    \includegraphics[width=\textwidth]{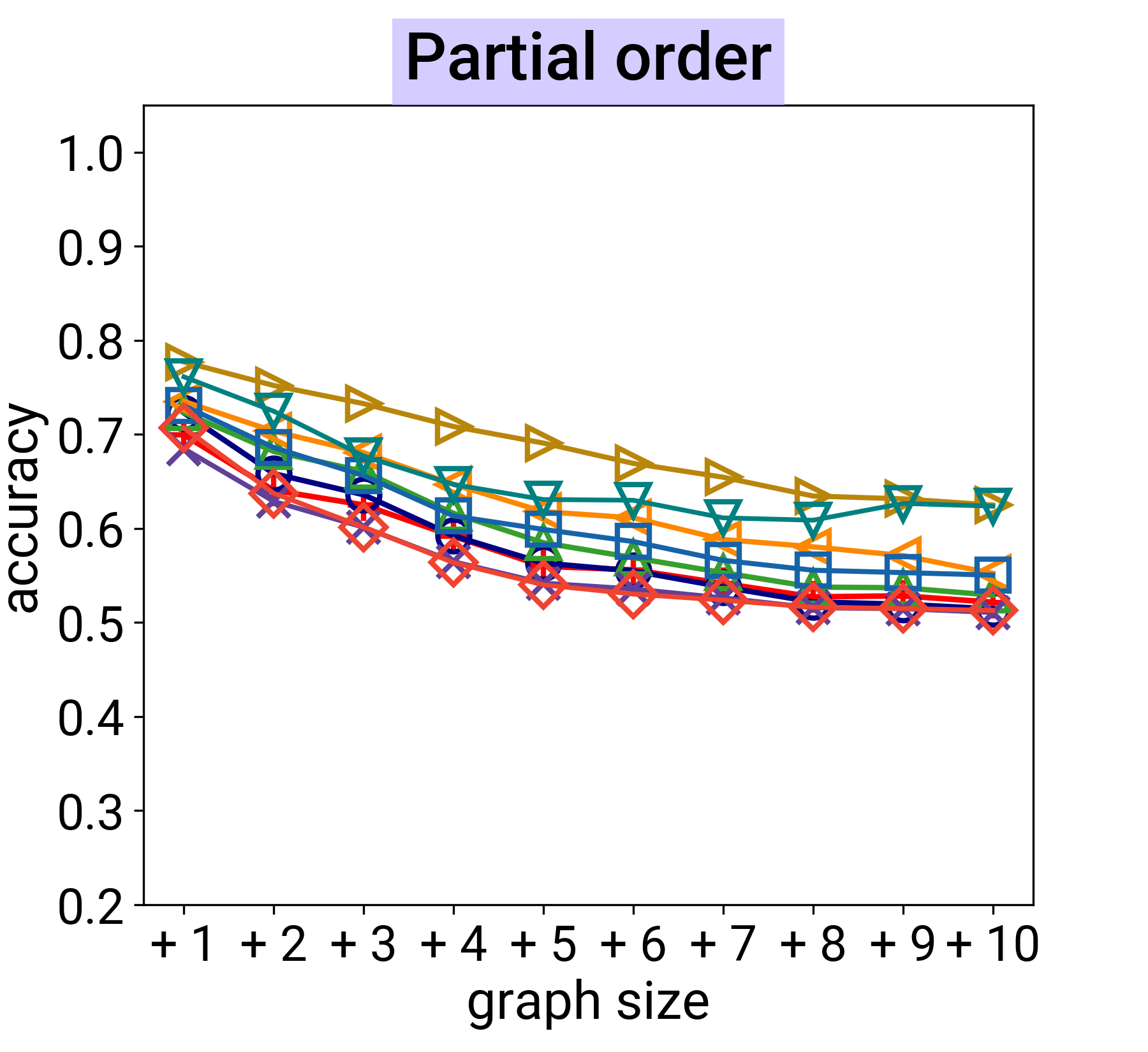}
  \end{subfigure}

  \par\smallskip
    \begin{subfigure}[b]{0.24\textwidth}
    \includegraphics[width=\textwidth]{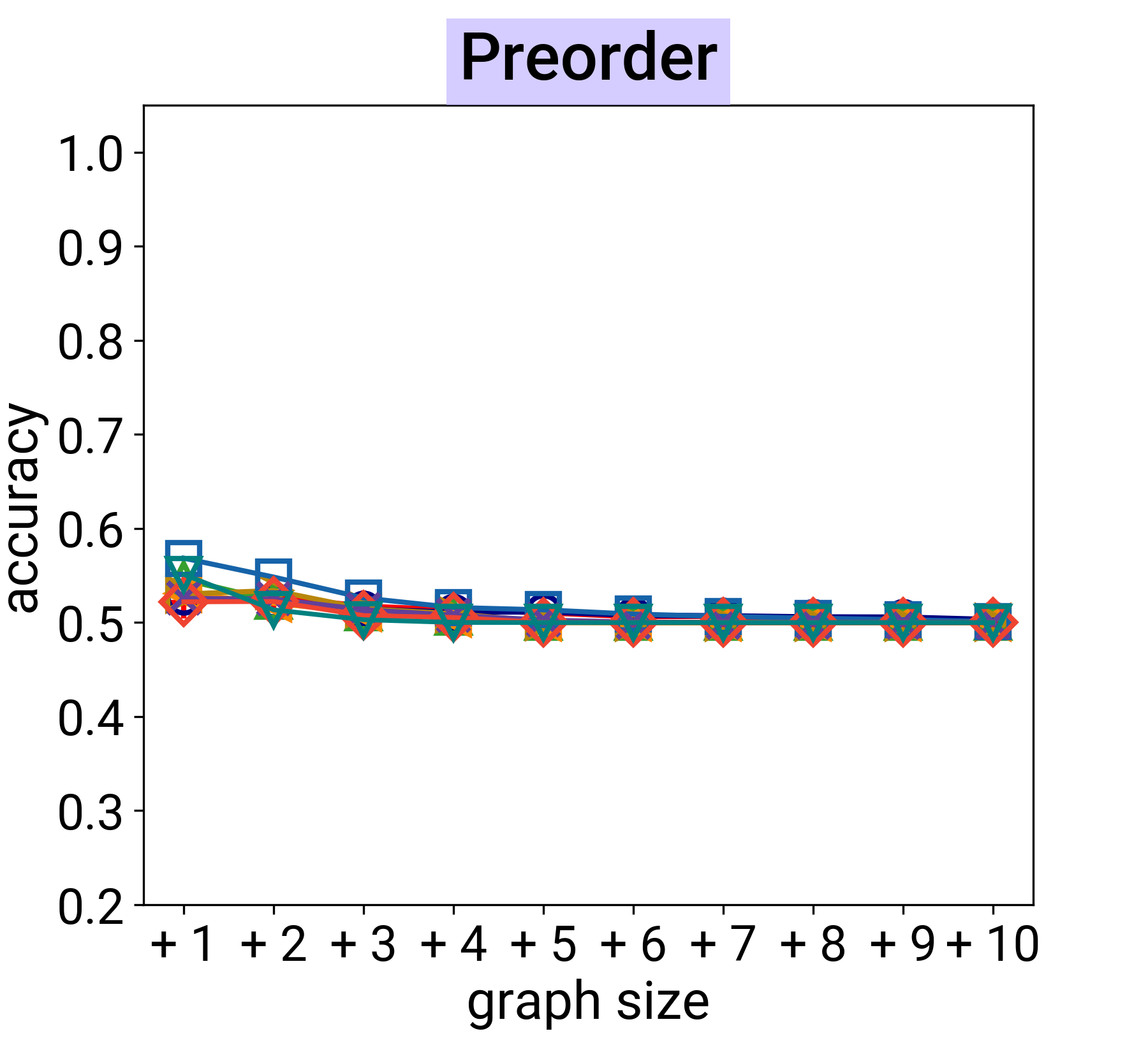}
  \end{subfigure}
    \begin{subfigure}[b]{0.24\textwidth}
    \includegraphics[width=\textwidth]{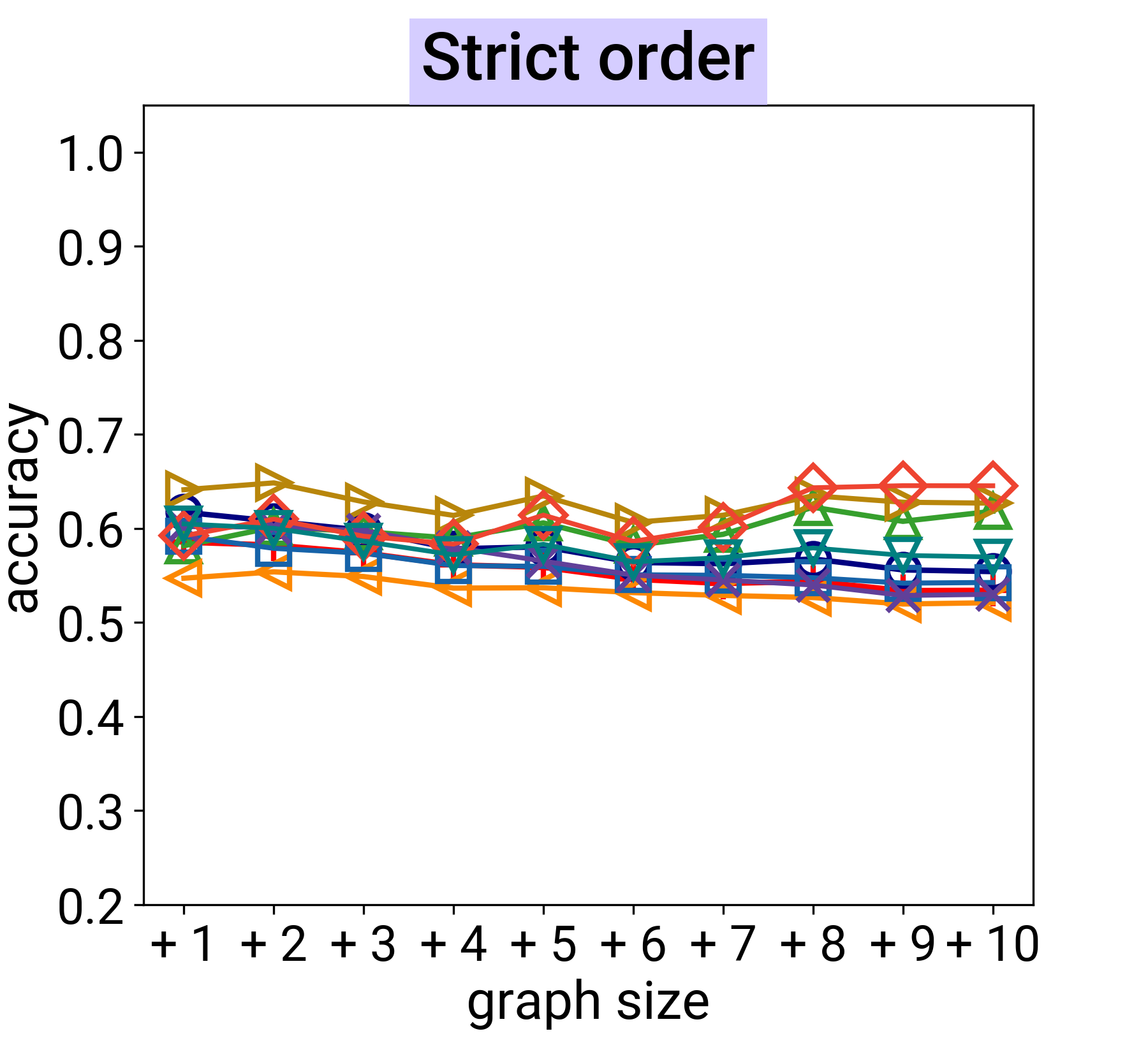}
  \end{subfigure}
  \begin{subfigure}[b]{0.24\textwidth}
    \includegraphics[width=\textwidth]{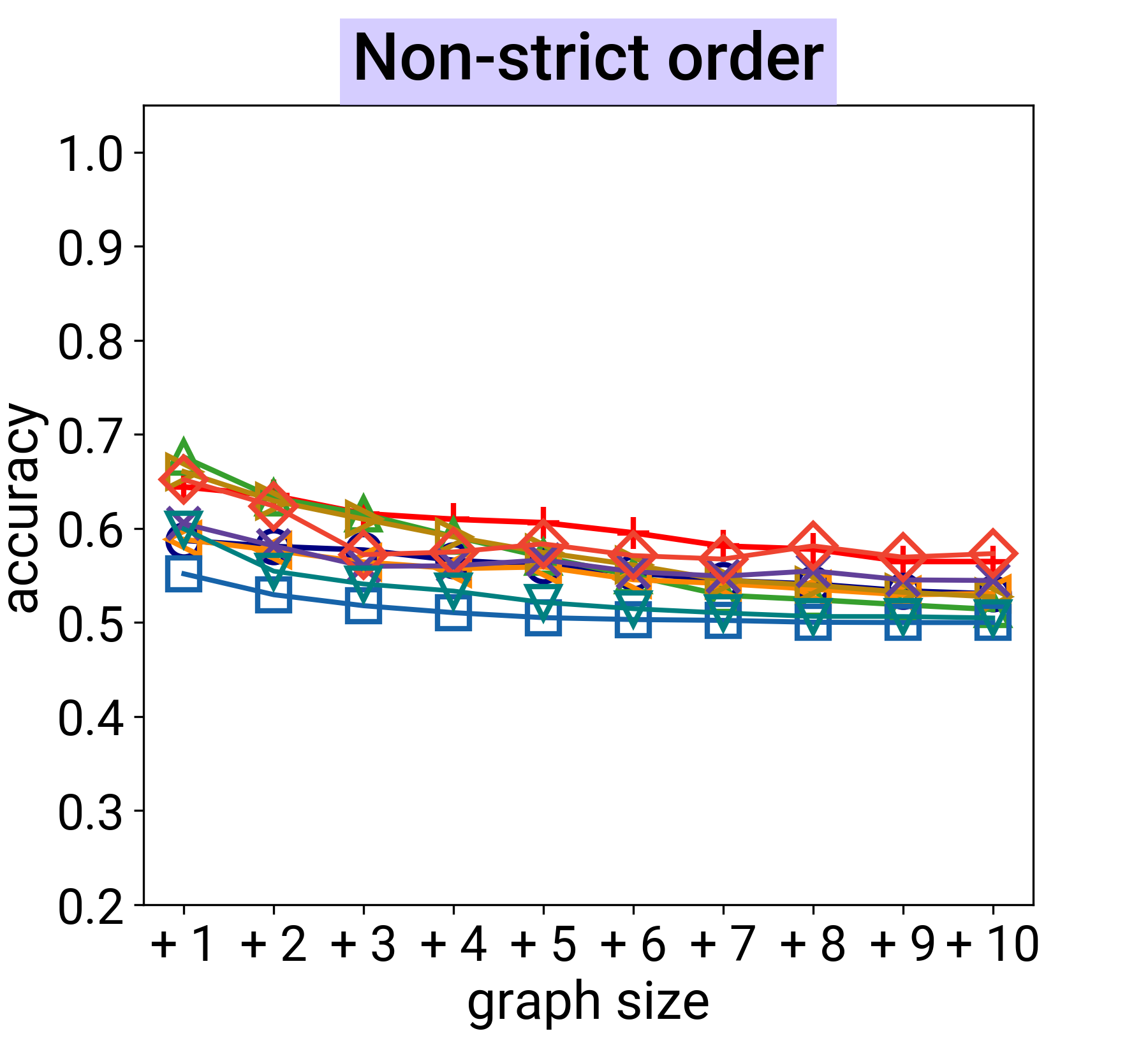}
  \end{subfigure}
    \begin{subfigure}[b]{0.24\textwidth}
    \includegraphics[width=\textwidth]{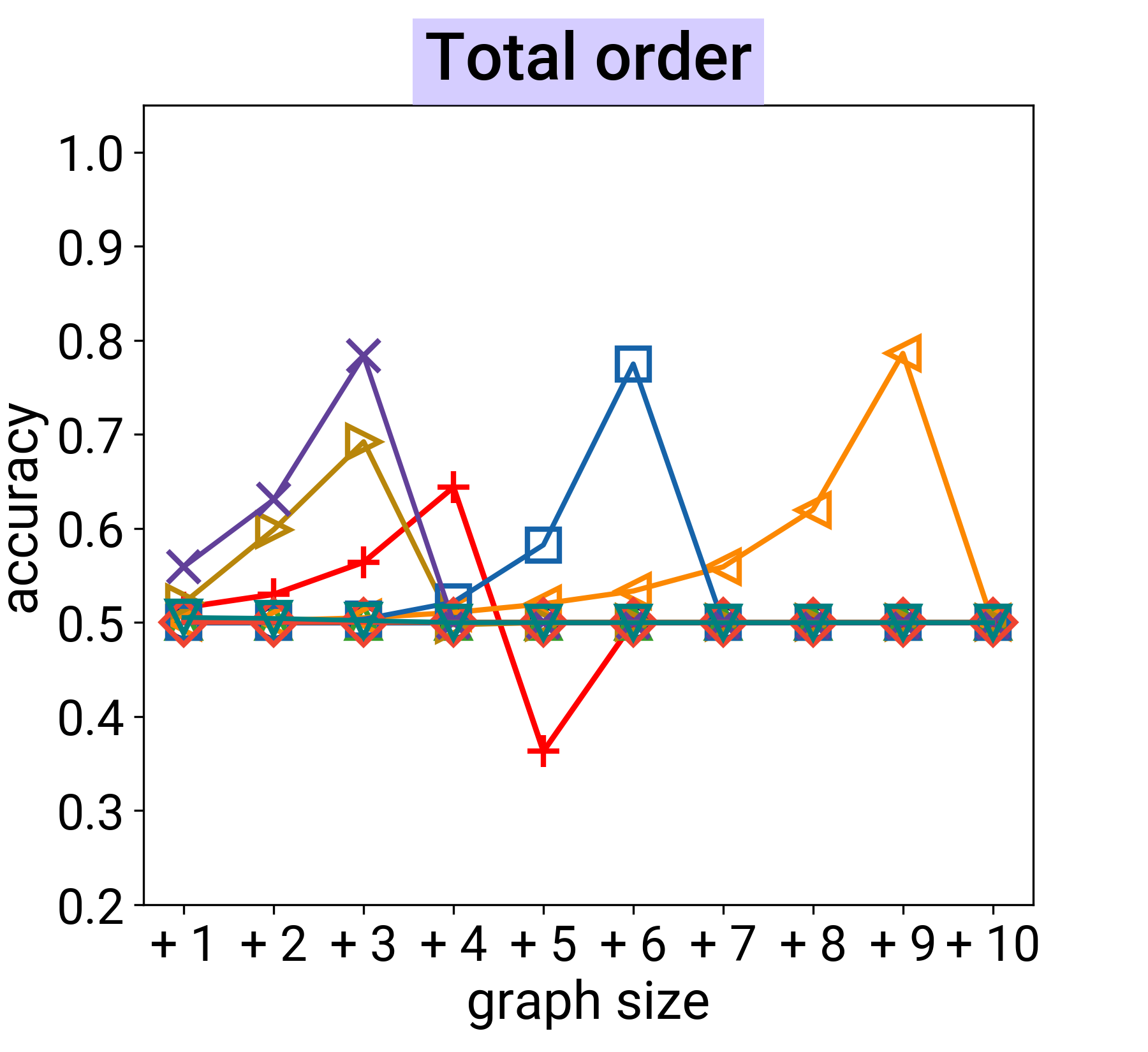}
  \end{subfigure}
  \vspace{-2ex}
  \caption{Global pooling performance across ten graph sizes under robustness aspect.}
  \vspace{-3ex}
  \label{fig:line_robustness}
\end{figure*}

Figure~\ref{fig:line_generalizability},~\ref{fig:line_sensitivity} and ~\ref{fig:line_robustness} illustrate how the performance of global pooling methods change with graph size.


